\documentclass{article}

\usepackage[final, main]{neurips_2026}

\usepackage{graphicx}
\usepackage[utf8]{inputenc}
\usepackage[T1]{fontenc}
\usepackage[english]{babel}
\usepackage{xcolor}
\usepackage{enumitem}
\usepackage{todonotes}
\usepackage{booktabs}
\usepackage{siunitx}
\usepackage{float}

\usepackage{amsmath}
\usepackage{tikz}
\usepackage{adjustbox}
\usetikzlibrary{positioning, shapes.multipart, arrows.meta}
\usepackage{listings}
\usepackage{subcaption}
\usepackage{hyperref}

\makeatletter
\renewcommand{\@notice}{}
\makeatother

\title{DARWIN: a Geospatial Foundation Model for Marine Intelligence}

\author{%
  Coen van den Elsen\thanks{Corresponding author: \texttt{coen@oceanos.earth}} \\
  Ocean OS \\
  \And
  Laurens Geffert \\
  Ocean OS \\
}

\begin{document}

\raggedbottom

\maketitle













\begin{abstract}
We present DARWIN, a geospatial foundation model for marine intelligence. Recent Earth-observation foundation models have concentrated on terrestrial domains. The marine realm poses distinct challenges: missingness, multi-modal physical and biogeochemical variables, sparse labels, and dynamics that are transient, multi-scale, and non stationary over time. DARWIN adapts a masked-autoencoder backbone to these properties through missingness-aware encoding, expanded spatial and temporal masking, and Fourier-style positional embeddings, and is pretrained on 26 years of global Copernicus reanalysis together with GEBCO bathymetry.

On predicting chlorophyll-\textit{a}, the pretrained backbone attains the lowest monthly relative error in every month of 2022, outperforming both no-encoder baselines and an identically architected random-init encoder. At 10\% of the supervised training data, DARWIN already surpasses the random-init model trained on 100\% of the data. Embedding analysis and qualitative reconstructions show that DARWIN captures coherent oceanographic structure rather than memorising input statistics, providing a scalable backbone for downstream marine applications such as anomaly detection, species distribution modelling, and environmental monitoring.
\end{abstract}

\section{Introduction}

The global oceans cover most of the Earth's surface. Oceans play a critical role in providing food, regulating the global climate, storing carbon, and supporting biodiversity \citep{roberts2017marine, hossain2019future}. However, we have a better understanding of the surface of Mars than we do of parts of the deep sea. If we want to make informed and sustainable decisions in the marine industries, then we need to improve our understanding of the marine realm.

Artificial intelligence (AI) has recently emerged as a powerful tool in various disciplines, including environmental science \citep{fan2023deep, ojji2024emerging, Joppa2017AIEnvironment}. Its ability to process vast amounts of data and uncover complex patterns has made it an invaluable asset in tackling real-world challenges. Despite its potential, traditional AI models face significant limitations. Increasing amounts of money are being spent on developing and training ever larger models with more and more data \citep{szwarcman2024prithvi, mai2023opportunities}. Acquiring labelled datasets remains a costly and labor-intensive process. Traditional task-specific models require extensive labelled data, but struggle to generalize across domains and variables, limiting their adaptability and scalability. This limitation requires the construction of new models for each specific application, further increasing the demand for resources.

With advancements in machine learning, particularly in deep learning and self-supervised learning, foundation models (FM's) have started to redefine how problems are approached \citep{achiam2023gpt, baris2025foundation, huang2024large, fu2024championing}. These are large models trained on large-scale datasets and have achieved state-of-the-art (SOTA) performance on a diverse range of tasks. Unlike traditional task-specific models that require full training for each narrowly defined task, foundation models are pre-trained on massive datasets and later adapted through fine-tuning or few-shot/zero-shot learning \citep{mai2023opportunities}. This approach enables the transfer of knowledge between tasks, significantly reducing the volume of labelled data needed for each, while improving the versatility of the model \citep{mai2023opportunities}. These models can be scaled efficiently and generalize well, eliminating the need to develop separate task-specific models for each application. Additionally, they can adapt across different domains, enhancing their versatility \citep{gao2025units, jakubik2023foundation, szwarcman2024prithvi, feng2025tessera}. This task-agnostic paradigm has unlocked new levels of performance and emergent behaviors, particularly in natural language processing, and is now gaining traction in earth science and remote sensing \citep{zhang2024geoscience}. Foundation models are able to outperform traditional models, due to their scalability and generalization abilities \citep{zhang2024opportunities, szwarcman2024prithvi, bodnar2024foundation}.

Researchers are increasingly exploring how foundation models can serve as generalist AI systems capable of addressing diverse environmental and geospatial tasks \citep{szwarcman2024prithvi}. Models are being used to analyze and interpret massive geospatial datasets, offering new insights into climate modelling, weather prediction, and land-use change detection \citep{price2023gencast, nguyen2023climax, lam2022graphcast, pathak2022fourcastnet}. These recent efforts have demonstrated that the use of foundation models can improve the traditional physics and AI-driven analysis and can extract more meaningful insights from remote sensing data sources \citep{szwarcman2024prithvi, price2023gencast, nguyen2023climax, lam2022graphcast, pathak2022fourcastnet}. Foundation models represent a shift from data gathering to data understanding, allowing AI systems to extract meaningful insights from complex environmental datasets. As research in this field progresses, these models have the potential to revolutionize geospatial analysis, making marine application data products more accurate, scalable, and efficient.

Existing AI models for oceanic analysis still face these limitations: labelled datasets are especially scarce in oceanographic applications due to the high cost of collecting ground-truth data, and models trained end-to-end on specific tasks lack the generalization needed across the diverse marine observation space \citep{szwarcman2024prithvi}. To address these challenges, we introduce DARWIN, a novel geospatial foundation model specifically designed for marine applications. Additionally, most existing AI models are not designed to handle the unique characteristics of marine environments, including dynamic and highly variable conditions, making their application to ocean monitoring and climate studies less effective.

\section{Background}
\subsection{Foundation Models in the Geospatial Domain}

Foundation models (FMs) are large, generalist models pre-trained on massive amounts of unlabelled data using self-supervised learning. Originally successful in natural language processing and computer vision, they are now transforming the geospatial domain by enabling broad generalization across tasks and sensor modalities. Unlike traditional task-specific models, FMs offer a reusable base that can be fine-tuned for downstream applications with minimal labelled data, an advantage particularly important in remote sensing, where annotation is costly and sparse \citep{Lauro2014Oceanographer}.

Recent advances have introduced several geospatial foundation models, demonstrating strong performance in tasks such as land cover classification, land use change detection, and weather forecasting. Prithvi has demonstrated the effectiveness of self-supervised learning for satellite imagery analysis, improving segmentation, classification, and forecasting accuracy \citep{szwarcman2024prithvi}. Similarly, SatMAE \citep{cong2022satmae} and Scale-MAE \citep{reed2023scale} leverage transformers to learn spatial and temporal dependencies from vast datasets, reducing the need for extensive labelled data while achieving superior performance across geospatial tasks. \citet{dawson2025sentinel} makes a foundation model for ocean color that shows improved performance on downstream tasks by treating the ocean as an image instead of a dynamical system. Terramind \citep{jakubik2023foundation} is the first truly generative multimodal model in the domain and shows SOTA performance. Aurora \citep{bodnar2024foundation} demonstrates effectiveness in high-resolution weather forecasting tasks at computational expense that is orders of magnitude smaller than existing numerical modelling systems. Terrestrial foundation models have numerous standardised benchmarks that they can be tested on, like Geo-Bench \cite{lacoste2023geo} and PANGAEA \cite{marsocci2024pangaea}. Clay \citep{clay2025} is an example of a flexible architecture creating embeddings optimized for search and numerous other tasks. Earthnet shows significant promise in high-frequency data assimilation and weather forecasting \citep{vandal2024global}. Tessera, AlphaEarth Foundations, Terramind, and Olmoearth all achieve SOTA performance metrics on different downstream tasks with new training techniques \citep{brown2025alphaearth, feng2025tessera, herzog2025olmoearth}, \citep{jakubik2025terramind}, \citep{tseng2026olmoearth}. The ocean realm does not have standardised benchmarks for foundation models yet.

\subsection{Self-Supervised Learning for Geospatial Representation Learning}

Training foundation models on large quantities of unlabelled data requires learning rich representations without manual annotation; Self-supervised learning (SSL) has emerged as a powerful paradigm for this, eliminating the need for extensive labelled datasets by leveraging intrinsic structures in the data.  SSL enables models to learn useful and generalizable representations by solving tasks that do not require labelled data. This makes it ideally suited for geospatial data, which often span multiple modalities and temporal scales. The paper on data2vec \citep{baevski2022data2vec} is a study that has been the basis of the later SSL paradigm. \citet{lee2021predicting} analyse why SSL is a good learning strategy, this can be seen as Deep Learning theory. The self-supervised domain consists of different techniques to learn representations, among which are masked autoencoders (MAE)\citep{he2022masked}, masked prediction of hidden units \citep{hsu2021hubert}, joint embedding architectures (JEA)\citep{caron2021emerging, oquab2023dinov2}, and  joint embedding predictive architectures (JEPA) \citep{assran2023self}.

In the geospatial domain, inspired by image processing, a widely adopted strategy is the Masked Autoencoder (MAE) framework \citep{he2022masked}. This framework is designed to reconstruct the original signal from partial observations \citep{he2022masked}. By dividing an image into non-overlapping patches, MAE randomly masks a significant portion of them, allowing the encoder, a Vision Transformer (ViT) \citep{dosovitskiy2020image}, to process only the visible patches. This drastically reduces computational complexity. A lightweight decoder reconstructs the missing regions using a reconstruction loss, often measured with Mean Squared Error (MSE). This method has proven highly effective for learning scalable, rich representations \citep{he2022masked, yue2023understanding}.

\subsection{Geospatial FM Architectures}

Geospatial foundation models (FM) are built on Vision Transformer (ViT) backbones \citep{dosovitskiy2020image}. Most of these models adopt an encoder-only or encoder-decoder architecture during pretraining. Unlike convolutional models, ViTs require positional encoding to capture the structure of the input. Standard ViTs use 2D positional encodings appropriate for images, but these are insufficient for higher-dimensional geospatial inputs, such as multi-temporal or multi-modal satellite imagery. Consequently, spatiotemporal positional encodings have been introduced to model three-dimensional input structures. For example, SatMAE employs separate sinusoidal embeddings for the X and Y spatial dimensions, which are concatenated and added to the input tokens to preserve spatial locality. Prithvi-1.0 uses 1D sinusoidal embeddings separately for height, width, and time, then combines them to capture spatiotemporal dependencies in multi-spectral satellite data \citep{jakubik2023foundation}. Prithvi-2.0 integrates geolocation and time metadata via a learned weighted sum of 2D sinusoidal embeddings, enabling the model to dynamically adjust the importance of metadata information \citep{szwarcman2024prithvi}. \cite{bodnar2024foundation} use a similar approach.

\subsection{Gaps in the Current FM Landscape and the Marine Challenge}
\label{sec:marine_background}

Despite the significant progress in developing foundation models for Earth observation, there remains a critical gap in their applicability to marine environments. Most existing geospatial foundation models have been developed and benchmarked on terrestrial datasets, predominantly focusing on land use, vegetation, and infrastructure \citep{marsocci2024pangaea}, \citep{Philipp2023FoMoBench}. These standardised benchmarks do not exist for the ocean realm. While \citet{yang2025deep} and \citet{dawson2025sentinel} introduced a foundation model for ocean color, their approach treats satellite imagery as static snapshots, effectively ignoring the critical temporal dynamics of marine systems. Furthermore, they omit the integration of the diverse physical and biogeochemical variables that are essential for resolving the complex, multi-scale dynamics governing the marine environment.

There are several differences between the terrestrial and the marine realm, and marine environments pose challenges that differentiate them from land-based settings.

Firstly, terrestrial features are usually persistent over time (rivers, forests, houses, roads) while marine features are often ephemeral and transient in nature (ocean eddies, plankton blooms, El Niño–Southern Oscillation) \citep{McGillicuddy2016Mesoscale} and exhibit complex and multi-scale temporal behavior that demands higher spatiotemporal fidelity in modelling.

Secondly, land surface offers good signal for both active (radar) and passive (optic) remote sensing systems, allowing rich, large-scale data collection. In the marine realm, on the other hand, most sensor systems do not penetrate the entire water column, meaning that data has to be collected from vessels and interpolated via geospatial processing, data mining, or modelling \citep{kutser2020remote}.

Thirdly, while sampling biases are certainly also present in many terrestrial datasets, this is amplified in marine environments, where a majority of the data is from coastal waters, the top of the water column, and the global North \citep{Swart2023SouthernOcean}. Observational data in oceanography is inherently sparse and noisy, often derived from satellite altimetry, in-situ drifters, and biogeochemical sensors. These data sources are typically lower in spatial resolution and may be irregular in temporal coverage \citep{Wang2023Fusion, Swart2023SouthernOcean}. Scarcity of labelled datasets in marine science severely limits the application of supervised learning methods \citep{Lauro2014Oceanographer}. Unlike urban or agricultural domains, where annotated data is often available through aerial surveys or governmental mapping agencies, marine datasets require costly field campaigns and sensor deployments. This further underscores the need for foundation models that can leverage self-supervised learning to extract structure from unlabelled data.

Current geospatial FMs are also not optimized for the multi-modal nature of marine observations. Variables such as sea surface temperature, chlorophyll concentration, salinity, and ocean color are recorded using diverse sensors with varying spatial and spectral characteristics. Integrating such data requires architectures capable of cross-modal representation learning, which is still an emerging area in the FM landscape.

At the same time, there are a number of data-intense applications which could benefit greatly from marine foundation model support. Species distribution modelling, an application widely used by ecologists and conservationists, conventionally uses climatic datasets of monthly, yearly, or decadal averages, but these features fail to capture extreme events that substantially influence species distributions (e.g. coral bleaching) \citep{Dinnage2024NicheFlow}. Additionally, marine disasters such as oil spills or contamination plumes are currently predicted with the help of numerical models but the compute time of these models can slow down disaster response efforts. \\

\section{Method for Pretraining}

\subsection{Data}
\subsubsection{The Dataset}

Our primary data is sourced from the Copernicus Marine Service (CMEMS)~\citep{copernicus_marine}, encompassing 26 years (1998--2023) of global monthly mean observations. CMEMS integrates satellite missions (ESA, EUMETSAT) and in-situ measurements into open-access, systematically curated data products. We use Level~4 (L4) reanalysis products: fully processed, gridded, and retrospectively reconciled with observational data using models such as NEMOVAR~\citep{mogensen2012nemovar}. The specific products are shown in \autoref{tab:cmems_products}. These products are maintained by domain experts who handle noise, outliers, and sensor malfunctions. By using L4 data products, we rely on their expertise and judgment calls for the data preprocessing.

Our focus is on surface-level observations for most variables, excluding subsurface data. This choice was made to keep computational training costs within budget, and to avoid complexities associated with acquiring, storing, and modelling subsurface measurements, which are often sparse and come with a higher degree of uncertainty. We do, however, include several environmental variables that act as proxies for sub-surface dynamics such as bathymetry and mixed-layer-depth.

Bathymetry data is sourced from GEBCO \citep{gebco}. We choose the GEBCO product over the static bathymetry layer from the Copernicus Marine Service due to the latter's lower spatial resolution and preprocessing artifacts. Moving forward, bathymetry information is referenced as a \textit{static}, whereas climatological and physical ocean variables are referenced as \textit{dynamic}.

\begin{figure}[H]
  \centering
  \begin{subfigure}[t]{0.48\textwidth}
    \centering
    \includegraphics[width=\textwidth]{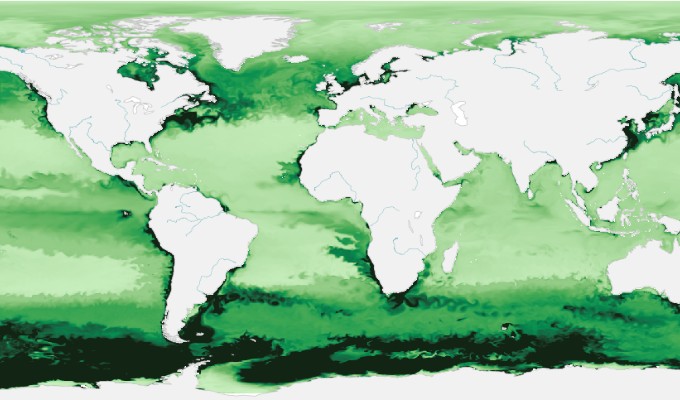}
    \caption{Mass concentration of chlorophyll in sea water (monthly mean)}
    \label{fig:chlorophyll}
  \end{subfigure}
  \hfill
  \begin{subfigure}[t]{0.48\textwidth}
    \centering
    \includegraphics[width=\textwidth]{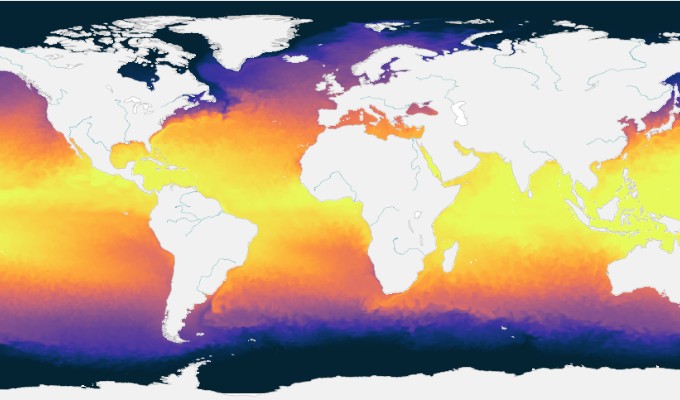}
    \caption{Sea water potential temperature (monthly mean)}
    \label{fig:temperature}
  \end{subfigure}
  \caption{Examples of variables in the dataset.}
  \label{fig:dataset_variables}
\end{figure}

The final pretraining dataset comprises 47 variables covering key physical, wind, biomass, color, and biogeochemical metrics (e.g., chlorophyll, oxygen, nutrients). The complete list of variables is shown in Appendix~\autoref{tab:copernicus_variables}.  It spans 312 monthly timesteps (1998–2023) on a \num{4316} $\times$ \num{2036} grid, where each cell corresponds to a $\tfrac{1}{12}^{\circ}$ horizontal resolution (approximately $8\,\mathrm{km} \times 8\,\mathrm{km}$ at the equator), providing high-resolution, consistent, and comprehensive coverage for geospatial model pretraining. Examples can be seen in \autoref{fig:dataset_variables}. Within the dataset, all dynamic variables corresponding to land grid cells (as defined by GEBCO bathymetry) are designated as NaN; no missing values exist for dynamic ocean cells. Further details regarding dataset construction, corrections, preprocessing, and storage optimizations are provided in \autoref{dataset_pipeline}.

\begin{table}[H]
\centering
\caption{Copernicus Marine Service (CMEMS) Global Ocean Data Products used in pretraining.}
\label{tab:cmems_products}
\begin{tabular}{lll}
\toprule
\textbf{Category} & \textbf{Product Name} & \textbf{Processing Level} \\
\midrule
\textbf{Physics} &  Physics Reanalysis & L4 (Reanalysis) \\
\textbf{Biogeochemistry} & Biogeochemistry Hindcast & L4 (Hindcast) \\
\textbf{Biomass} &  Low and Mid Trophic Levels Biomass & L4 (Hindcast) \\
\textbf{Ocean Colour} & Colour (Copernicus-GlobColour) & L4 (Interpolated) \\
\textbf{Wind} & Monthly Mean Sea Surface Wind and Stress & L4 (Satellite-Model) \\
\bottomrule
\end{tabular}
\end{table}


\subsubsection{Sampling}

\begin{figure}[H]
  \centering
  \includegraphics[width=0.9\textwidth]{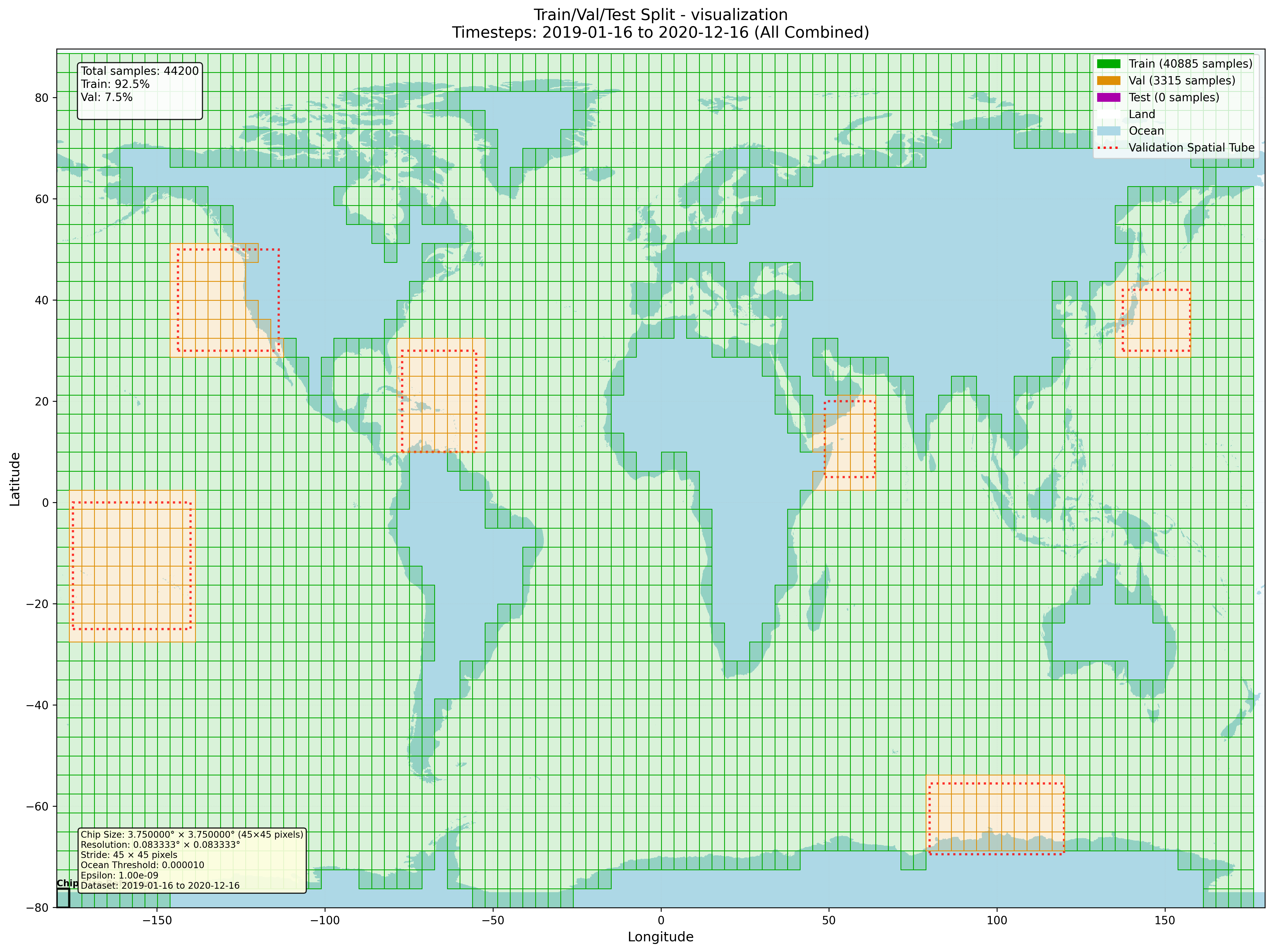}
  \caption{Training and validation split}
  \label{fig:data_splits}
\end{figure}

To train the model, we extract spatiotemporal chips of size $45 \times 45 \times 12$ (latitude × longitude × time), representing one year of observations over a 360×360~km region. This window size is chosen as it balances spatial context with seasonal temporal coverage. We generate spatially non-overlapping tiles, which have a stride of 1 timestep to cover all seasonal patterns. After removal of all the land-only tiles, we have \num{1023400} samples. We first experimented with a 90/10 random split for validation, but found that in some cases the model's overfitting was hidden. We therefore select a diverse set spatiotemporal holdout to properly examine how the model generalizes to unseen regions, as can be seen in \autoref{fig:data_splits}. The final partition yields \num{946645} training and \num{76755} validation samples.

\subsection{Modelling approach}

\begin{figure}[H]
  \centering
  \includegraphics[width=\textwidth]{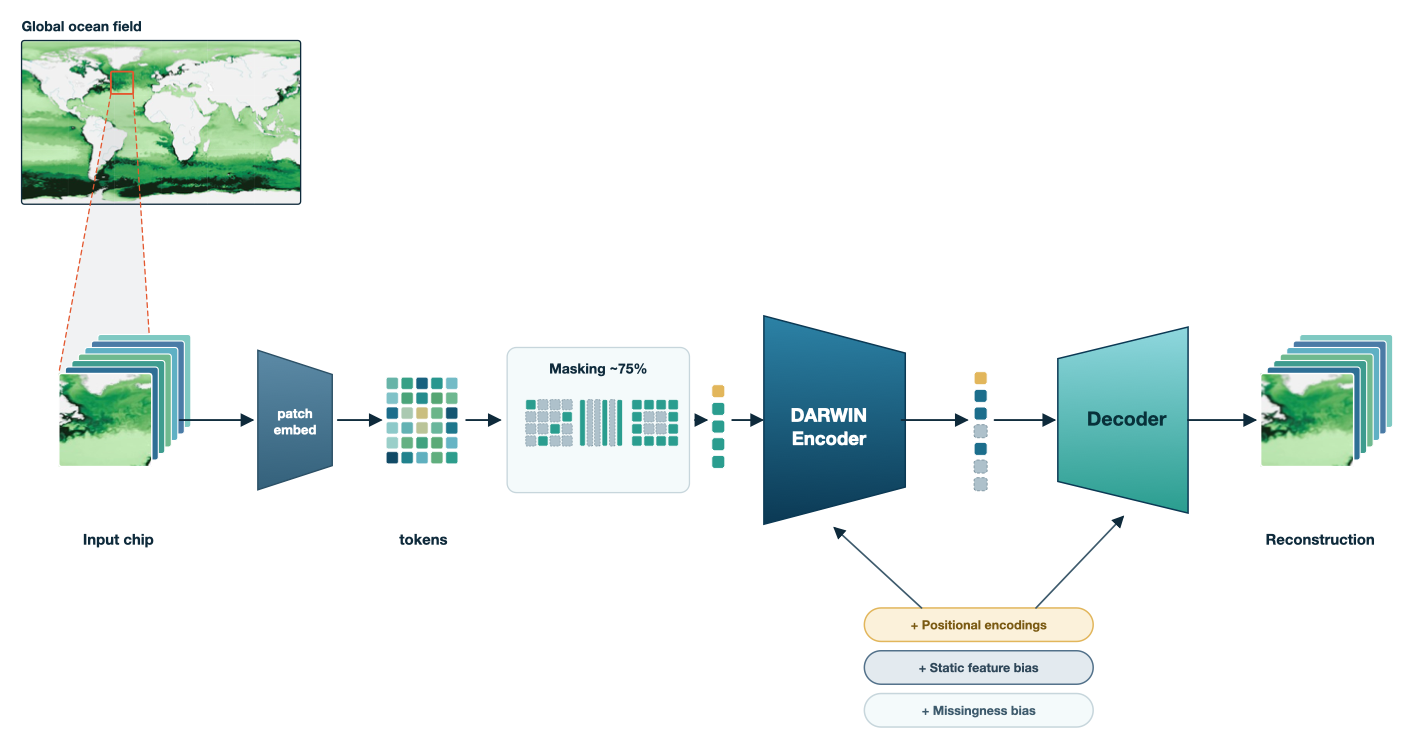}
  \caption{DARWIN architecture overview. For visualisation purposes, the time dimension is omitted.}
  \label{fig:architecture}
\end{figure}

DARWIN retains the masked autoencoder (MAE) backbone introduced by Prithvi \citep{szwarcman2024prithvi}: a Vision Transformer (ViT) encoder–decoder operating on 3D spatiotemporal patches. Inputs are divided into non-overlapping patches using an embedder, and the encoder processes only the unmasked subset. The decoder reconstructs masked patches, with loss computed exclusively over masked ocean pixels. A schematic of the overall MAE framework is shown in \autoref{fig:architecture}.

We modify this approach with the following: \\

\begin{itemize}
    \item \textbf{Missingness-aware encoding.} We feed the model all variables per pixel together with a binary missingness indicator and binary land indicator per variable. This approach biases the model to learn about missingness and whether it is caused by land masking or lack of information. We only decode actual variables and compute loss exclusively over observed ocean pixels, so that the model can learn \emph{aware} of missingness without being asked to reconstruct it. Although our ablations did not show measurable improvement in MAE, we retain this component as it provides an explicit structure for downstream tasks where missingness patterns are informative and the MAE reconstruction loss itself is not a reliable indicator for potential downstream performance.

    \item \textbf{Static encodings.} We want the model to learn about the complex dynamics and interdependencies between environmental variables. However, it would be unreasonable to expect the model to learn static variables such as bathymetry for spatiotemporal holdout regions. We also do not want geographic info to leak into the embeddings in an unwanted way, as that can cause spatial overfitting. We therefore want the ability to add bathymetry to the encoder and decoder independently and also on the masked patches to inform the spatiotemporal patterns of the dynamic variables. That is why we did not concatenate static variables (ocean mask and bathymetry min/max/mean/std) to the input data but instead added the information as a bias to the embeddings of both the encoder and the decoder. This allows the model to condition on static context wherever useful, while leaving it free not to dedicate capacity to re-representing static geographic information. Static channels are patchified by a 3D convolution, projecting from the $N_\text{static}{=}5$ input channels to the model's embedding dimension $D$ (with separate, learned weights in the encoder and the decoder); the resulting per-spatial-patch embedding is added to the dynamic patch tokens. This design allows static embeddings to be included or omitted at inference, giving downstream tasks explicit control over whether geographic context is injected. This is useful when spatial generalization is required and overfitting to known locations must be avoided. Although pretraining ablations did not show significant MAE pretraining difference, we expect this flexibility to matter in spatiotemporal holdout evaluations and embedding quality.

    \item \textbf{Expanded masking strategies.} We introduce temporal-slice masking (entire months removed) and spatial tube/block masking (large contiguous regions masked) in addition to random masking. During initial experimentation we noticed that value interpolation from spatial neighbours dominated the reconstruction. This forces the model to learn both complex variable interdependencies and temporal dynamics. We hypothesize that this helps with seasonal continuity and mesoscale structure, improving long-range interpolation and reasoning (see \autoref{appendix:masking_strategies} for details).

    \item \textbf{Improved positional encodings.} {We add temporal, spatial, and scale awareness by replacing standard sinusoidal encodings with Fourier-style embeddings for chip-level latitude, longitude, and time-slice hour of year (see \autoref{appendix:positional_encodings} for details). We refer to the implementation of \citet{bodnar2024foundation} for further details. We add chip-scale to mitigate latitude bias. In addition, we add a year positional encoding to capture non-periodic long-term trends such as climate change drift. We implement all of this with a dropout to force the model to generalize instead of learning a lookup table.}
\end{itemize}

\subsection{Loss}
The loss function is defined as:

\[
\mathcal{L}_{\text{MSE}} = \frac{1}{|M|} \sum_{i \in M} \left( x_i - \hat{x}_i \right)^2
\]

where \( M \) is the set of all pixel indices in masked, ocean-only regions, \( x_i \) is the ground truth, and \( \hat{x}_i \) is the model's reconstruction.\\

The objective is to minimize the Mean Squared Error (MSE) between the reconstructed and original values, but only over \textit{masked} ocean regions. This encourages the model to infer spatial and temporal structure rather than memorize visible inputs. Prior work has shown that restricting loss computation to masked regions improves generalization by forcing the model to rely on contextual information \citep{he2022masked}. To prevent learning from irrelevant land data, we explicitly exclude these regions using a binary mask. This mask excludes land pixels from the loss computation. As a result, the model focuses exclusively on learning ocean dynamics. The ignored land tokens may serve as implicit registers, which could further enhance learning efficiency \citep{darcet2023vision}.

\subsection{Pretraining setup}
The setup of the pipeline, monitoring and working in the cloud are discussed in a separate blogpost \footnote{\url{https://coenvde.github.io/blogs/training-at-larger-scale/index.html}} related to this research. The full parameter overview in \autoref{appendix:training_config}.

Given that our data modality is climate instead of language, and the initial experiments did not show a performance gain for larger models, we selected a relatively compact model architecture compared to other foundation models. The features of interest in the marine domain, such as eddies, currents, and coastal gradients, are also more abstract and less structured than terrestrial features such as buildings or roads, further motivating a streamlined design to reduce overfitting and manage training cost. Architectural decisions were guided by domain experts, and our hyperparameter setup draws inspiration from the Clay project \citep{clay2025}, Aurora \citep{bodnar2024foundation}, Olmo \citep{herzog2025olmoearth}, Tessera \citep{feng2025tessera}, and Prithvi \citep{szwarcman2024prithvi}. As a practical rule of thumb, we set the embedding dimension to be no smaller than the number of input channels, ensuring the patch projection is non-lossy. This follows established practice, as discussed with a researcher from the Aurora team.

We process the world in chips. Each chip is divided into non-overlapping spatiotemporal patches (representation of pixels) of size $(P_t, P_h, P_w) = (1, 3, 3)$, yielding a token grid of $(12, 15, 15)$ per chip. For input, training chips of shape $(B, C, T, H, W)$ = $(B, 47, 12, 45, 45)$ were split into $N$ patches of size $(P_t, P_h, P_w) = (1, 3, 3)$. This meant that the patch grid was now: (time patches, height patches, width patches) = $(12, 15, 15)$. This patching scheme ensures that each chip contains a central patch, and every patch itself has a central pixel, which simplifies spatial alignment during inference. This patch size reduces total token count while preserving essential spatial detail. A random subset of patches was masked, and only the visible ones were passed through the encoder. The decoder received these encoded representations along with learned mask tokens and reconstructed the original input.

\section{Pretraining results}



\subsection{Reconstruction: Initial qualitative assessment}

\begin{figure}[H]
  \centering
  \begin{subfigure}{\textwidth}
    \centering
    \includegraphics[width=0.7\textwidth]{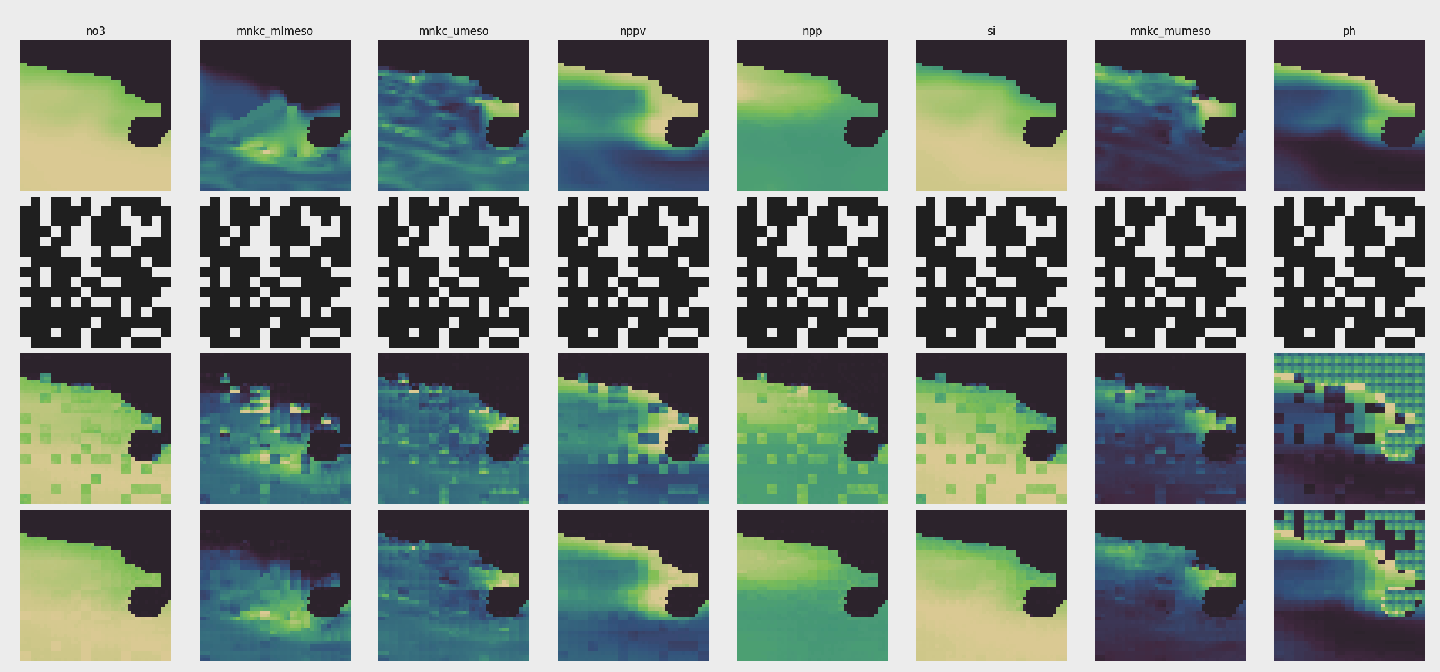}
    \caption{Random masking}
    \label{fig:train_reconstruction_random}
  \end{subfigure}

  \begin{subfigure}{\textwidth}
    \centering
    \includegraphics[width=0.7\textwidth]{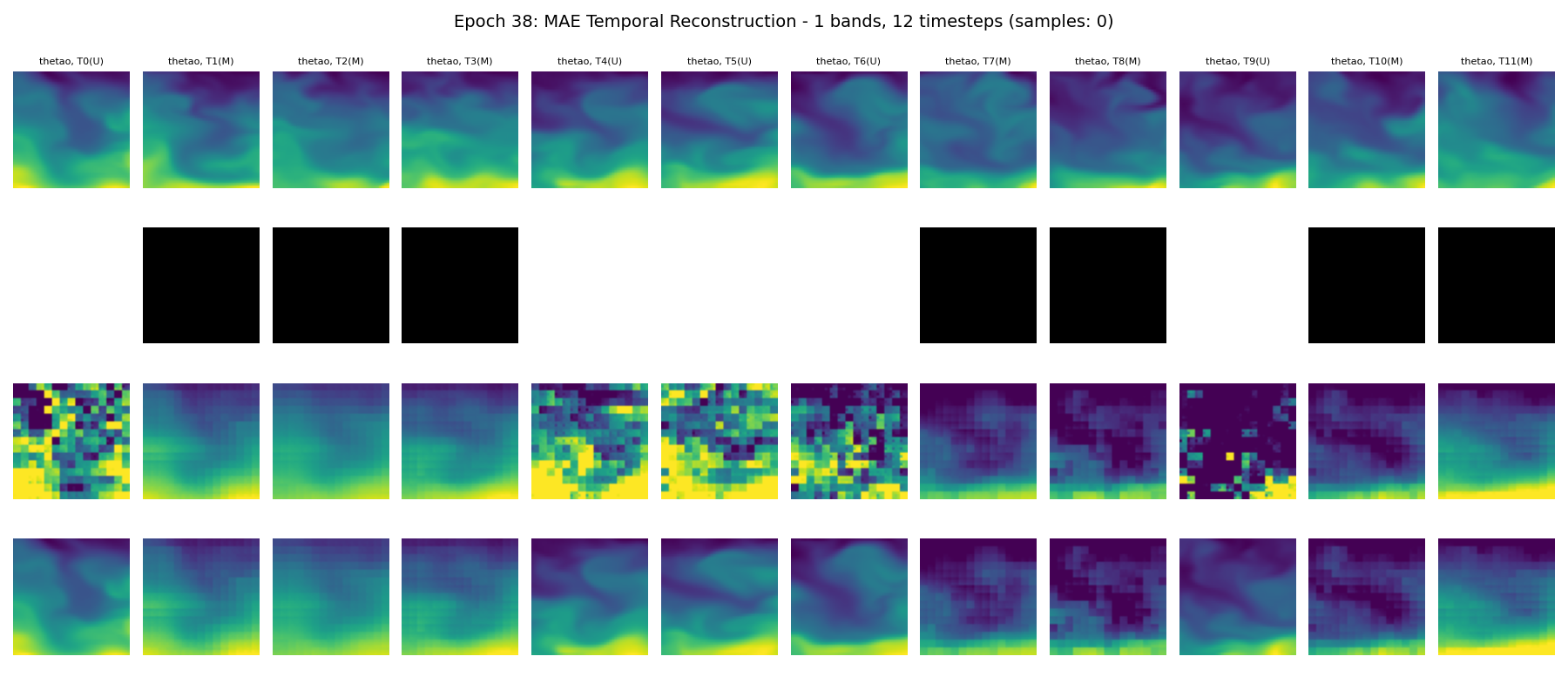}
    \caption{Temporal masking}
    \label{fig:train_reconstruction_spatial}
  \end{subfigure}

  \begin{subfigure}{\textwidth}
    \centering
    \includegraphics[width=0.7\textwidth]{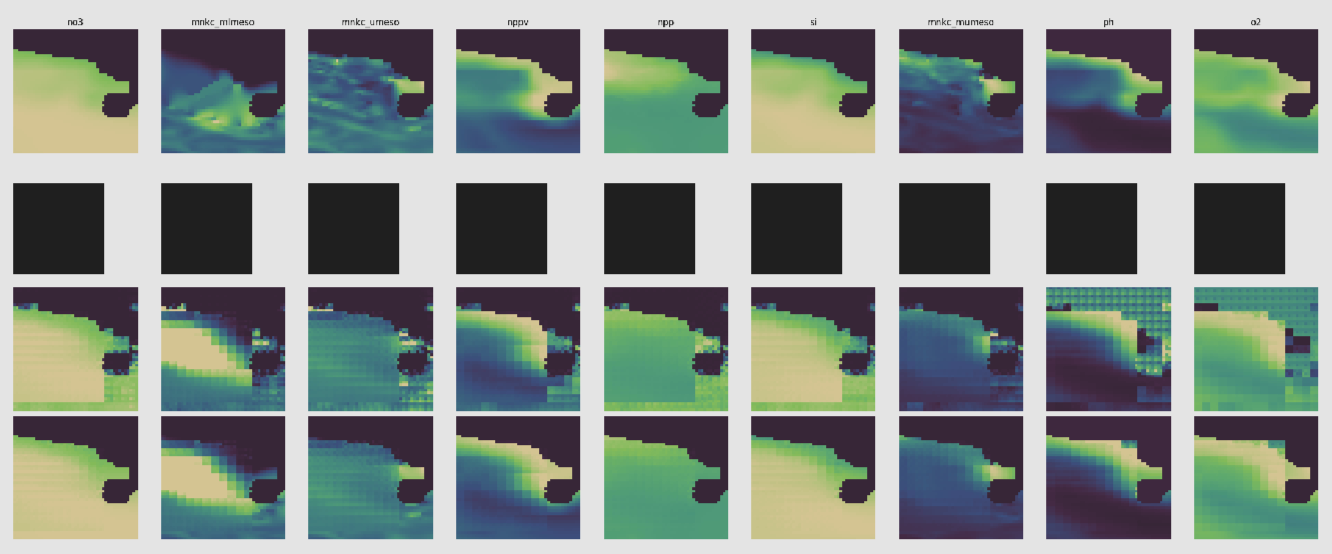}
    \caption{Spatial masking}
    \label{fig:train_reconstruction_temporal}
  \end{subfigure}

  \caption{Reconstruction of all variables in a single chip at the end of pretraining under different masking strategies. Each column represents a different variable. Rows: (1) original input, (2) mask applied during training (black = masked), (3) model prediction, (4) overlay of prediction and original values in unmasked regions.}
  \label{fig:train_reconstruction_all}
\end{figure}

As a first qualitative assessment of the model’s performance, we visualise reconstructions on the validation set during experimentation. \autoref{fig:train_reconstruction_random}, \autoref{fig:train_reconstruction_spatial}, and \autoref{fig:train_reconstruction_temporal} show the reconstruction of variables with each masking strategy in a single timestep of a chip from the validation set. They visually confirm the temporal and spatial abilities of (variable) reconstruction. The temporal reconstruction is visibly worse compared to the others, hypothetically because the task is harder and the granularity of the temporal dimension is too coarse to provide meaningful signal.

\begin{figure}[H]
  \centering
  \includegraphics[width=0.7\textwidth]{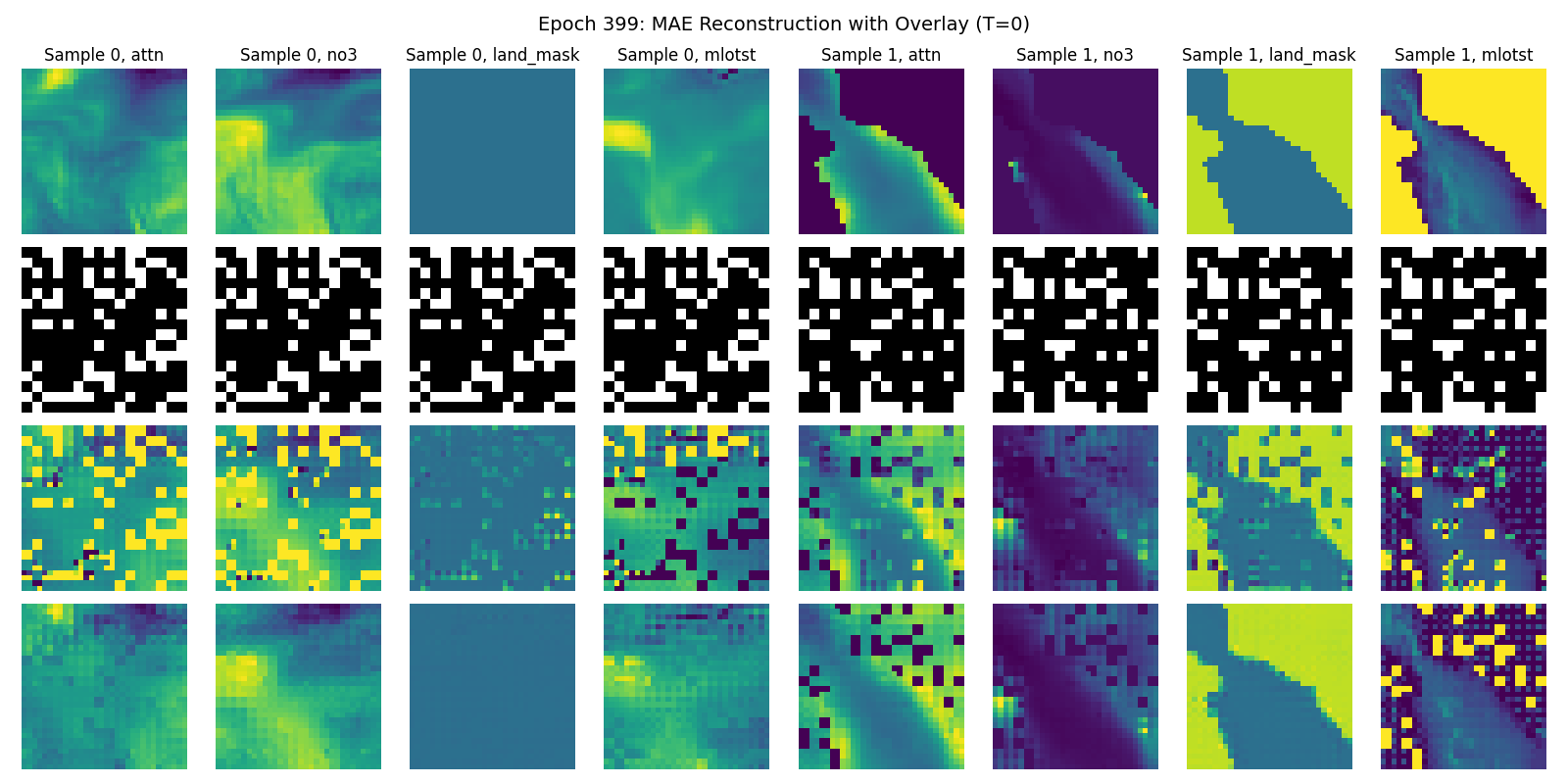}
  \caption{
    Reconstruction of a chip during experimentation. Each column shows a different variable; a clear example of density estimation is visible in the NO\textsubscript{3} channel on the land pixels. Rows: (1) original input, (2) mask applied during training (black = masked), (3) model prediction, (4) overlay of prediction and original values in unmasked regions.
  }
  \label{fig:density_chip}
\end{figure}

On the land pixels in \autoref{fig:density_chip}, the model appears to be learning a density function. Notably, in the prediction of NO\textsubscript{3}, the model does not simply replicate known values at isolated pixels. Instead, it produces smoothed predictions around these points. This behavior, observed consistently during both pretraining and experiment runs, suggests that the model may be capturing underlying processes rather than just filling in missing data.

\subsection{Exploring the Embedding Space}\label{sec:embedding_space}

To evaluate the representational quality of the DARWIN encoder, we analyze the distribution and structure of the latent embeddings produced from the validation set. Embeddings are extracted by running the encoder in inference mode with all masking ratios set to zero, sliding a $45 \times 45 \times 12$ chip at one-pixel stride across the global ocean and retaining the centre patch token at the final temporal step. This analysis includes both a quantitative statistical profile of the embedding space and a qualitative visualisation of feature clusters.

\subsubsection{Statistical Distribution and Hypersphere Characteristics}

\begin{figure}[H]
  \centering
  \begin{minipage}[c]{0.45\textwidth}
    \centering
    \captionof{table}{Scalar statistics of embeddings.}
    \label{tab:embedding_stats_scalar}
    \begin{tabular}{lcc}
    \toprule
    \textbf{Metric} & \textbf{Global} & \textbf{Europe} \\
    \midrule
    Minimum & -4.308 & -2.948 \\
    Maximum & 5.657 & 3.752 \\
    Mean ($\mu$) & -0.011 & -0.014 \\
    Std.\ Dev.\ ($\sigma$) & 0.427 & 0.433 \\
    \bottomrule
    \end{tabular}

    \vspace{1em}

    \captionof{table}{$L_2$ Norm statistics of embeddings.}
    \label{tab:embedding_stats_norm}
    \begin{tabular}{lcc}
    \toprule
    \textbf{Metric} & \textbf{Global} & \textbf{Europe} \\
    \midrule
    Mean Norm & 9.649 & 9.789 \\
    Min Norm & 6.866 & 8.949 \\
    Max Norm & 11.838 & 11.810 \\
    Std.\ Dev.\ Norm & 0.326 & 0.290 \\
    \bottomrule
    \end{tabular}
  \end{minipage}\hfill
  \begin{minipage}[c]{0.5\textwidth}
    \centering
    \includegraphics[width=\textwidth]{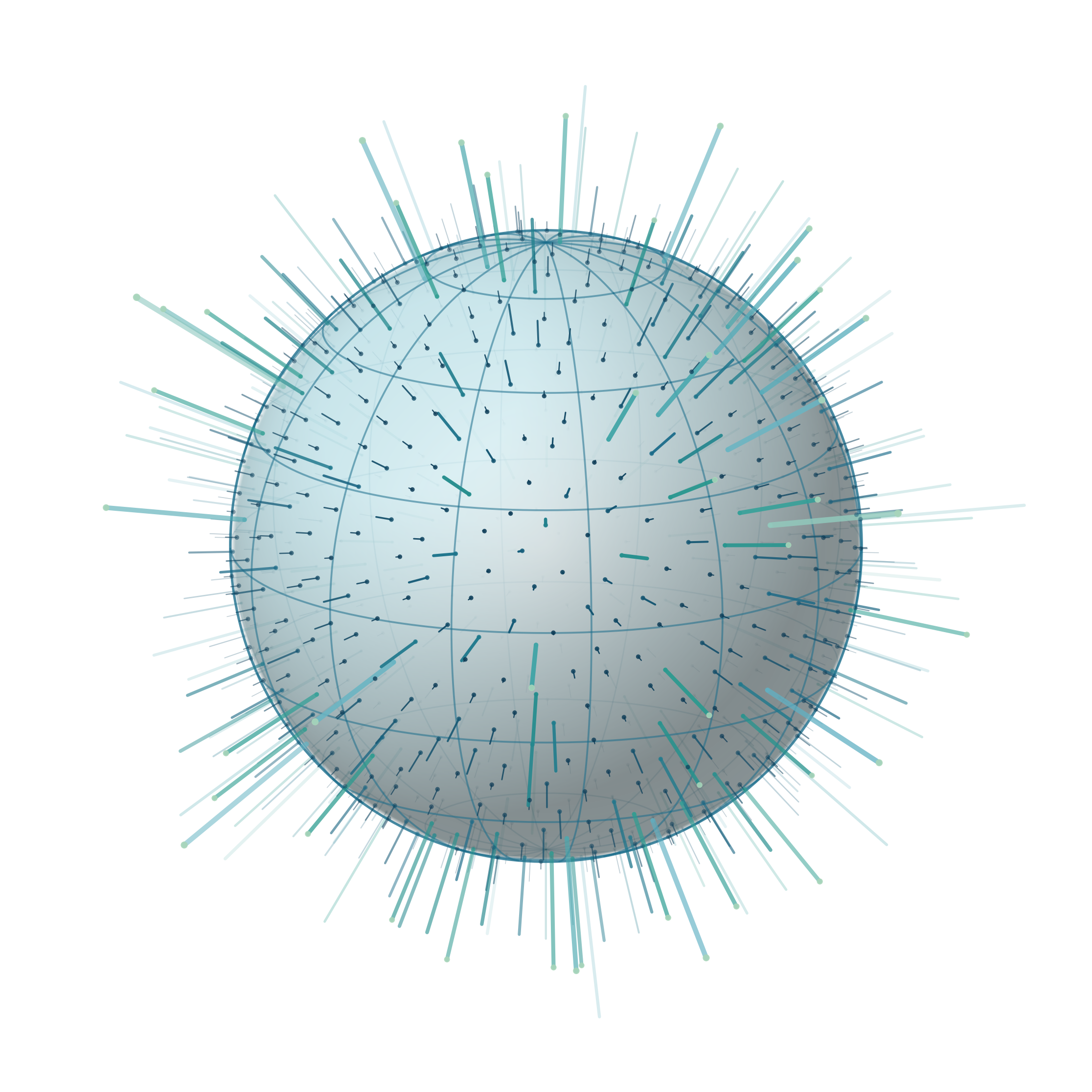}
    \caption{Conceptual visualisation of the embeddings distributed on a spiky surface of a high-dimensional hypersphere.}
    \label{fig:spiky_hypersphere}
  \end{minipage}
\end{figure}

The embeddings were computed at global scale for the target month of January 2020 (12-month input window: February 2019 - January 2020). We report statistics on the scalar embedding values and their per-token $L_2$ norms in Table~\ref{tab:embedding_stats_scalar} and Table~\ref{tab:embedding_stats_norm}, where the Europe column gives the same statistics restricted to the European marine subset of the global embedding map ($20^{\circ}$W--$15^{\circ}$E, $42^{\circ}$N--$65^{\circ}$N). Both views reveal a consistent and stable latent space.

The scalar values within the embeddings are tightly concentrated around zero, with a small standard deviation ($\sigma \approx 0.43$) but heavy tails reaching roughly $\pm 5$. The $L_2$ norms exhibit a low standard deviation ($\sigma \approx 0.3$) relative to a mean magnitude of approximately 10. This suggests that the model maps input data onto a spiky surface of a high-dimensional hypersphere. In such a manifold, the Euclidean distance is dominated by the angular separation between vectors. This indicates that the model learns to encode marine features primarily through the \textit{direction} (orientation) of the embedding vectors.

\subsubsection{Qualitative Visualisation via PCA}

\begin{figure}[H]
     \centering
     \begin{subfigure}[b]{0.54\textwidth}
         \centering
         \includegraphics[width=\textwidth]{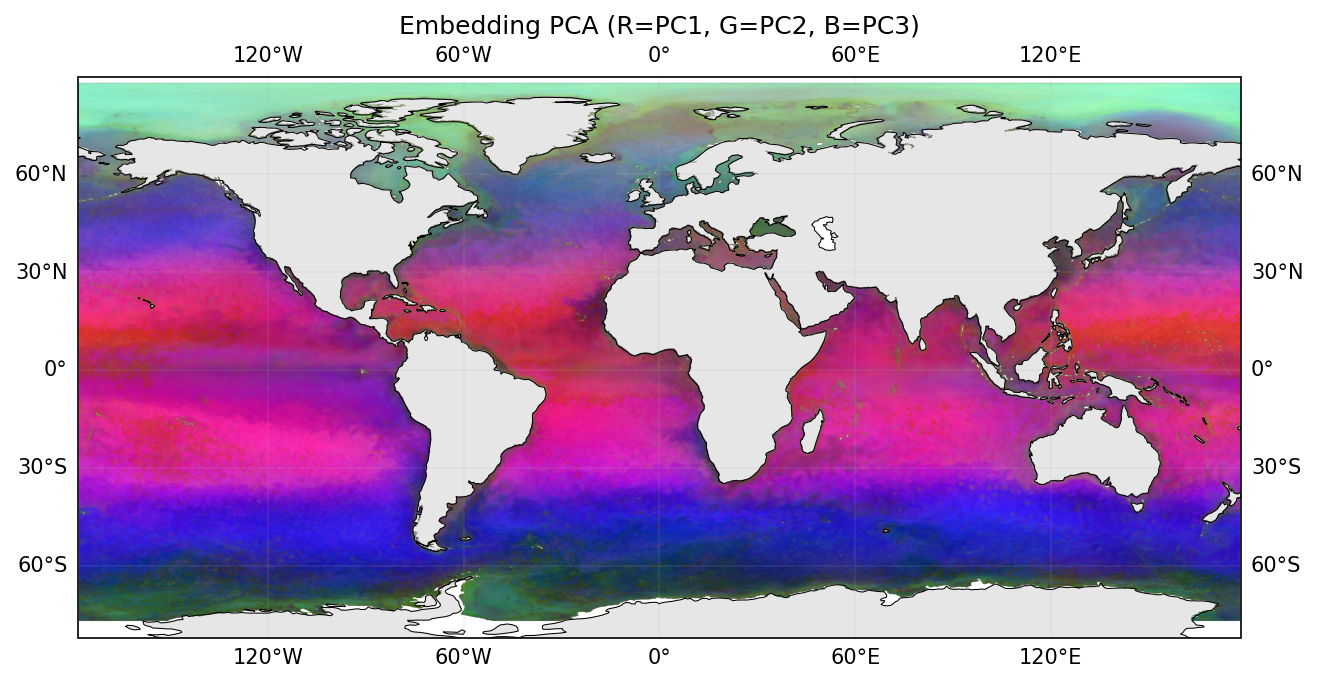}
         \caption{Global snapshot (January 2020)}
         \label{fig:pca_global}
     \end{subfigure}
     \hfill
     \begin{subfigure}[b]{0.45 \textwidth}
         \centering
         \includegraphics[width=\textwidth]{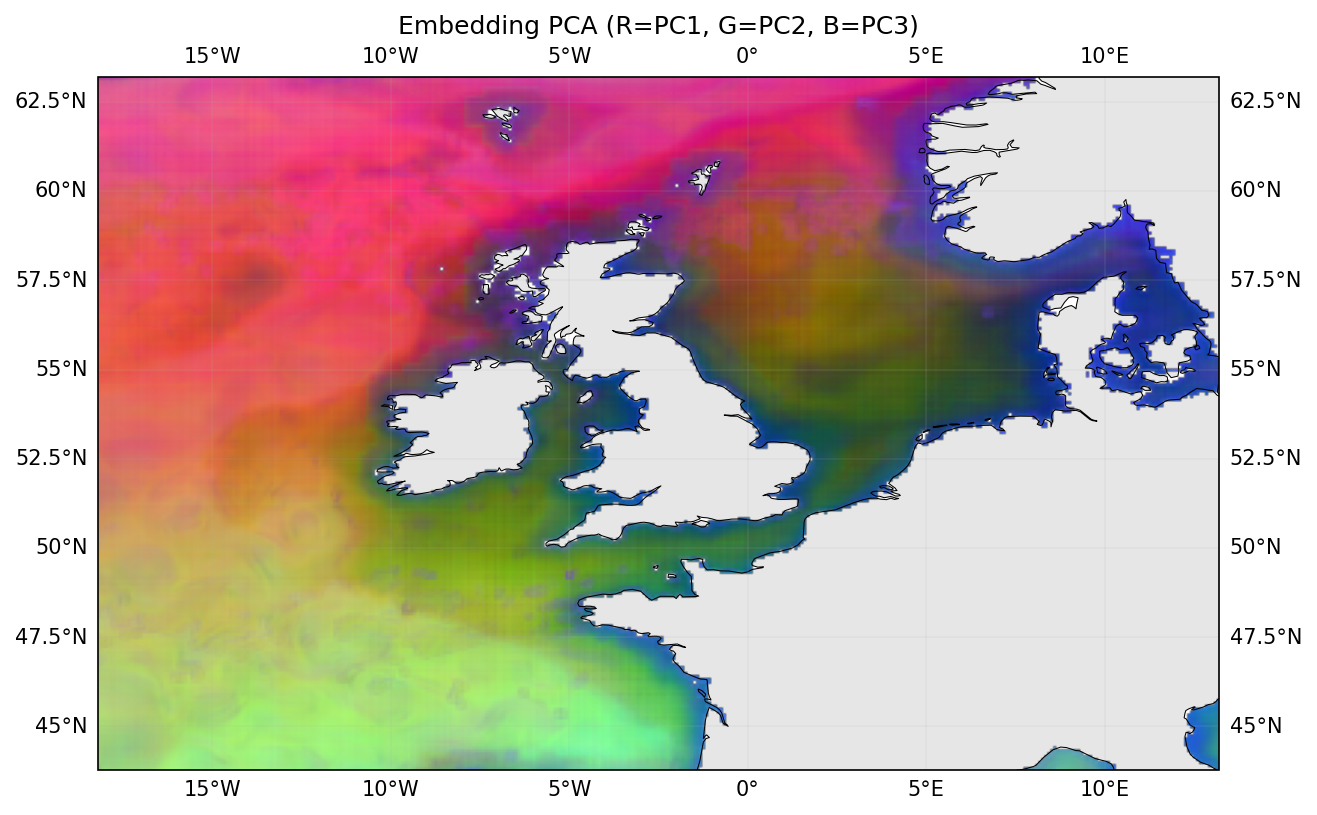}
         \caption{European snapshot (January 2020)}
         \label{fig:pca_europe}
     \end{subfigure}
     \caption{RGB visualisation of the top three principal components of DARWIN embeddings. Spatial coherence in the colors indicates that the model has learned to represent distinct oceanographic regimes across different scales. The PCA is run on the individual snapshot, so colors do not correspond across plots.}
     \label{fig:PCA_combined}
\end{figure}

To visualise the high-dimensional relationships captured by the model, we perform Principal Component Analysis (PCA) on the embeddings. \autoref{fig:PCA_combined} illustrates the first three principal components mapped to RGB channels.

The resulting visualisation shows clear, coherent spatial patterns that correspond to oceanographic structures.

\section{Validating DARWIN - Downstream Task}

\subsection{Methodology}

Representation-learning losses are not a good metric for evaluating model and embedding quality. For this reason, (standardised) downstream tasks are commonly used to benchmark the performance of foundation models. In contrast to terrestrial methods, there are no standardised benchmarks tasks readily available for the ocean yet. To evaluate DARWIN's practical utility, we select chlorophyll-\textit{a} prediction as the primary downstream task. This choice is motivated by three factors: chlorophyll-\textit{a} is directly observable from space via Sentinel-3 OLCI, providing reliable labels at scale; it is also directly related to the oceanographic processes that DARWIN is designed to represent; and it serves as a proxy for data-scarce tasks of high real-world relevance, such as harmful algal bloom detection and prediction.

For chlorophyll-\textit{a} fine-tuning, we use a separate DARWIN encoder that was \emph{trained from scratch} on the same global Copernicus Marine reanalysis stack as the full model, but with all ocean-colour, plankton, and biomass-related dynamic channels removed from the self-supervised objective. This design deliberately withholds direct biomass, color, and pigment information from the backbone during foundation-model training, so that the downstream chlorophyll task cannot rely on trivial colour-channel leakage. We did retain SPM and KD490 variables even though they are highly correlated to chlorophyll because we argue they represent both geophysical (run-off, turbidity) as well as biochemical processes (nutrient concentration, algal blooms). Predicting chlorophyll-*a* concentration without this information would be a more challenging task for the model and may lead to some model candidates not besting a climatological baseline (historic month-averages, resulting in MAE of 1.06 for validation year 2022, details not included here). This approach is also more similar to the approach taken by \citep{dawson2025sentinel} who use Sentinel 3 OLCI data as input features to model in-situ chlorophyll-*a* concentration.  \autoref{tab:included_static}, \autoref{tab:included_dynamic}, and \autoref{tab:omitted_dynamic} in \autoref{appendix:variables} detail the exact subset of variables included and omitted during this specific pretraining run.

\paragraph{Sliding-window inference.} To produce dense chlorophyll predictions across the global ocean, we tile each output month with the same spatially non-overlapping $45 \times 45$ chips used during training (spatial stride equal to the chip size; temporal stride of one month). Every ocean pixel is therefore predicted exactly once. An alternative is fine-grained, pixel-stride sliding inference, in which the encoder is evaluated at every one-pixel shift where only the centre pixel is saved; this is the scheme we use to extract the embeddings analysed in Section~\ref*{sec:embedding_space}. Pixel-stride inference hypothetically yields better predictions, as each pixel benefits from having maximal context, instead of being at the border, but at roughly $45^2 \approx 2{,}000\times$ the computational cost of chip-stride inference, which was a limiting factor. The chip-stride scheme keeps test-time evaluation tractable.

\subsubsection{Data}\label{sec:methodology_data}

We utilize the \textit{ESA Ocean Colour Climate Change Initiative (CCI) v6.0} dataset for downstream evaluation. Specifically, we use the L3 mapped monthly chlorophyll-\textit{a} product at $4\,\mathrm{km}$, spatial resolution. The training and validation sets comprise observations from 1998 - 2021, employing the same spatiotemporal holdout regions defined during DARWIN’s pretraining. The calendar year 2022 is reserved exclusively as an independent test set.

The decision to prioritize L3 over L4 data is based on two factors. First, L3 values are directly traceable to observed ocean states; using L4 data would risk benchmarking DARWIN against the assumptions of a specific interpolation algorithm rather than actual observations. Second, the monthly composition of L3 data significantly reduces the "sparsity" typical of daily satellite swaths, providing sufficient global coverage for robust training.

We acknowledge that L3 monthly composites are not spatially uniform. Persistently cloudy regions, such as the Southern Ocean or high latitudes during winter, remain under-sampled. Consequently, performance metrics should be interpreted with the understanding that they represent fewer independent observations. The temporal and spatial splitting policy matches the pretraining split (\autoref{fig:data_splits}).

\paragraph{Standardization.} Chlorophyll-\textit{a} is heavy-tailed and nonnegative. For every valid pixel we apply a \textit{log1p} transform $t = \log(1+x)$ with $x \ge 0$ and $x$ finite; invalid or missing pixels are excluded from the loss. Training and validation metrics in ``loss space'' use a \textbf{global} $z$-score of $t$: we compute the mean $\mu_{\log}$ and standard deviation $\sigma_{\log}$ of $t$ over \emph{all} valid pixels in the merged aligned chlorophyll store (full training corpus), then standardise as $(t - \mu_{\log})/\sigma_{\log}$, with $\sigma_{\log}$ floored to a small constant for numerical stability. Reported physical-space MAE/RMSE apply the inverse map $\mathrm{expm1}(z\,\sigma_{\log} + \mu_{\log})$ to predictions and labels.

\begin{figure}[H]
  \centering
  \includegraphics[width=\textwidth]{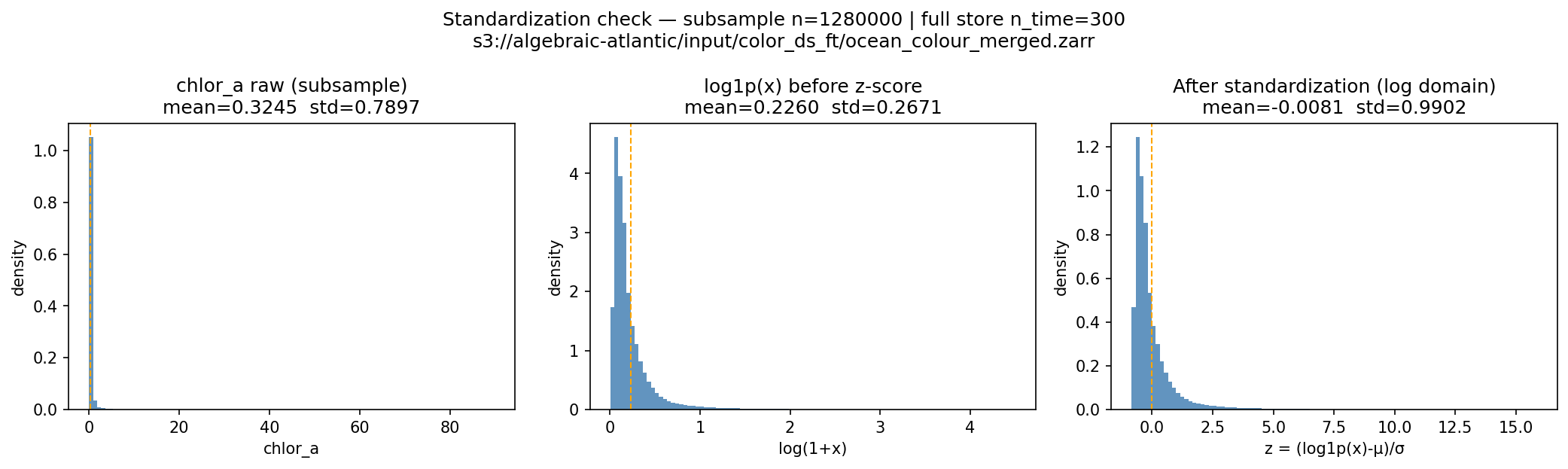}
  \caption{Histograms illustrating the target standardization process for chlorophyll-\textit{a}. The original heavy-tailed distribution (left) is normalized using a $\log(1+x)$ transformation (centre) and subsequently standardised using a global $z$-score (right) to compute training and validation metrics in loss space.}
  \label{fig:oc+chl_stats}
\end{figure}

\paragraph{Data-efficiency subsampling.} To assess how DARWIN scales with the amount of supervised data, we run experiments in which the training set is subsampled while the validation and test sets remain fixed. For a target training fraction $f \in (0, 1]$ and $n$ available training chips, we draw a single random permutation of the index set $\{0,\ldots,n-1\}$ with a fixed seed of $42$, and retain the first $k = \max\!\bigl(1,\,\mathrm{round}(f \cdot n)\bigr)$ entries of that permutation as the training subset. Because every fraction uses a \emph{prefix} of the same permutation, the resulting subsets are nested across fractions, i.e.\ $\mathcal{D}_{0.1} \subset \mathcal{D}_{0.2} \subset \cdots \subset \mathcal{D}_{1.0}$. Increasing the fraction therefore only adds chips to the previous subset rather than swapping them, isolating the effect of \emph{more data} from \emph{different data} in the resulting label-efficiency curves.

\begin{table}[H]
\centering
\caption{Chlorophyll downstream sample counts (number of $45{\times}45$, 12-timestep chip windows).}
\label{tab:downstream_sample_counts}
\begin{tabular}{lrrr}
\toprule
\textbf{Train fraction} & \textbf{Train} & \textbf{Val} & \textbf{Test} \\
\midrule
$1.0$ (100\,\%) & \num{871165} & \num{70635} & \num{40800} \\
$0.1$ (10\,\%)  & \num{87116}  & \num{70635} & \num{40800} \\
$0.01$ (1\,\%)  & \num{8712}   & \num{70635} & \num{40800} \\
$0.001$ (0.1\,\%) & \num{871}  & \num{70635} & \num{40800} \\
\bottomrule
\end{tabular}
\end{table}

Details regarding the downstream data alignment, preprocessing, and target standardization can be found in \autoref{sec:downstream_preprocessing}.

\subsubsection{Setup}\label{sec:training_eval_protocol}

To evaluate the efficacy of the foundation model, we design a series of comparative experiments that systematically vary baselines, prediction granularity, encoder initialisation, and head architecture.

\paragraph{Baselines.}
We evaluate two families of baselines that do not use a foundation-model encoder:
\begin{enumerate}[nosep]
    \item \textbf{Deep-learning baselines.} Lightweight models that map the raw input variables directly to a chlorophyll prediction, without an encoder. We ablate these across prediction granularity (pixel vs.\ patch), inclusion of static inputs (with vs.\ without bathymetry, and head architectures (Linear, MLP, and single-layer Attention). This grid of configurations isolates the contribution of spatial context sharing (patch vs.\ pixel), geographic prior information (static inputs), and model expressivity (head complexity).
    \item \textbf{Tree-based baselines.} We train XGBoost and LightGBM tree regressors as strong non-neural baselines, using the same predictor set as DARWIN-NoColour: 26 dynamic reanalysis variables and five static GEBCO channels, with all ocean-colour channels  withheld to prevent trivial leakage. Further details can be found in \autoref{appendix:tree_reg}.
\end{enumerate}

\paragraph{Prediction granularity.}
The foundation-model fine-tuning task is framed as a dense patch prediction problem: the model predicts a $3 \times 3$ pixel patch for every embedding token. For the deep-learning baselines, we additionally ablate pixel-level prediction (one output per input pixel) to quantify the benefit of spatial context sharing through patch-level processing.

\paragraph{Encoder initialization.}
To quantify the benefit of self-supervised pretraining, every encoder-based configuration is trained twice: once with DARWIN-NoColour's pretrained weights (which can intuitively be seen as an informed prior for downstream applications as weight initialization), and once from a random initialization of an architecturally identical model. Both variants are evaluated with the backbone either frozen (only the head is trained) or unfrozen (end-to-end fine-tuning).

\paragraph{Head architectures.}
We compare four prediction heads on top of the encoder, ordered by expressivity:
\begin{enumerate}[nosep]
    \item[(i)] \textbf{Linear probe:} a single linear layer mapping each token embedding to the target patch.
    \item[(ii)] \textbf{MLP:} a multi-layer perceptron with non-linear activations.
    \item[(iii)] \textbf{Decoder head (added encodings):} a single-layer transformer decoder that re-injects absolute latitude/longitude embeddings, patch-scale embeddings, absolute temporal and yearly signals, and the static bathymetry/mask---mirroring the structure of the pretraining decoder.
    \item[(iv)] \textbf{Vanilla attention head:} a single-layer transformer that adds only a fixed 3D sinusoidal grid positional embedding to the patch tokens.
\end{enumerate}
The key difference between (iii) and (iv) is an explicit \textbf{inductive bias}: the decoder head can condition predictions directly on static, calendar-time, and absolute spatial and scale information at every forward pass, whereas the vanilla attention head withholds that shortcut.

\paragraph{Loss calculation.}
All models are trained with mean squared error (MSE) computed exclusively over valid ocean pixels; pixels with missing values (NaNs) are excluded from the loss.

\paragraph{Evaluation.}
All models are evaluated on the full 2022 test year. To evaluate the predictive accuracy of the models, we calculate performance metrics across all valid pixels. To assess real-world performance, we compute the physical-space Mean Absolute Error (MAE) in milligrams per cubic meter (mg\,m$^{-3}$) after applying the inverse transformation $\mathrm{expm1}(z\,\sigma_{\log} + \mu_{\log})$ to both predictions and ground truth labels.

\subsection{Downstream Task Results}\label{sec:finetune-results}

\subsubsection{Quantitative analysis}

We observe that the bigger, more expressive configurations (particularly the unfrozen encoder with attention-based heads) tend to overfit rapidly. To account for this, we always report metrics from the best validation checkpoint rather than the final training step.

\begin{figure}[H]
  \centering
  \includegraphics[width=\textwidth]{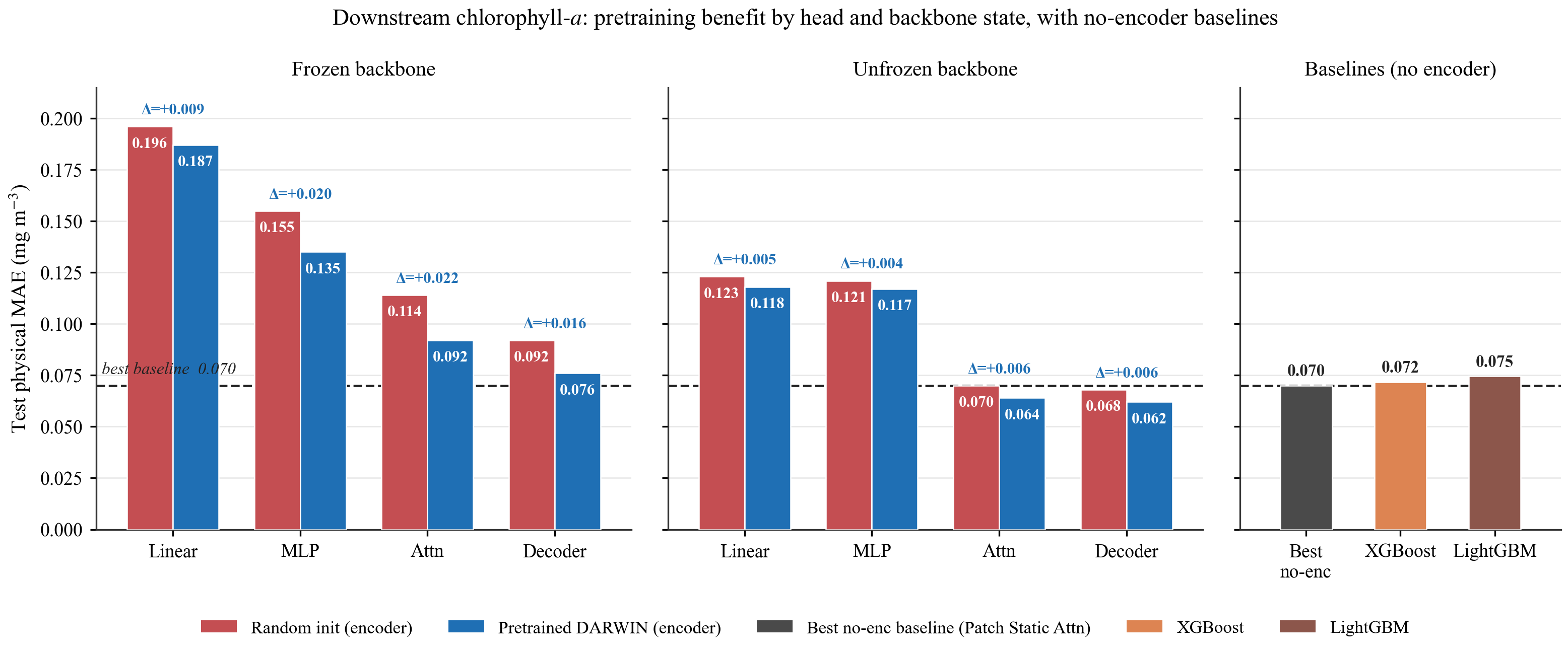}
  \caption{Test physical MAE on the held-out 2022 chlorophyll-\textit{a} test set across the four prediction heads, split by backbone state (frozen vs.\ unfrozen) and encoder initialisation (random vs.\ pretrained DARWIN). The dashed line marks the best no-encoder baseline (Patch Static Attn, $0.070$ mg\,m$^{-3}$). The $\Delta$ above each pair is the pretraining benefit (random $-$ DARWIN).}
  \label{fig:downstream_test_mae}
\end{figure}

Across all configurations, prediction heads follow a consistent ordering by expressivity: Linear $<$ MLP $<$ Vanilla Attention $<$ Decoder head. Each step up in non-linearity and parameter count yields a clear improvement, and the decoder head's re-injection of positional and static encodings provides a further marginal gain over the vanilla attention head. This ordering suggests that static and positional information has not leaked into the encoder's patch embeddings during pretraining (the decoder head's advantage comes precisely from re-supplying that context at prediction time). This property is desirable for downstream applications where spatial generalization, rather than spatial memorization, is the goal. The tree-based baselines (XGBoost at $0.072$~mg\,m$^{-3}$, LightGBM at $0.075$~mg\,m$^{-3}$) perform competitively with, but not better than the best no-encoder deep-learning baseline ($0.070$~mg\,m$^{-3}$), which benefits from spatial context sharing across neighbouring patches within each chip. The full baseline ablation across pixel/patch granularity and static inputs is shown in Appendix~\autoref{fig:downstream_baseline_ablation}. Finally, we evaluated the full pretrained foundation model, which attained the strongest downstream performance of all configurations, consistent with a larger and better pretraining dataset. The corresponding comparison is shown in Appendix~\autoref{fig:full_darwin_comparison}.

\begin{figure}[H]
  \centering
  \includegraphics[width=0.95\textwidth]{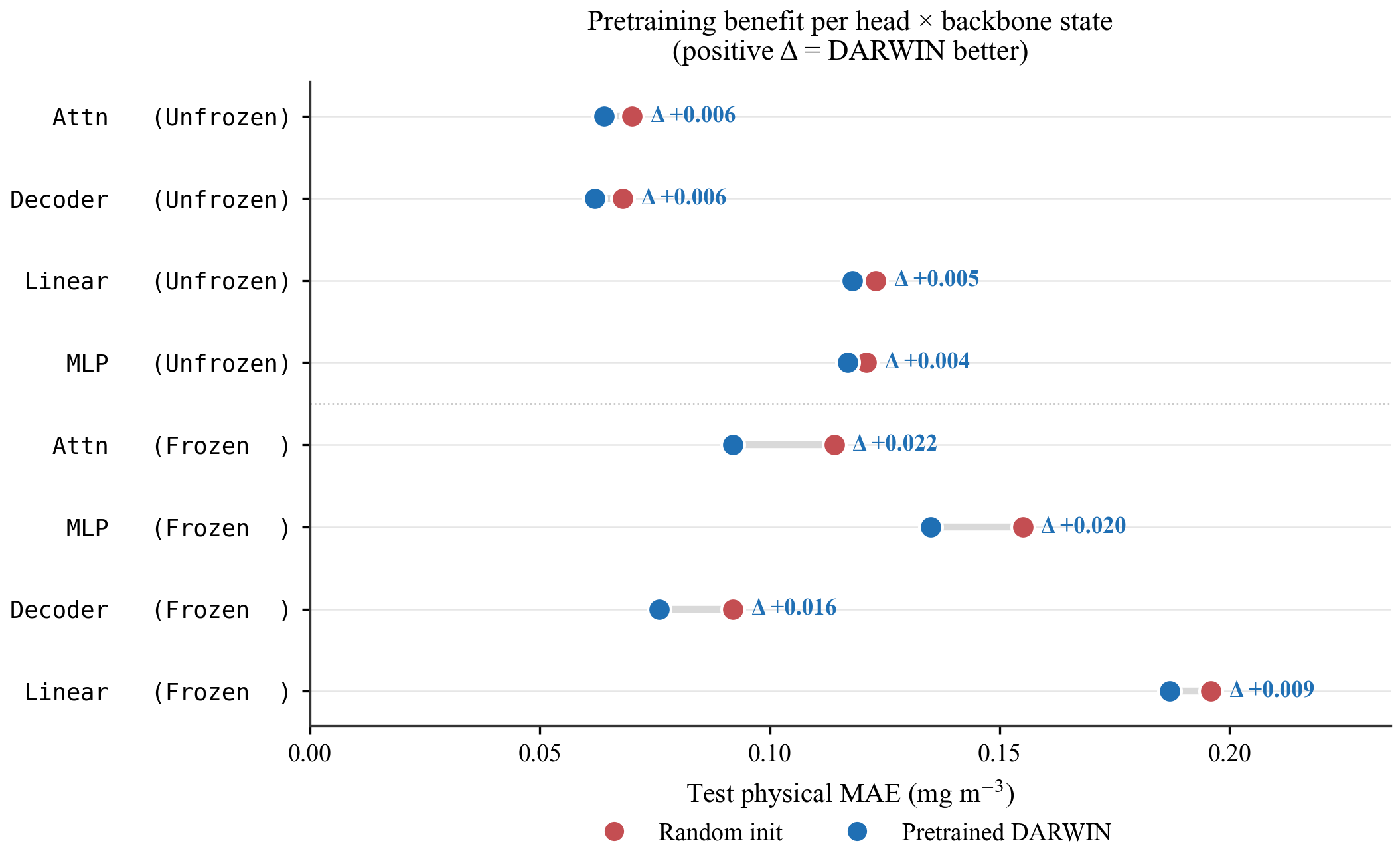}
  \caption{Pretraining benefit per (head $\times$ backbone state) configuration, sorted by $\Delta$ within each backbone group. Each row pairs the random-init and pretrained-DARWIN test MAE for the same architecture; the gap between the dots equals the gain attributable to pretraining alone.}
  \label{fig:downstream_gain_dumbbell}
\end{figure}

Comparing encoder initialisations, DARWIN's pretrained weights consistently outperform random initialisation across every head and backbone configuration. The benefit is most pronounced in the frozen regime, where the encoder cannot adapt to the downstream task and the quality of the pretrained representations is tested more directly. When the backbone is unfrozen, end-to-end fine-tuning narrows the gap (particularly for the more expressive decoder and attention heads, where both initialisations converge to similar performance) but pretrained DARWIN still retains a consistent edge. The overall best results are achieved by the pretrained DARWIN encoder, unfrozen, with the decoder head ($0.062$~mg\,m$^{-3}$ test MAE). Notably, the large performance gap between the frozen linear probe and the frozen MLP head suggests that the encoder's learned representations encode chlorophyll-relevant information in a non-linear manifold: a simple linear mapping cannot decode it, but a shallow non-linear head recovers much of the signal.

For the deep-learning baselines, incorporating static bathymetry inputs yields a consistent performance improvement across all configurations. We also find that patch-level baselines systematically outperform pixel-level baselines, supporting the hypothesis that spatial context sharing across neighbouring pixels is helpful for predicting chlorophyll concentration. A complete ablation of these baseline choices is presented in \autoref{fig:downstream_baseline_ablation}.

\subsubsection{Data-efficiency comparison}
\label{sec:data_efficiency_tables}

To assess label efficiency, we select the best-performing configurations from the full-data experiments and re-train them at reduced training fractions ($0.1\%$, $1\%$, $10\%$, and $100\%$). Specifically, we compare: (1)~the pretrained DARWIN encoder (unfrozen) with the decoder head, (2)~the randomly initialised encoder (unfrozen) with the decoder head, and (3)~the best no-encoder baselines (Patch Static Attention and Patch Static MLP). These configurations were chosen because they were the most performant modelling approaches. The MLP baseline is included as an additional reference point. Tree-based methods are excluded from the data-efficiency comparison as they cannot scale to the big data regimes used.  As in the full-data setting, we report metrics from the best validation checkpoint for each configuration, since models with larger capacity tend to overfit quickly at smaller data fractions.

\begin{figure}[H]
  \centering
  \includegraphics[width=\textwidth]{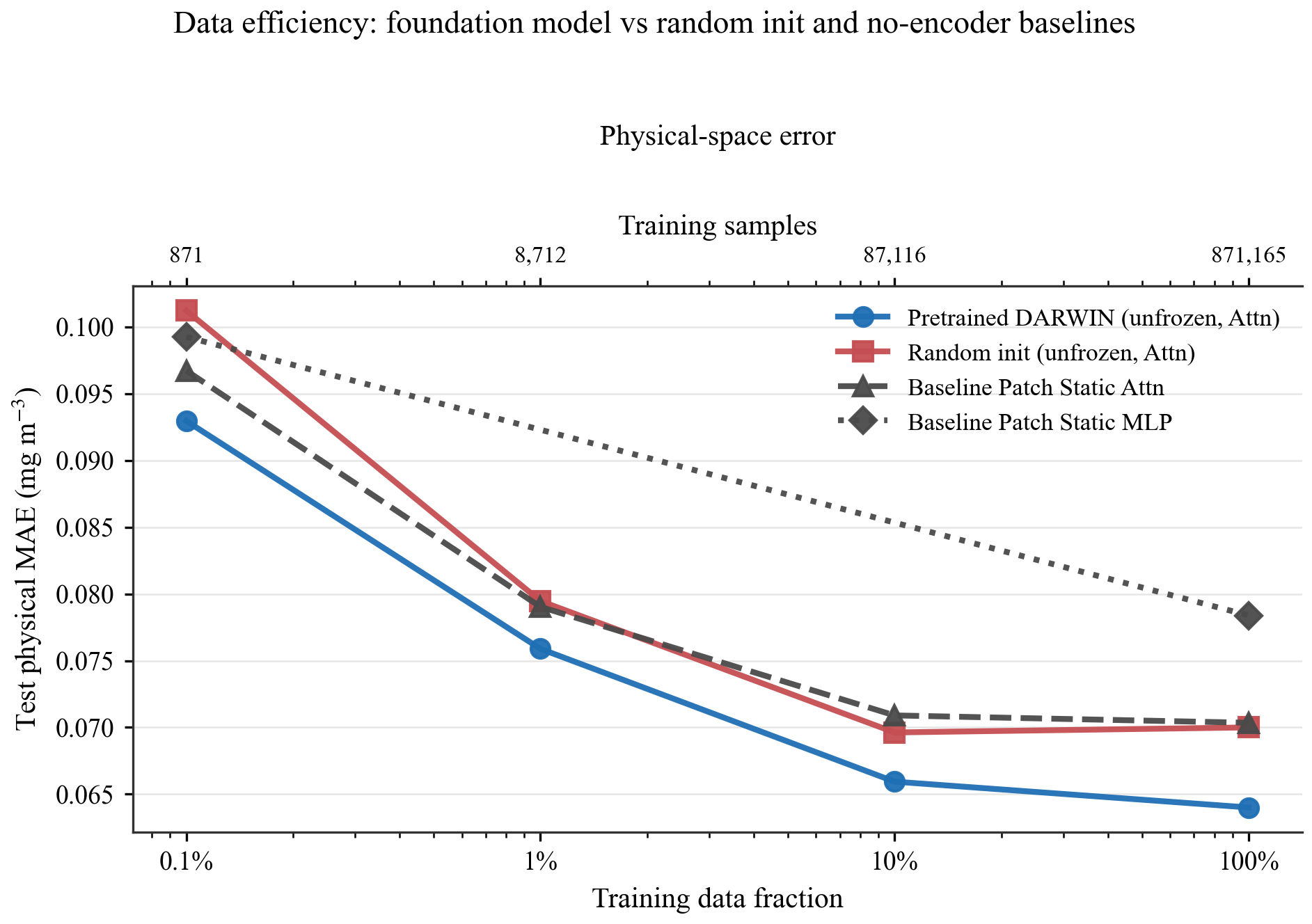}
  \caption{Data efficiency across training-set fractions for the attention head, evaluated on the 2022 chlorophyll-\textit{a} test holdout. The top axis on each panel reports the absolute number of $45{\times}45{\times}12$ training chips. The Patch Static MLP baseline is shown only at the two endpoints ($0.1\%$ and $100\%$) for reference.}
  \label{fig:downstream_data_efficiency}
\end{figure}

As illustrated in \autoref{fig:downstream_data_efficiency}, the pretrained DARWIN configuration consistently outperforms the randomly initialized encoder and the deep-learning baselines across all data scales. The label-efficiency benefits of self-supervised pretraining are substantial: DARWIN achieves a test MAE of $0.066$~mg\,m$^{-3}$ using only $10\%$ of the training data, surpassing the full-data performance of the randomly initialized model ($0.070$~mg\,m$^{-3}$ at $100\%$). Remarkably, even in the extreme low-data regime ($0.1\%$, corresponding to just 871 training samples), DARWIN maintains predictive accuracy and perform better than all compared models (even low parameter MLP). This suggests that the larger models have a simplicity bias in the optimization process.

\subsection{Qualitative analysis}\label{sec:downstream_qualitative}

We restrict the qualitative analysis to the three best-performing configurations: the \textit{Attention Baseline} (no foundation-model backbone), the \textit{Random Init} backbone fine-tuned unfrozen (with decoder head), and the \textit{Pretrained DARWIN} backbone fine-tuned unfrozen (with decoder head).

All three models reproduce the broad global chlorophyll-\textit{a} distribution for February 2022 (Figure~\ref{fig:qual_pred_log_feb}; linear-scale maps in Appendix Figure~\ref{fig:qual_pred_feb}).

\begin{figure}[H]
  \centering
  \includegraphics[width=0.95\textwidth]{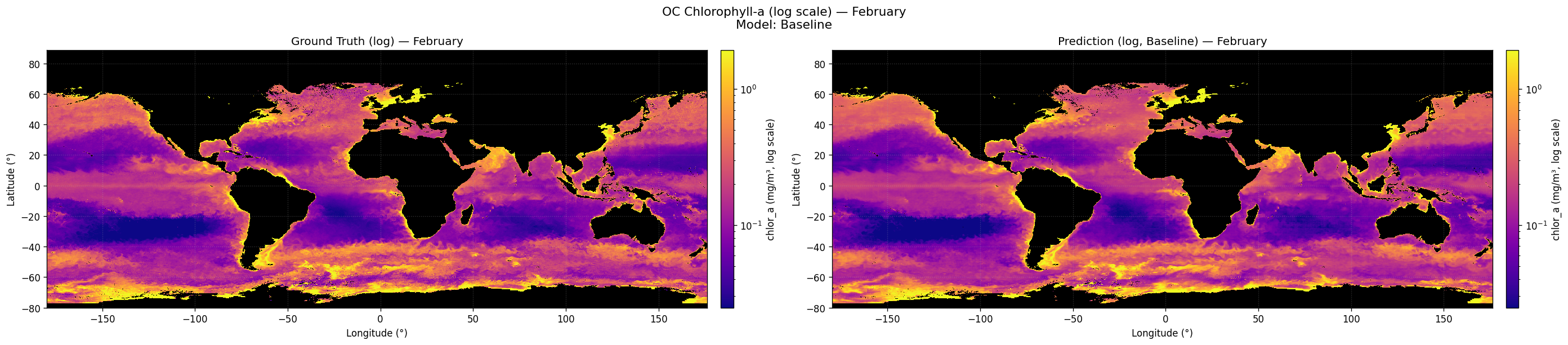}\\[0.3em]
  \includegraphics[width=0.95\textwidth]{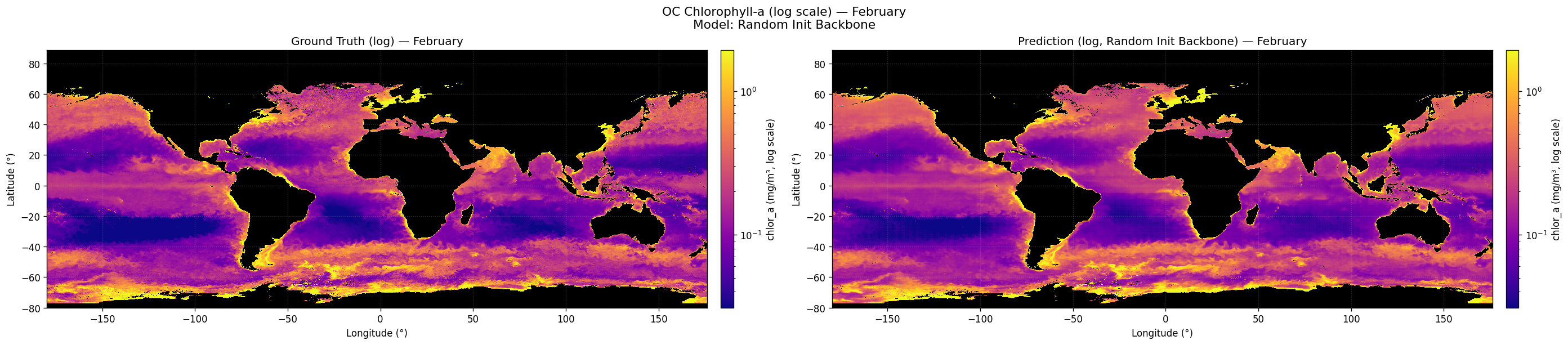}\\[0.3em]
  \includegraphics[width=0.95\textwidth]{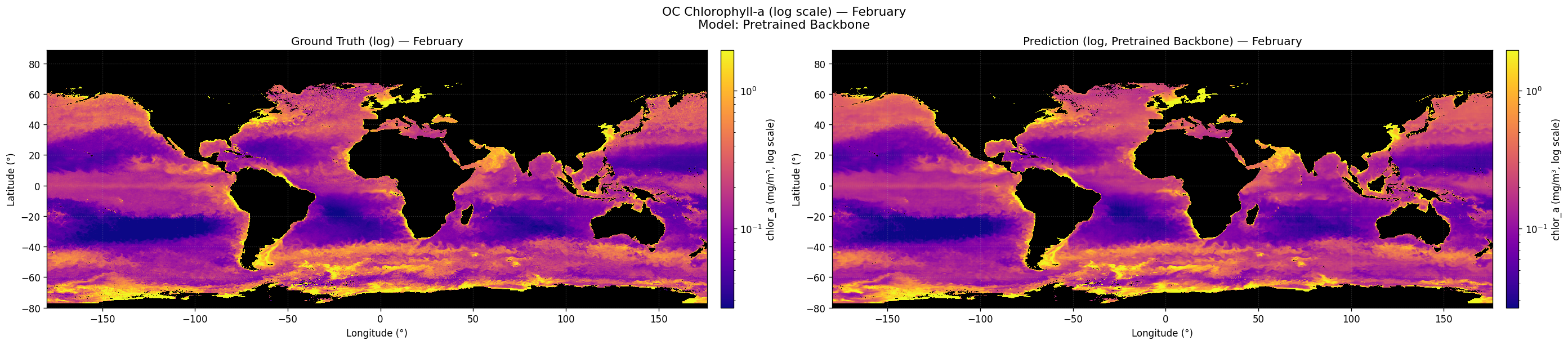}
  \caption{Predicted vs.\ observed chlorophyll-\textit{a} ($\log_{10}$ scale) for February 2022.
          Rows: Baseline, Random Init, Pretrained DARWIN.}
  \label{fig:qual_pred_log_feb}
\end{figure}

\begin{figure}[H]
  \centering
  \begin{tabular}{@{}cc@{}}
    \includegraphics[width=0.485\textwidth]{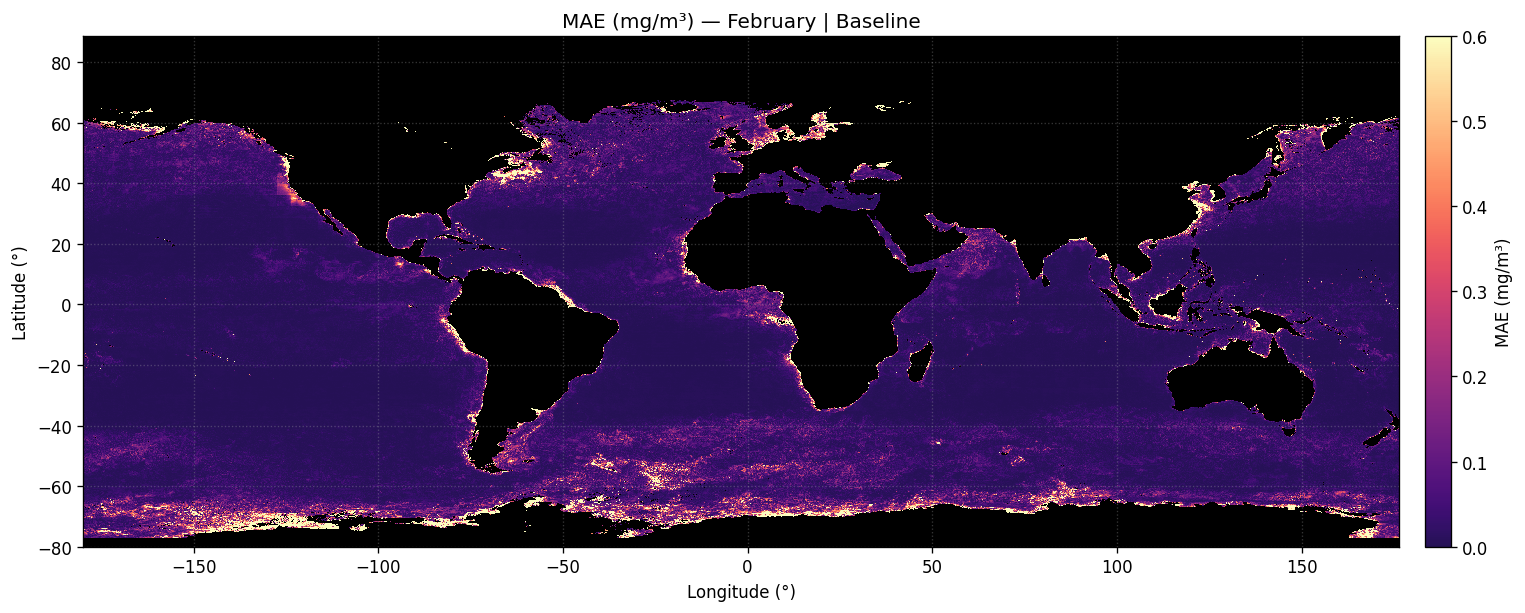} &
    \includegraphics[width=0.485\textwidth]{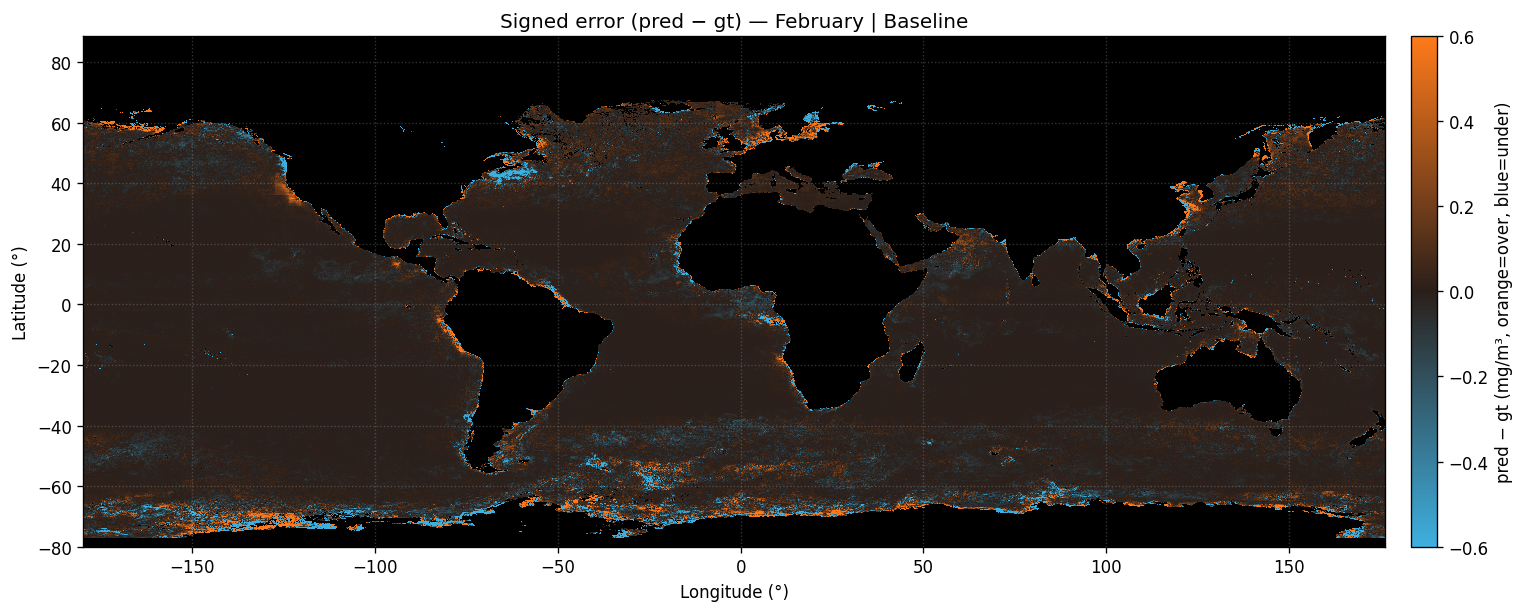} \\[0.2em]
    \includegraphics[width=0.485\textwidth]{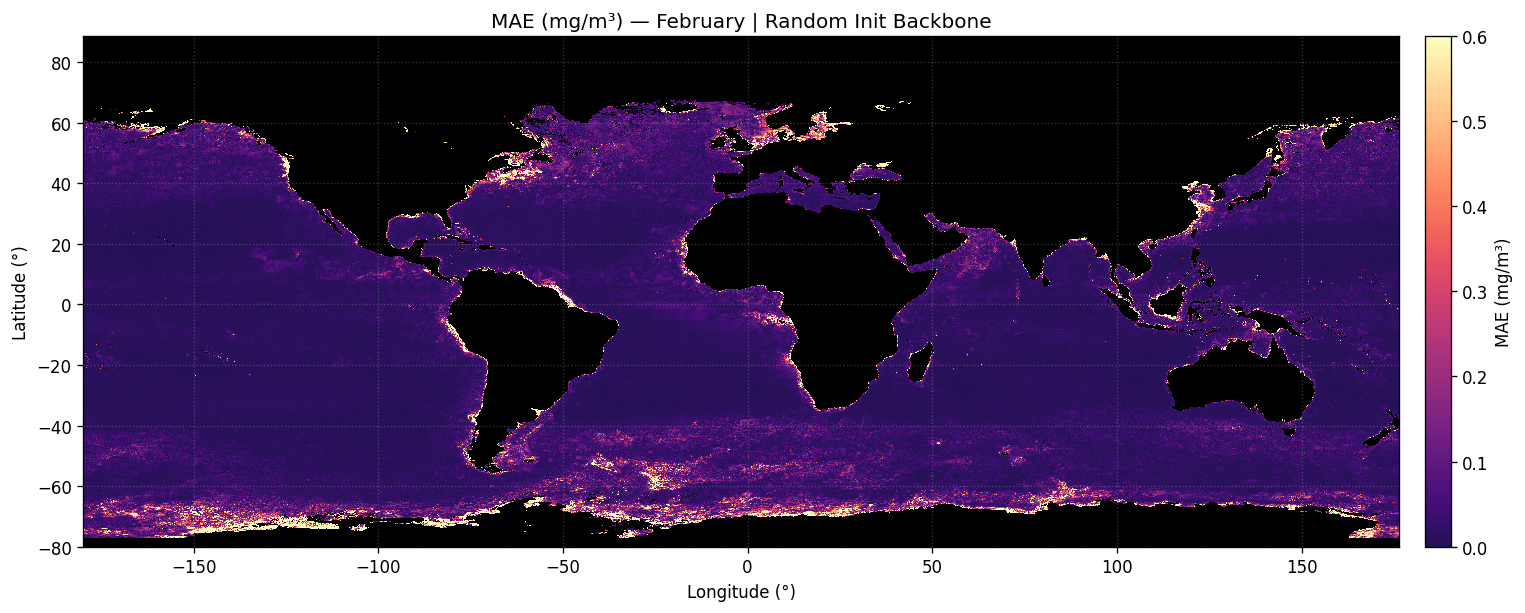} &
    \includegraphics[width=0.485\textwidth]{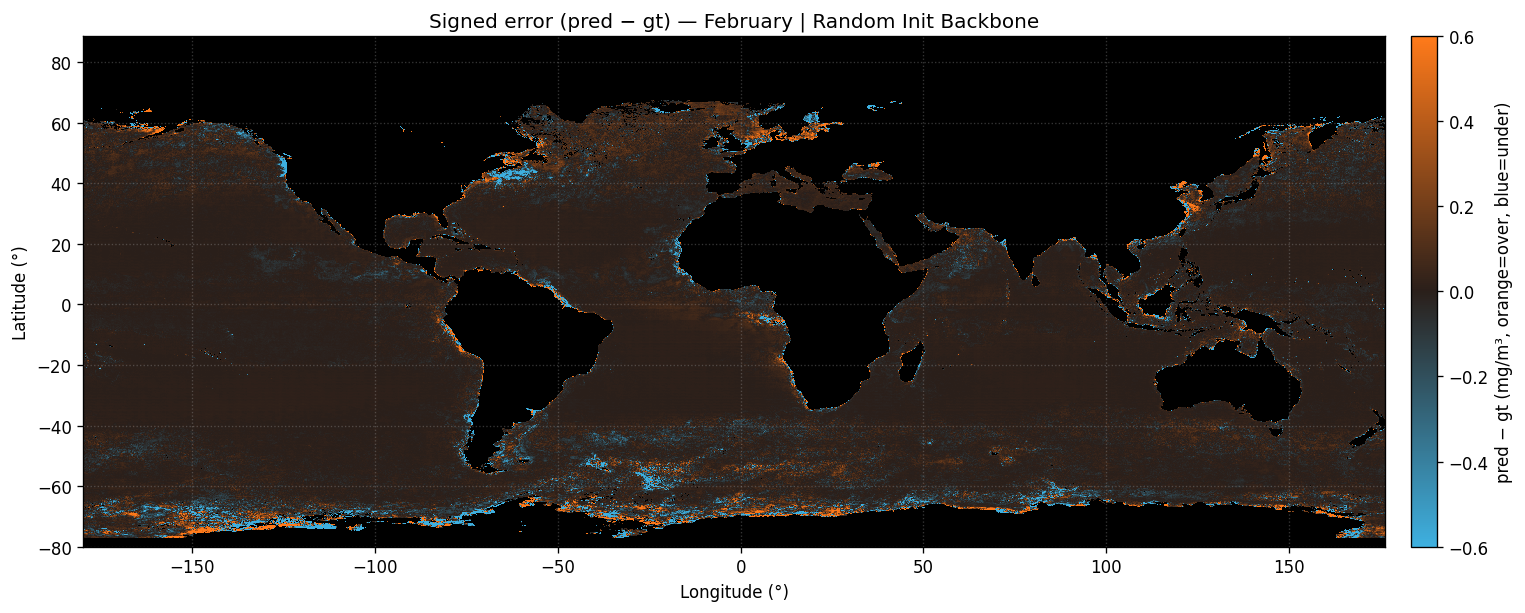} \\[0.2em]
    \includegraphics[width=0.485\textwidth]{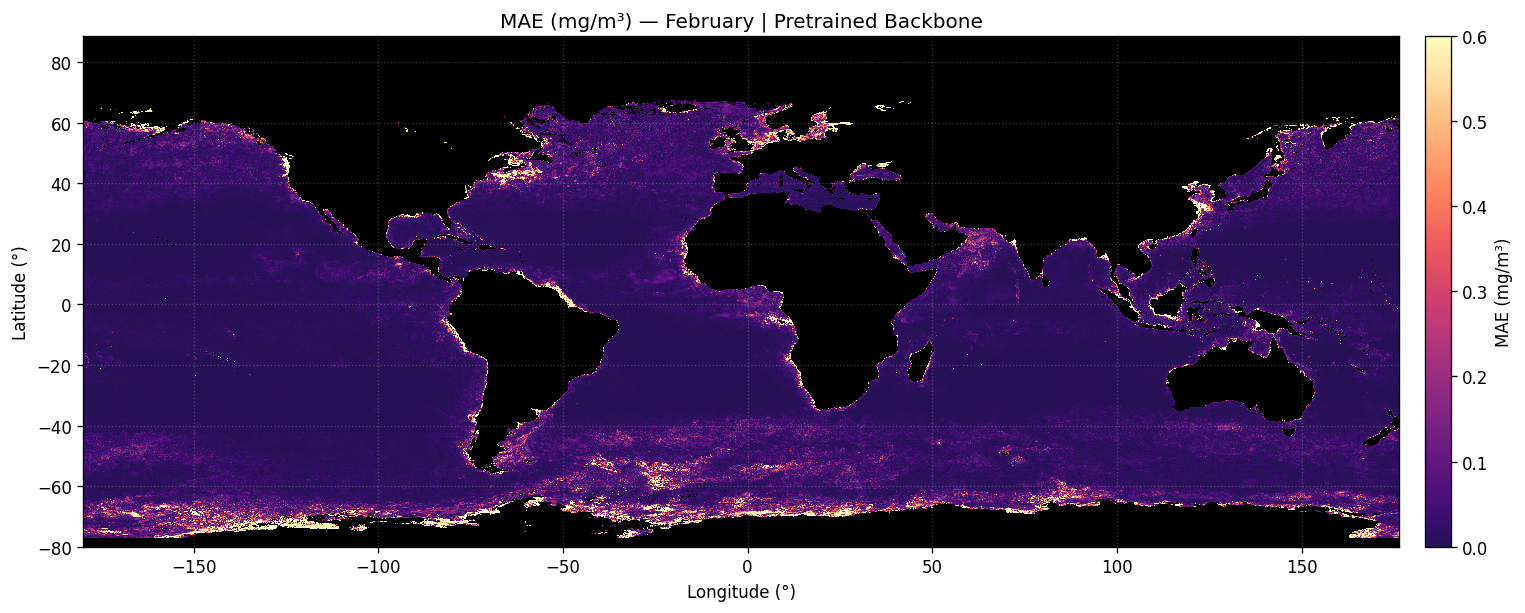} &
    \includegraphics[width=0.485\textwidth]{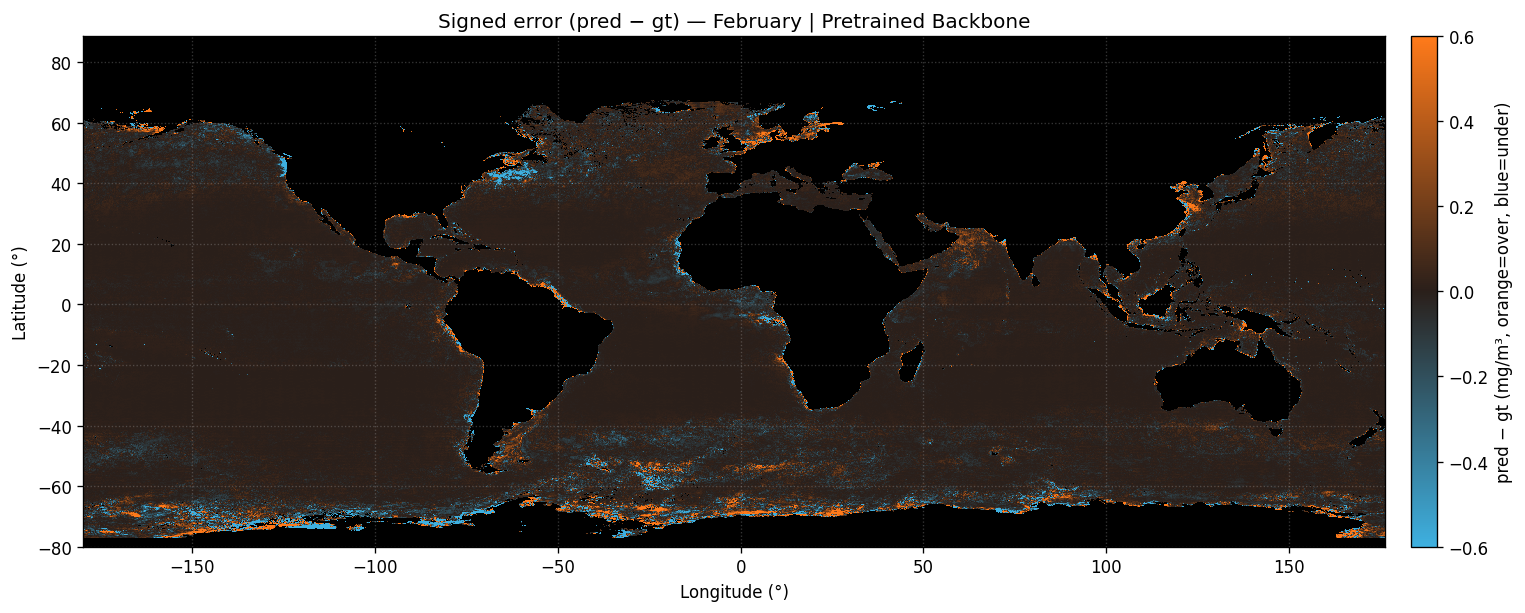}
  \end{tabular}
  \caption{Per-pixel MAE (left column: absolute; right column: signed, pred\,$-$\,gt) for February 2022.
           Rows: Baseline, Random Init, Pretrained DARWIN.
           The baseline performs significantly worse near California.}
  \label{fig:qual_mae_feb}
\end{figure}

A pronounced spatial divergence emerges in the error maps (Figure~\ref{fig:qual_mae_feb}). Along the California coast, the baseline substantially over-predicts chlorophyll-\textit{a}. Both encoder-based configurations (random init and pretrained DARWIN) do not exhibit this systematic positive bias, suggesting that the richer spatiotemporal context captured by the transformer backbone helps resolve the fine-scale coastal dynamics. A close-up of the California region is shown in Appendix Figure~\ref{fig:supp_california_mae_feb}.

\begin{figure}[H]
  \centering
  \begin{tabular}{@{}cc@{}}
    \includegraphics[width=0.485\textwidth]{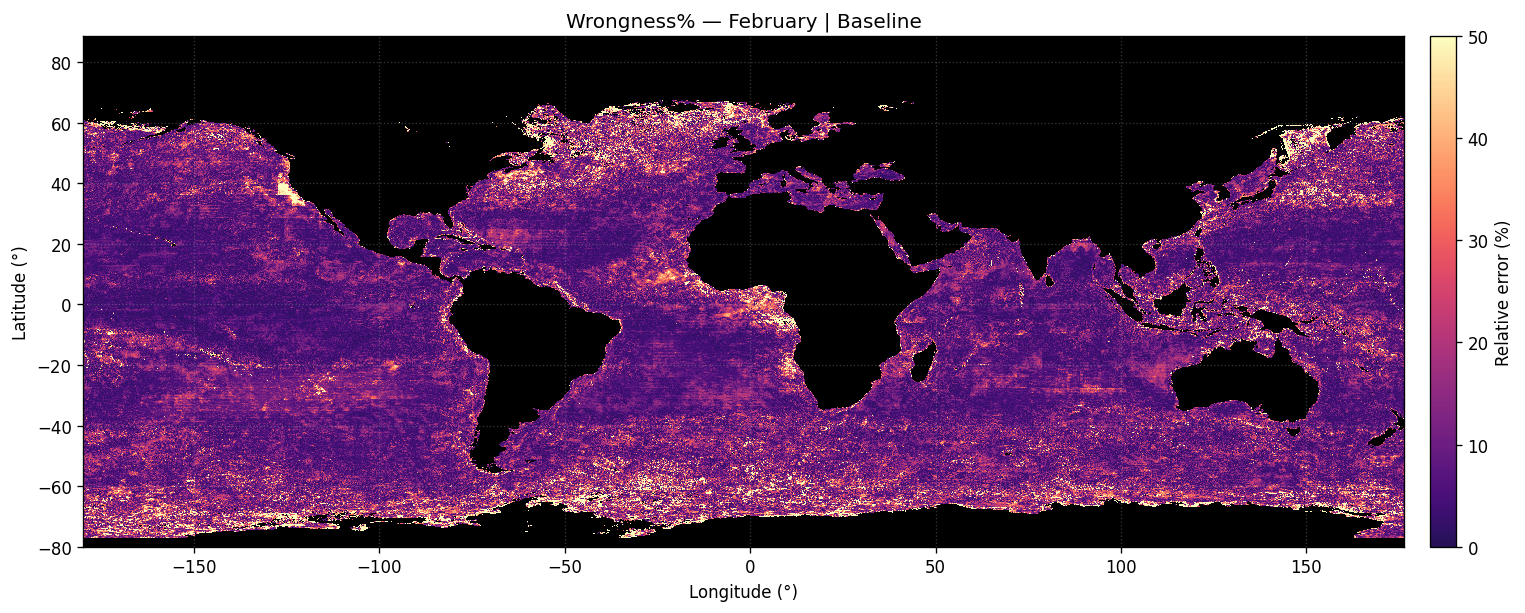} &
    \includegraphics[width=0.485\textwidth]{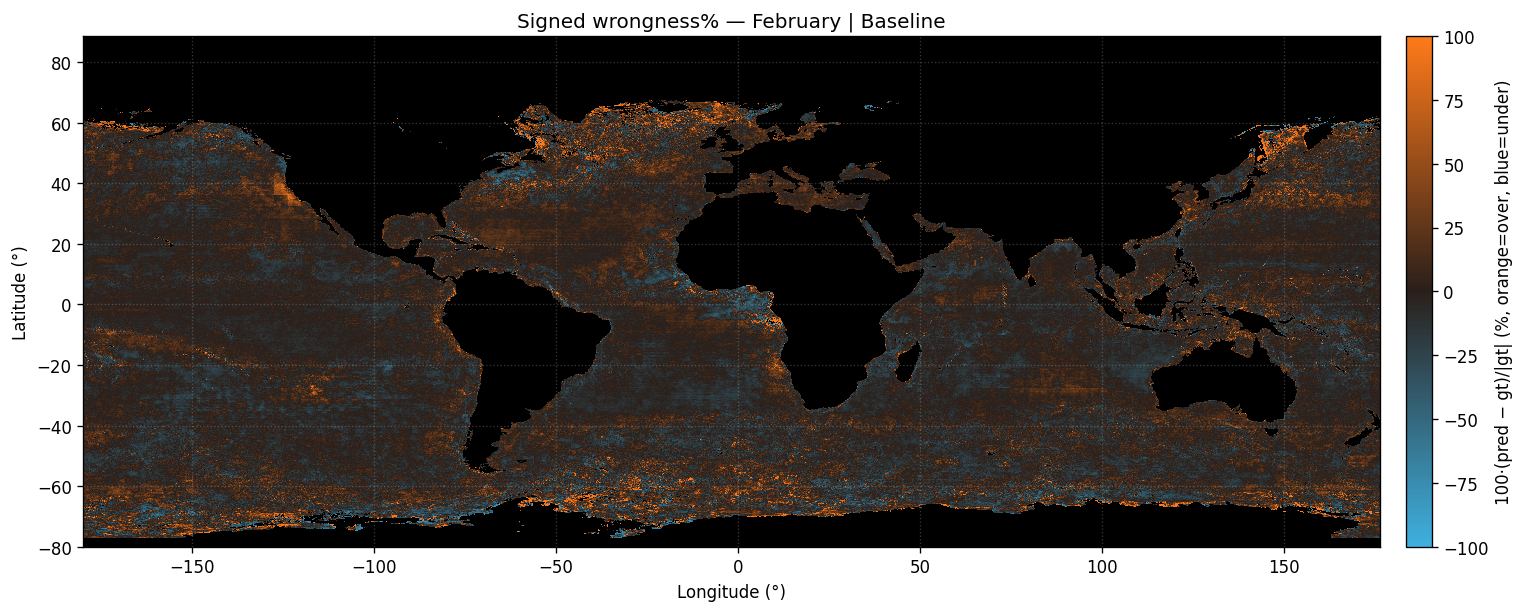} \\[0.2em]
    \includegraphics[width=0.485\textwidth]{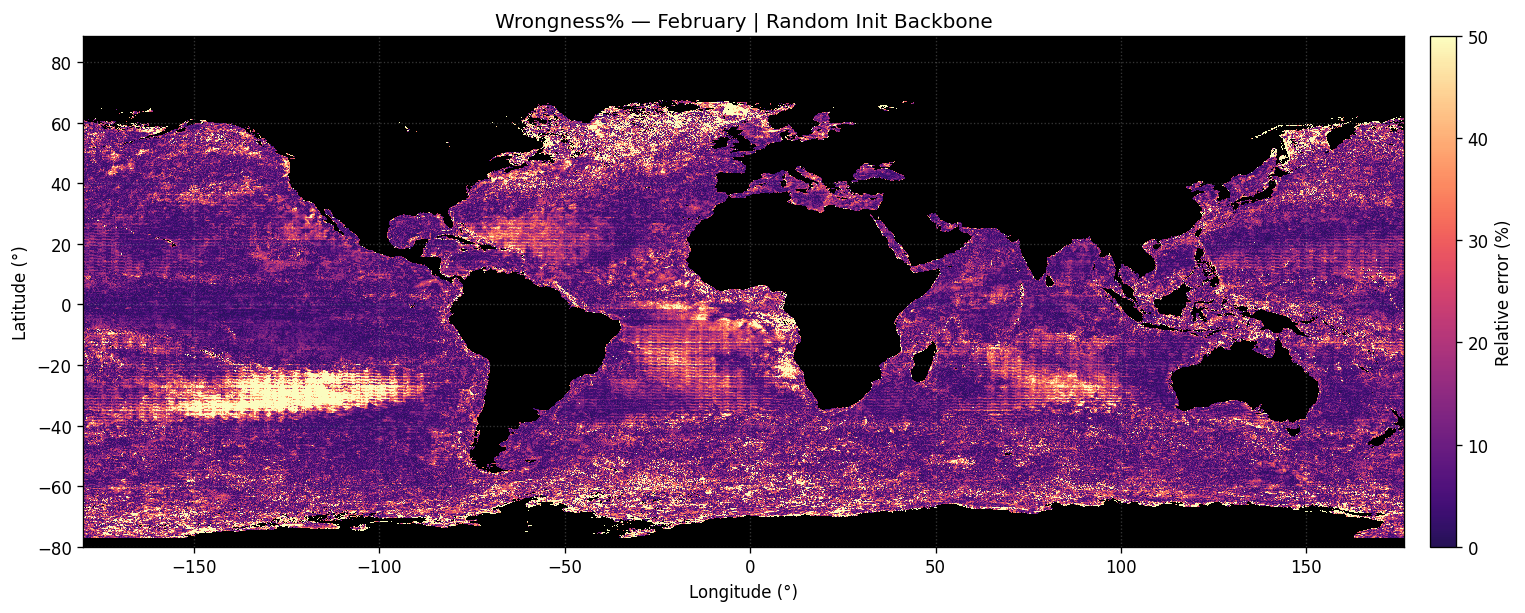} &
    \includegraphics[width=0.485\textwidth]{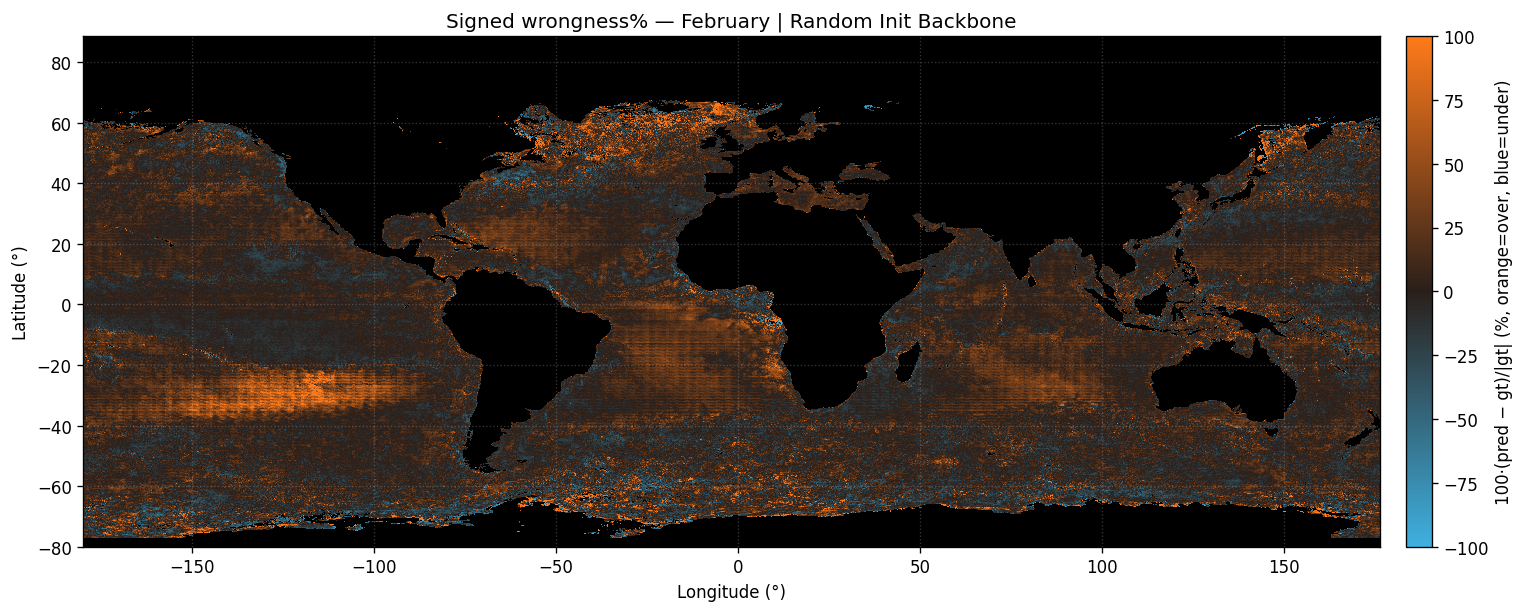} \\[0.2em]
    \includegraphics[width=0.485\textwidth]{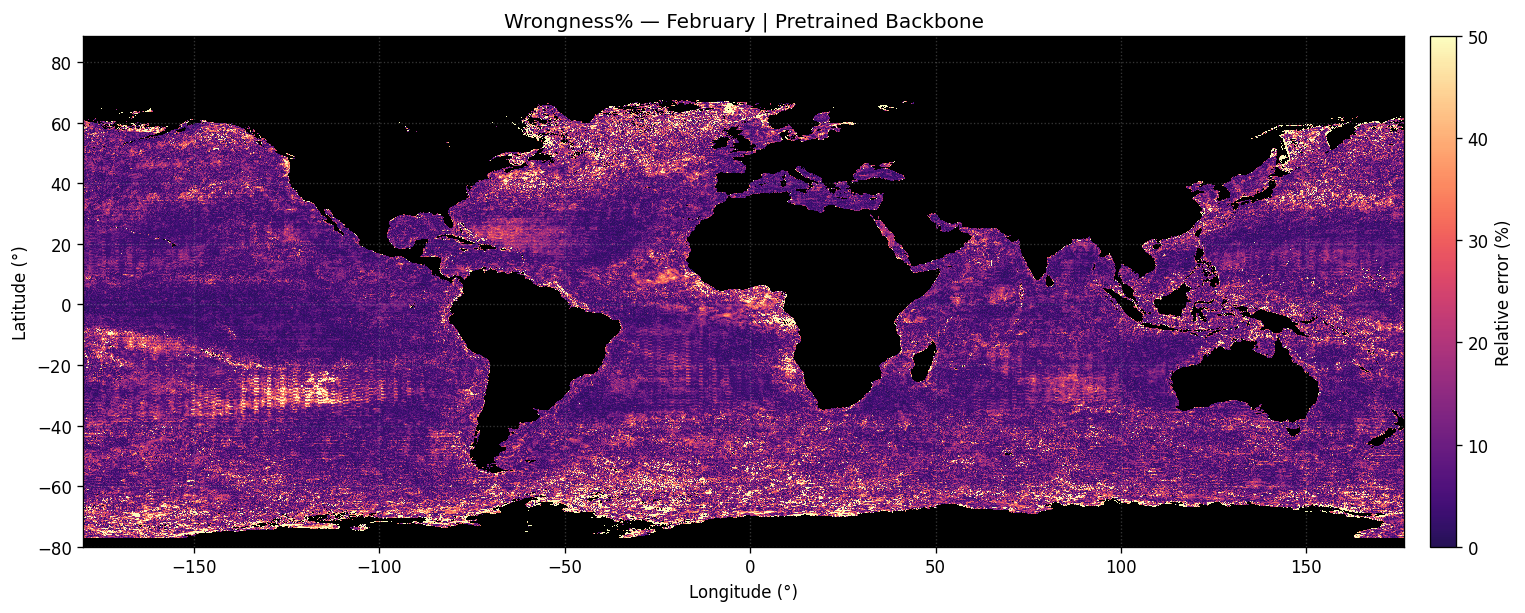} &
    \includegraphics[width=0.485\textwidth]{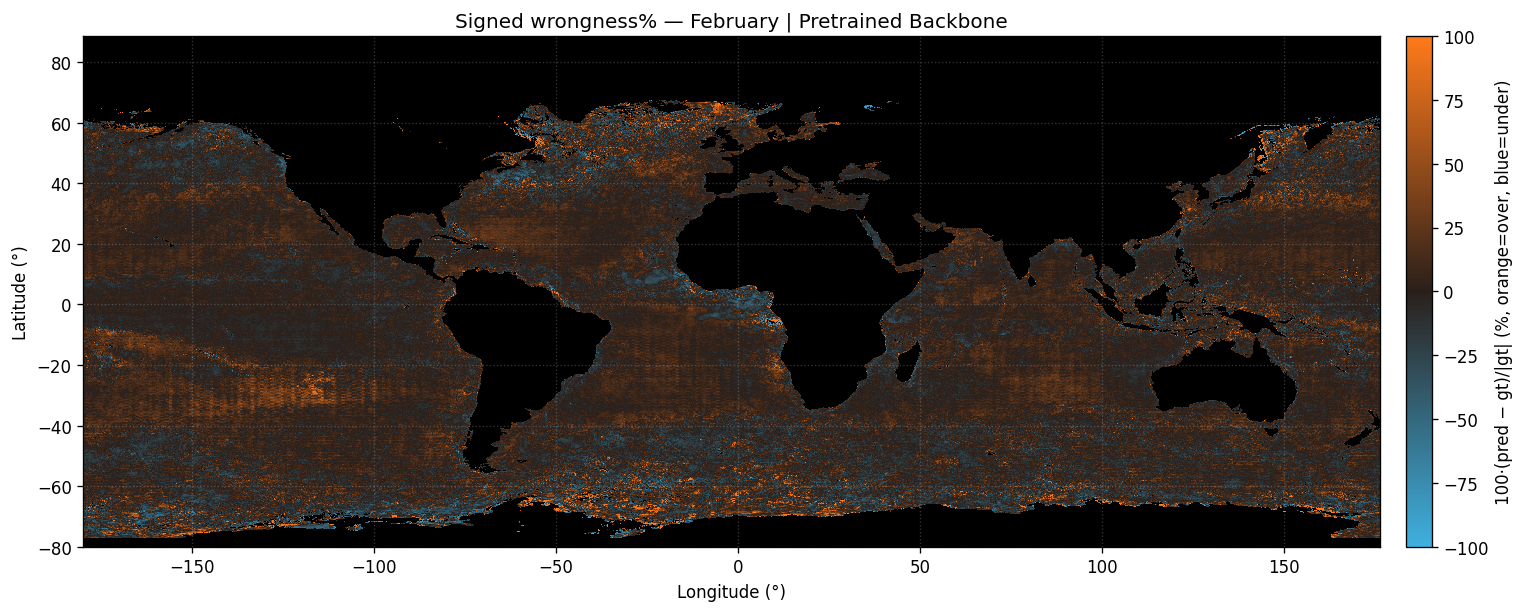}
  \end{tabular}
  \caption{Per-pixel relative error (wrongness, \%) for February 2022 (left: absolute; right: signed).
           Rows: Baseline, Random Init, Pretrained DARWIN.
           The difference in error is mainly around California and the larger area to the south-west.}
  \label{fig:qual_wrongness_feb}
\end{figure}

The wrongness maps (Figure~\ref{fig:qual_wrongness_feb}) reveal a complementary pattern. In regions where chlorophyll concentrations are very low, the baseline achieves the lowest relative error. Among the encoder-based models, pretrained DARWIN is substantially less wrong than random init in these same low-concentration regions. The absolute error in these areas is \textit{negligibly} small, so the practical consequence of the baseline's relative advantage is limited.

In contrast, in regions where chlorophyll concentrations are high, pretrained DARWIN is the most accurate model. Random init improves over the baseline in these regions but does not close the gap to the pretrained configuration. This can be seen in more detail in the California zoom-in in Appendix Figure~\ref{fig:supp_california_wrongness_feb}.

This points to a clear and ecologically interpretable conclusion: \textit{where it matters most} (high-chlorophyll, biologically productive regions), pretrained DARWIN is the best-performing model, with random init as a distant second. In oligotrophic regimes the baseline has a marginal relative advantage, but pretrained DARWIN remains close while random init degrades more noticeably in relative terms.

A further qualitative difference concerns spatial smoothness. In regional close-ups (Appendix Figures~\ref{fig:supp_europe_pred_log_jun} and~\ref{fig:supp_california_pred_log_jun}; June, European waters and California coast), the random-init predictions appear visibly pixelated and patchy, whereas pretrained DARWIN and the baseline both produce spatially coherent, smooth fields. This suggests that DARWIN's pretrained representations encode spatial regularity (reflecting learned ocean physics) in a way that does not emerge spontaneously from random initialization, even after full end-to-end fine-tuning on the same data.

\begin{figure}[H]
  \centering
  \includegraphics[width=\textwidth]{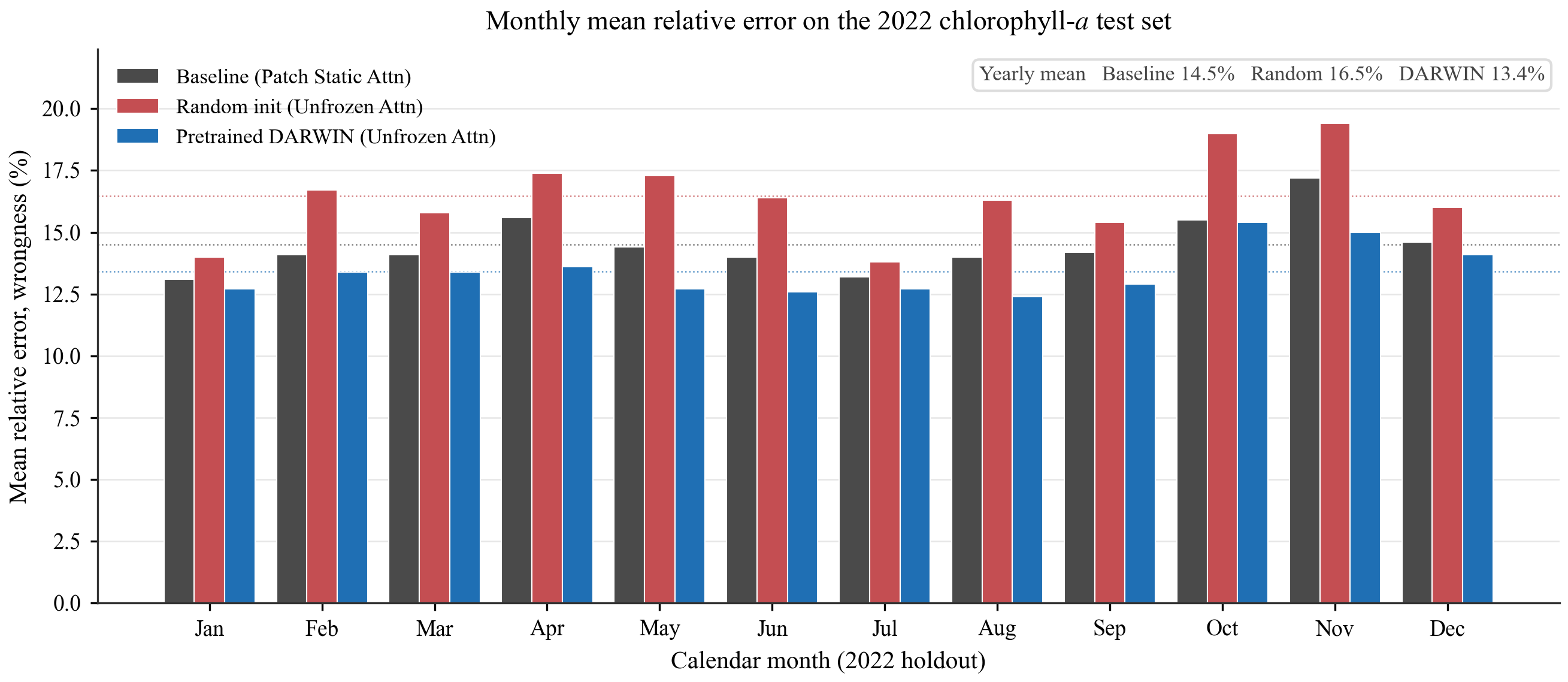}
  \caption{Monthly mean relative error (\textit{wrongness}, $100\,|\hat y - y|/\max(|y|,\varepsilon)$, $\varepsilon=10^{-3}$~mg\,m$^{-3}$) on the 2022 chlorophyll-\textit{a} test holdout. Dotted horizontal lines mark each model's yearly mean. Pretrained DARWIN is the lowest-error configuration every month.}
  \label{fig:downstream_monthly_wrongness}
\end{figure}

\paragraph{Monthly aggregated relative error.}
\autoref{fig:downstream_monthly_wrongness} aggregates wrongness over the full global mosaic per month. The pretrained DARWIN model is the lowest-error configuration in every month of the 2022 holdout, the random-init backbone is the highest, and the  baseline sits between them. The yearly mean is $13.4\%$ for pretrained DARWIN, $14.5\%$ for the baseline and $16.5\%$ for random init.

\section{Discussion}


The marine-specific adaptations introduced in this research address the structural challenges outlined in the introduction. Missingness-aware encoding and expanded masking strategies directly target the sparse, heterogeneous nature of ocean observations, while Fourier-style positional embeddings capture the multi-scale, non-stationary dynamics that distinguish marine from terrestrial domains. Our experiments provide consistent evidence that our self-supervised pretraining methodology on reanalysis data yields transferable representations of marine state. On the 2022 chlorophyll-\textit{a} holdout, pretrained DARWIN attains the lowest monthly relative error in every month and the lowest yearly mean (13.4\%), outperforming both the best no-encoder baseline (14.5\%) and an identically architected random-init encoder (16.5\%). The pretraining gain is positive across all eight head $\times$ backbone configurations we tested, and is most pronounced when the backbone is kept frozen (the regime in which representation quality is exposed most directly). The gap between a frozen linear probe and a frozen MLP head further indicates that DARWIN's encoder embeds chlorophyll-relevant signal in a non-linear manifold rather than as a linearly decodable feature.

The benefit compounds in the low-label regime: with only 10\% of the training set, pretrained DARWIN already surpasses a randomly initialised model trained on the full 100\%, and remains best even at 0.1\% (871 chips), a property especially valuable in marine science, where annotated data is scarce and biased toward coastal and surface waters. Spatially, pretraining matters most where it matters most. In high-chlorophyll, biologically productive regions DARWIN is the most accurate model and produces smooth, physically coherent fields, while the random-init backbone yields visibly pixelated predictions on the same data. The baseline retains only a marginal relative edge in regimes where absolute errors are negligibly small. Combined with the structured embedding geometry (a near-isotropic, high-dimensional hypersphere whose PCA projections recover coherent oceanographic regimes) these results indicate that the model captures genuine ocean structure. Three caveats qualify these findings: the $8\,\mathrm{km} \times 8\,\mathrm{km}$  grid and monthly cadence constrain the resolvable phenomena and likely underlie the comparatively weaker temporal reconstructions seen during pretraining; the MAE objective spends capacity on pixel-level reconstruction that may not align with the abstraction useful downstream; and our only downstream validation focuses on a single surface variable.

The embedding analysis suggests the model captures coherent spatial regimes; the comparatively weaker temporal reconstructions during pretraining suggest temporal dynamics are harder to compress at monthly cadence. Downstream tasks requiring anomaly detection or gap-filling may benefit most, while forecasting applications will likely require explicit temporal fine-tuning. DARWIN is best positioned for tasks that leverage spatial context over large areas and tolerate monthly resolution such as species distribution modelling. It is likely less suited to high-frequency phenomena such as storm responses or short-lived bloom events, and to subsurface prediction, where surface reanalysis provides limited observational constraint.

\section{Roadmap for Future work}
\label{chp:roadmap}

\paragraph{Improving the SSL approach:} A significant direction for future research involves moving beyond the Masked Autoencoders (MAE) approach for representation learning. While effective and interpretable to an extent, they are often inefficient as they waste computational power reconstructing high-frequency pixel noise (such as individual blades of grass or lighting variations) that are irrelevant to high-level semantic understanding \cite{leefrequency}. An alternative approach would be a joint embedding frameworks like DINOv2 \citep{oquab2023dinov2}, \citep{simeoni2025dinov3} and transitioning toward architectures based on the Joint Embedding Predictive Architecture (JEPA) \citep{assran2023self}. The core advantage of the JEPA approach lies in its shift from predicting missing pixels to predicting missing representations (embeddings). By forcing the model to predict the underlying reality in a compressed mathematical space, it develops a form of "common sense" or object permanence, reasoning about the big picture rather than local noise. This paradigm serves as a primitive "World Model," where the predictor learns to handle uncertainty by not committing to exact pixel-level details. Building on this, the V-JEPA (Video JEPA) 2.1 \citep{mur2026v} framework extends these spatial predictions into the temporal dimension. By predicting what happens next in time rather than just what exists elsewhere in space, the model develops a more robust understanding of marine dynamics.

\paragraph{Multi-scale, multi-modal pretraining beyond L4:}
A significant limitation of this work is that pretraining relies almost entirely on L4 reanalysis products, which were or have been coarsened onto a single $8 \times 8$~km monthly grid. This has two consequences: the model is trained on what existing operational systems \emph{predict} the ocean to be rather than on what sensors directly observe, and the spatial and temporal structure present in higher-resolution native sources is discarded before it ever reaches the encoder. Incorporating input features from direct satellite observations, along with in-situ measurements, would allow DARWIN to learn from observations rather than learning the assumptions baked into operational ocean models. These input data sources arrive on heterogeneous native grids and at heterogeneous cadences. The goal would be to combine them \emph{without} a coarsening step that erases the very resolution we are trying to preserve.

\paragraph{Grid-agnostic encoding: continuous embeddings and RoPE:}
Realising a multi-modal, multi-scale vision requires breaking an assumption baked into the current architecture: that all inputs share a single discrete pixel grid. To ingest heterogeneously gridded sources without re-coarsening them, discrete patchification could be replaced with a continuous embedding strategy (cross-attention from each source's native grid onto a shared latent token set) following methods such as MOSAIC \citep{zhdanov2026sparse} and ENF \citep{wessels2025grounding}. This would likely remove patching artefacts, let the model natively combine data at all available resolutions, and provide a natural foundation for downstream super-resolution. Moving away from Absolute Positional Encoding (APE) as the carrier of spatial structure. APE injects absolute grid coordinates directly into the model, probably tying learned features to specific patch-grid locations rather than to the geometric relationships between them. Adopting Rotary Positional Embeddings (RoPE), which let attention operate purely on the \emph{relative} distance between tokens and never feed absolute position into the model.

To fully realize the potential of DARWIN as a foundation model for marine intelligence, a next step lies in the systematic development of downstream tasks. These applications serve as benchmarks to evaluate the generalizability and practical utility of the learned representations. One key application is species distribution modelling (SDM), a widely used ecological tool for understanding and predicting the presence of marine species based on environmental variables. Another promising downstream direction is the prediction of coral bleaching events. These are driven by rapid environmental changes, particularly in sea surface temperature and chlorophyll concentration. By training a dedicated classifier or regression head on top of DARWIN’s embeddings, the model could be used to detect early warning signals for bleaching, using in-situ measurements or satellite-based labels for supervision. Similar methods can be extended to detect marine hazards, such as harmful algal blooms, oil spills, and sediment plumes. These events are traditionally modeled with physical simulations, but may benefit from DARWIN’s ability to learn complex, multivariate patterns across space and time.

\section{Conclusion.}
We introduce DARWIN, a marine-specific geospatial foundation model that adapts the masked-autoencoder paradigm to the structural realities of ocean data through missingness-aware encoding, expanded spatial and temporal masking strategies, and Fourier-style positional embeddings. Trained on 26 years of Copernicus reanalysis, DARWIN learns coherent representations that serve as a strong prior for downstream marine prediction, demonstrated by attaining the lowest monthly relative error in every month of 2022 on chlorophyll-\textit{a} prediction, outperforming both no-encoder and randomly initialised baselines, and surpassing the fully supervised random-init model at only 10\% of the labelled training data. Taken together, the results establish DARWIN as a viable and performant foundation model for marine intelligence.

\section*{Environmental impact}

All runs were conducted using Amazon Web Services in region us-east-1, which has a carbon efficiency of 0.37 kgCO$_2$eq/kWh.

For \textbf{pretraining}, a cumulative of 296 aggregate GPU hours of computation was performed on hardware of type A100 PCIe 40/80GB (TDP of 250W). Emissions for this are estimated to be \textbf{27.38 kgCO$_2$eq}, none of which were directly offset by the cloud provider.

For \textbf{experiments and downstream task runs}, a cumulative of approximately \num{5000} aggregate GPU hours of computation was performed on hardware of type A10G (TDP of 150W). Emissions for this are estimated to be \textbf{277.5 kgCO$_2$eq}, none of which were directly offset by the cloud provider.

Estimations were conducted using the \href{https://mlco2.github.io/impact#compute}{MachineLearning Impact calculator} presented in \cite{lacoste2019quantifying}. These calculations exclude the use of LLM's, which were used cautiously in this research. Running these models requires significant computational resources and water consumption. While exact figures are not available, we highlight the environmental impact and encourage future research to consider environmental responsibility when conducting scientific studies.

\section*{Acknowledgments}
We gratefully acknowledge the AWS GenAI for geospatial program for providing the computing credits that supported this work, and Copernicus Marine for granting us direct programmatic access to their data on AWS S3. Sincere gratitude to Dr. Andrew Corbett for his valuable advice throughout this project. We would also like to thank Wessel Bruinsma PhD for his insightful input and Daan Heijke for serving as an excellent sounding board during discussions.  Finally, our thanks go to the entire team at Ocean OS (David Lamb, Joshua McCarty, Rhys Feeney, Kate Maurer-Song, Vincent Opitz, and Jasmine Brown) for their continued support.

\bibliographystyle{plainnat}
\bibliography{bibentries}

\appendix

\section{Model implementation}
\subsection{Positional Encodings}
\label{appendix:positional_encodings}

DARWIN combines a fixed three-dimensional sinusoidal grid encoding with four Fourier-based encodings that condition the model on the absolute geographic, spatial-scale, and temporal context of each chip. The grid encoding provides absolute position of patches within a chip, while the Fourier streams inject the absolute coordinates of the chip on Earth and in time. The design follows Aurora~\citep{bodnar2024foundation}, with adaptations for variable-resolution chips and CLS-token conditioning.

\paragraph{Three-dimensional grid encoding.}
Each patch token receives a fixed sin-cos positional embedding over the $(T, H, W)$ grid of patch indices. The embedding dimension is split across width ($\tfrac{6}{16}D$), height ($\tfrac{6}{16}D$), and time ($\tfrac{4}{16}D$), with each axis encoded original transformer formulation. The CLS token receives a zero grid embedding. This stream provides relative position within a chip and is the only positional signal added to every patch token unconditionally.

\paragraph{Absolute spatial encoding.}
The geographic location of the chip is encoded with a Fourier expansion of its central latitude and longitude. The centre is computed as the mean latitude and the circular mean longitude over the chip pixels, the latter using an $\arctan2$ of the mean sine and cosine to remain robust at the $\pm 180^{\circ}$ meridian. The latitude and longitude each occupy half of the embedding dimension. Wavelengths are log-spaced between $\lambda_{\min} = 0.01$ and $\lambda_{\max} = 180$ for latitude, and $\lambda_{\min} = 0.01$ and $\lambda_{\max} = 360$ for longitude, allowing the embedding to capture both local variations and global-scale patterns. The result is a single embedding per chip that is added to the CLS token, which thereby acts as a learned summary of the chip's geographic context.

\paragraph{Scale encoding.}
To make the model aware of the physical extent of a chip --- which varies with both latitude and the source dataset's resolution --- a scale encoding is included. For a chip bounded by latitudes $\phi_1, \phi_2$ and longitudes $\theta_1, \theta_2$, its physical area is
\[
A = R^2 \, (\sin \phi_2 - \sin \phi_1)(\theta_2 - \theta_1),
\]
with $R = 6371$~km and $\theta_1, \theta_2$ in radians; the longitudinal span is wrapped to the principal branch so that anti-meridian crossings are handled correctly. The square root $\sqrt{A}$ is fed to a Fourier expansion whose wavelengths span from the smallest patch area at the poles to the total surface area of the Earth. Taking the square root yields units of kilometres and means the encoding scales linearly with linear resolution, so that high-resolution chips are mapped to small-scale values and coarser chips to larger ones. Like the absolute spatial encoding, the scale embedding is added to the CLS token.

\paragraph{Temporal encoding.}
Time is represented at two complementary scales, both added to every patch token of the corresponding timestep (broadcast over space):
\begin{itemize}
    \item \textbf{Hour-to-year cycles.} The absolute hour of each timestep is encoded with a Fourier expansion whose wavelengths are log-spaced between one hour and $24 \times 365.25$ hours. This single stream simultaneously captures diurnal, weekly, and seasonal periodicities. The range assertion is disabled so that absolute hour counts since a reference epoch can be passed in unchanged.
    \item \textbf{Year (interannual trend).} The year value is encoded with the standard non-periodic sin-cos positional encoding (base $10{,}000$) used in the original transformer. Because the year is a slowly varying scalar relative to the encoding's wavelengths, this stream behaves as a smooth ramp and lets the model learn non-periodic effects such as climate drift.
\end{itemize}

\paragraph{Combining the encodings.}
Each of the four Fourier encodings is independently passed through an encoding-specific linear layer followed by dropout, then multiplied by a per-stream scalar gate before being added to its target token(s). Concretely, the absolute spatial and scale embeddings are added to the CLS token (as it is chip level information), while the hourly and yearly embeddings are added to every patch token of the matching timestep. The dropout encourages the model to rely on the broader positional context rather than memorising any single encoding, while the per-stream linear layers allow the network to learn how much of each Fourier basis to retain.

\subsection{Expanded Masking Strategies}
\label{appendix:masking_strategies}

DARWIN extends the random masking strategy of standard MAEs with two additional schemes designed to strengthen temporal and spatial reasoning. At each training step, exactly one of the three strategies (random, temporal, or spatial) is sampled uniformly at random and applied to the entire batch. For the temporal and spatial schemes the sampled mask is shared across all samples in the batch, while the random scheme draws an independent mask per sample.

Inputs are organised as a sequence of patch tokens corresponding to a $(T, H, W)$ grid; in our setup with $45\times45$ pixel chips of 12 timesteps and a $(1, 3, 3)$ patch size, this gives a $(12, 15, 15)$ patch grid.

\paragraph{Random masking.}
A subset of patches is masked independently per sample:
\[
\forall (t,h,w) \in \mathcal{M}: \quad X_{t,h,w,:} \mapsto \texttt{[MASK]},
\]
where $\mathcal{M}$ is sampled per sample by ranking i.i.d.\ uniform noise and
keeping the lowest-ranked $\lceil (1-r_{\text{rand}})\,T H W \rceil$ tokens
visible. We set $r_{\text{rand}} = 0.75$, following standard MAE practice.

\begin{figure}[H]
  \centering
  \includegraphics[width=\textwidth]{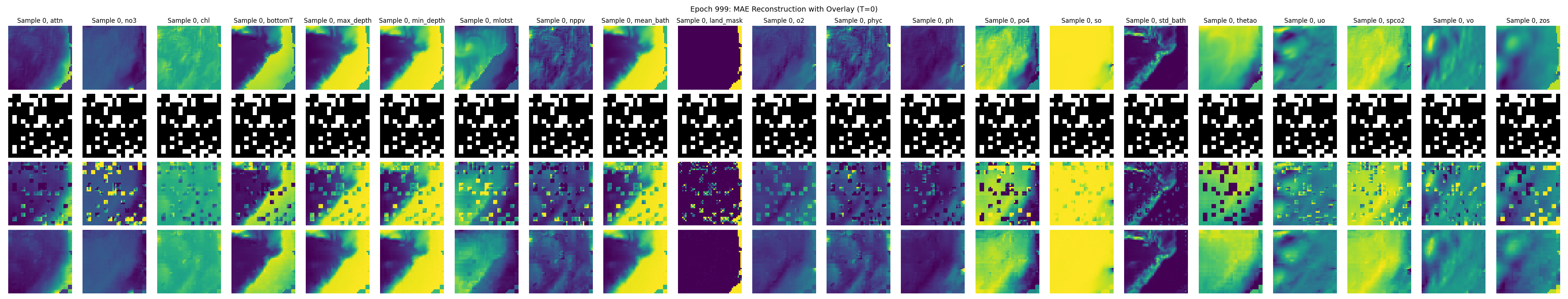}
  \caption{Example of random masking.}
  \label{fig:rand_mask_v2}
\end{figure}

\paragraph{Temporal-slice masking.}
A subset $\mathcal{T} \subset \{1,\dots,T\}$ of complete temporal slices is masked, with $|\mathcal{T}| = \lfloor r_{\text{time}} T \rfloor$ and $\mathcal{T}$ drawn as a uniform permutation of the time axis (so the masked months are not constrained to be contiguous):
\[
\forall t \in \mathcal{T}: \quad X_{t,:,:,:} \mapsto \texttt{[MASK]}.
\]
We set $r_{\text{time}} = 0.6$, which with $T = 12$ corresponds to 7 of 12 months hidden. (with this ratio, at least two consecutive months are always masked). This forces the model to infer missing months from seasonal continuity rather than from local spatial context within the same timestep.

\begin{figure}[H]
  \centering
  \includegraphics[width=0.9\textwidth]{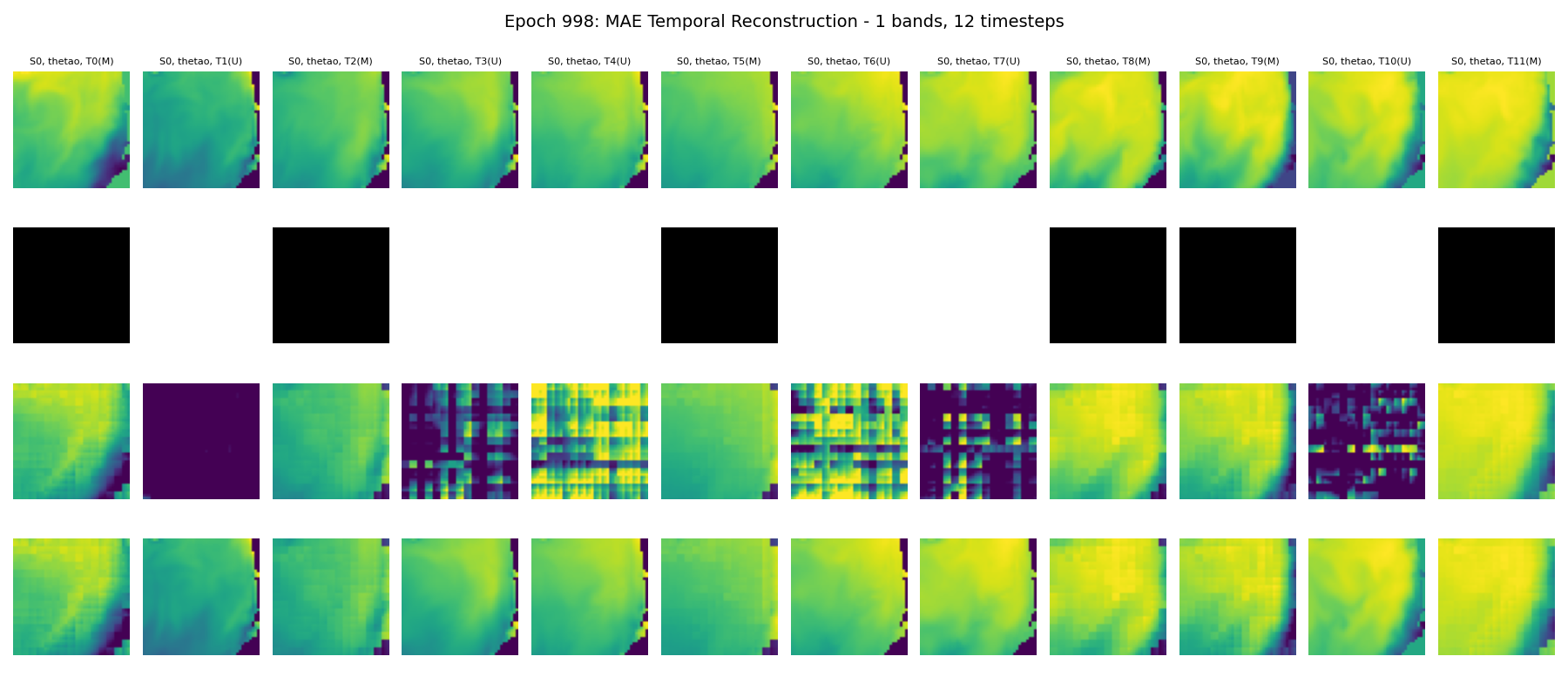}
  \caption{Example of time-slice masking.}
  \label{fig:temp_mask_v2}
\end{figure}

\paragraph{Spatial column masking.}
A single axis-aligned tube is selected in the spatial plane and masked across \emph{every} timestep, producing a column through time rather than a finite spatiotemporal tube:
\[
X_{:,\,h_0:h_0+k,\,w_0:w_0+k,\,:} \mapsto \texttt{[MASK]},
\]
where the side length is $k = \lfloor r_{\text{spat}} \cdot H \rfloor$ and $(h_0, w_0)$ is sampled uniformly so the square fits inside the chip. Note that $r_{\text{spat}}$ specifies the \emph{linear} side as a fraction of the chip side, not the masked fraction of patches: with $r_{\text{spat}} = 0.75$ and $H = W = 15$, the square covers $11 \times 11 = 121$ of $225$ spatial patches across all $T = 12$ timesteps, i.e.\ approximately $54\%$ of the total patches. This forces the model to reconstruct mesoscale spatial patterns (e.g.\ eddies, fronts) using only the unmasked surrounding context in every timestep, since the masked region never reappears within the chip.

\begin{figure}[H]
  \centering
  \includegraphics[width=\textwidth]{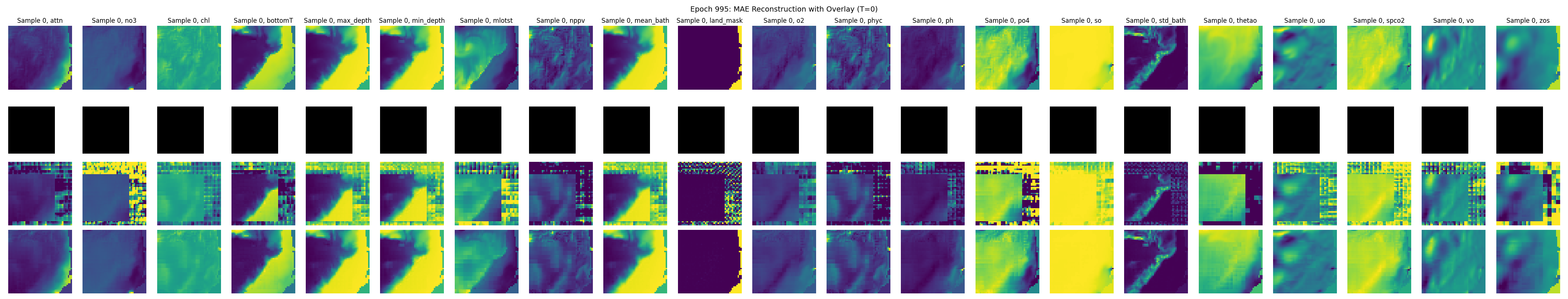}
  \caption{Example of spatial column masking.}
  \label{fig:spat_mask_v2}
\end{figure}

\paragraph{Sampling scheme.}
At each forward pass, one of the three strategies above is selected uniformly at random ($p = 1/3$ each) and applied to the whole batch. The three strategies are therefore mixed across training iterations rather than blended within a single sample, exposing the model to the full distribution of reconstruction tasks while keeping each individual forward pass internally consistent.

\section{Variables in dataset}

\subsection{Used in pretraining}
\label{appendix:variables}

\begin{table}[H]
\centering
\caption{Data Variables Used for Pre-training.}
\label{tab:copernicus_variables}
\small
\begin{tabular}{lll}
\toprule
Variable & Description & Unit \\
\midrule
\textbf{Physics} & & \\
Temperature & Sea Water Potential Temperature & [°C] \\
Salinity & Sea Water Salinity & [1e-3] \\
Sea surface height & Sea Surface Height Above Geoid & [m] \\
Velocity (zonal) & Eastward Sea Water Velocity & [m\,s$^{-1}$] \\
Velocity (meridional) & Northward Sea Water Velocity & [m\,s$^{-1}$] \\
Sea ice concentration & Sea Ice Area Fraction & [1] \\
Sea ice thickness & Sea Ice Thickness & [m] \\
Sea ice velocity (u) & Eastward Sea Ice Velocity & [m\,s$^{-1}$] \\
Sea ice velocity (v) & Northward Sea Ice Velocity & [m\,s$^{-1}$] \\
Bottom temperature & Sea Floor Potential Temperature & [°C] \\
Mixed layer depth & Ocean Mixed Layer Thickness & [m] \\
\midrule
\textbf{Biogeochemistry} & & \\
Chlorophyll & Chlorophyll-\textit{a} Mass Concentration & [mg\,m$^{-3}$] \\
Nitrate & Nitrate Concentration & [mmol\,m$^{-3}$] \\
Phosphate & Phosphate Concentration & [mmol\,m$^{-3}$] \\
Silicate & Silicate Concentration & [mmol\,m$^{-3}$] \\
Oxygen & Dissolved Oxygen Concentration & [mmol\,m$^{-3}$] \\
Primary production (v) & Net Primary Production (Volume) & [mg\,m$^{-3}$\,day$^{-1}$] \\
Iron & Iron Concentration & [mmol\,m$^{-3}$] \\
Phytoplankton (C) & Phytoplankton Concentration in Carbon & [mmol\,m$^{-3}$] \\
pH & Sea Water pH (Total Scale) & [1] \\
Surface pressure CO$_2$ & Surface Partial Pressure of CO$_2$ & [Pa] \\
\midrule
\textbf{Biomass} & & \\
Zooplankton (C) & Mass Content of Zooplankton (Carbon) & [mg\,m$^{-2}$] \\
Micronekton (epi) & Epipelagic Micronekton Biomass & [g\,m$^{-2}$] \\
Micronekton (umeso) & Upper Mesopelagic Micronekton Biomass & [g\,m$^{-2}$] \\
Micronekton (mumeso) & Migrant Upper Mesopelagic Biomass & [g\,m$^{-2}$] \\
Micronekton (lmeso) & Lower Mesopelagic Micronekton Biomass & [g\,m$^{-2}$] \\
Micronekton (mlmeso) & Migrant Lower Mesopelagic Biomass & [g\,m$^{-2}$] \\
Micronekton (hmlmeso) & Highly Migrant Lower Mesopelagic Biomass & [g\,m$^{-2}$] \\
Primary productivity & Net Primary Productivity of Biomass & [mg\,m$^{-2}$\,day$^{-1}$] \\
Euphotic depth & Euphotic Zone Depth & [m] \\
\midrule
\textbf{Ocean Colour} & & \\
DIATO / DINO & Diatoms / Dinophytes Chl-a & [mg\,m$^{-3}$] \\
HAPTO / GREEN & Haptophytes / Green Algae Chl-a & [mg\,m$^{-3}$] \\
PROCHLO / PROKAR & Prochlorococcus / Prokaryotes Chl-a & [mg\,m$^{-3}$] \\
MICRO / NANO / PICO & Micro/Nano/Picophytoplankton Chl-a & [mg\,m$^{-3}$] \\
KD490 & Diffuse Attenuation Coefficient & [m$^{-1}$] \\
SPM & Suspended Particulate Matter & [g\,m$^{-3}$] \\
\midrule
\textbf{Atmospheric/Other} & & \\
Wind speed & Sea Surface Wind Speed & [m\,s$^{-1}$] \\
Eastward wind & Eastward Wind Velocity & [m\,s$^{-1}$] \\
Northward wind & Northward Wind Velocity & [m\,s$^{-1}$] \\
Eastward stress & Surface Downward Eastward Stress & [Pa] \\
Northward stress & Surface Downward Northward Stress & [Pa] \\
Wind stress magnitude & Magnitude of Surface Downward Stress & [Pa] \\
Land mask & Land-Sea Mask & [1 or 0] \\
Bathymetry & Mean Seafloor Depth & [m] \\
\bottomrule
\end{tabular}
\end{table}

\subsection{Data division for downstream task pretraining}

\begin{table}[H]
\centering
\caption{Included static variables for the chlorophyll-\textit{a} downstream task.}
\label{tab:included_static}
\begin{tabular}{@{}lp{10cm}@{}}
\toprule
\textbf{Category} & \textbf{Variables} \\
\midrule
\textbf{Included Static (5)} & \texttt{ocean\_mask}, \texttt{min\_bathy}, \texttt{max\_bathy}, \texttt{mean\_bath}, \texttt{std\_bath} \\
\bottomrule
\end{tabular}
\end{table}

\begin{table}[H]
\centering
\caption{Included dynamic variables for the chlorophyll-\textit{a} downstream task.}
\label{tab:included_dynamic}
\begin{tabular}{@{}lp{10cm}@{}}
\toprule
\textbf{Category} & \textbf{Variables} \\
\midrule
\textbf{Included Dynamic (26)} & \texttt{bottomT}, \texttt{thetao}, \texttt{so}, \texttt{zos}, \texttt{uo}, \texttt{vo}, \texttt{usi}, \texttt{vsi}, \texttt{eastward\_wind}, \texttt{northward\_wind}, \texttt{eastward\_stress}, \texttt{northward\_stress}, \texttt{wind\_stress\_magnitude}, \texttt{wind\_speed}, \texttt{o2}, \texttt{ph}, \texttt{sithick}, \texttt{siconc}, \texttt{mlotst}, \texttt{fe}, \texttt{no3}, \texttt{po4}, \texttt{si}, \texttt{spco2}, \texttt{KD490}, \texttt{SPM} \\
\bottomrule
\end{tabular}
\end{table}

\begin{table}[H]
\centering
\caption{Omitted dynamic variables for the chlorophyll-\textit{a} downstream task.}
\label{tab:omitted_dynamic}
\begin{tabular}{@{}lp{10cm}@{}}
\toprule
\textbf{Category} & \textbf{Variables} \\
\midrule
\textbf{Omitted Dynamic (21)} & \texttt{DIATO}, \texttt{DINO}, \texttt{GREEN}, \texttt{HAPTO}, \texttt{MICRO}, \texttt{NANO}, \texttt{PICO}, \texttt{PROCHLO}, \texttt{PROKAR}, \texttt{chl}, \texttt{phyc}, \texttt{npp}, \texttt{nppv}, \texttt{zeu}, \texttt{zooc}, \texttt{mnkc\_epi}, \texttt{mnkc\_hmlmeso}, \texttt{mnkc\_lmeso}, \texttt{mnkc\_mlmeso}, \texttt{mnkc\_mumeso}, \texttt{mnkc\_umeso} \\
\bottomrule
\end{tabular}
\end{table}

\section{Dataset Pipeline}
\label{dataset_pipeline}

\subsection{Creation of the Global Dataset}

The construction of the terabyte-scale unified global ocean dataset required preprocessing and merging multiple oceanographic products into a unified dataset. This process involved defining a master grid, regridding/interpolating, and harmonizing land-sea masks to ensure a continuous signal for self-supervised pretraining. It consists of monthly mean data.

\paragraph{Master Grid and Temporal Reference} We used the Physics Reanalysis product as the primary reference for the spatial and temporal grid. We defined the global extent by trimming all variables to a shared bounding box: Latitude $[-80.0, 89.58]$ and Longitude $[-179.83, 179.75]$. This resulted in a final grid dimension of $\num{2036} \times \num{4316}$.

\paragraph{Land-Sea Masking} The ocean–land mask was derived from the physics mlotst (mixed-layer thickness) variable over the full dataset period. We chose mlotst because it exhibits fewer transient NaNs than the velocity components  (\texttt{uo}, \texttt{vo}), giving a stable global coastline. We apply a conservative criterion: a cell is classified as ocean only if it is valid at every timestep, so intermittently missing pixels are excluded. We treat this mask as ground truth for the marine realm. Any dynamic values falling on land pixels were treated as such, while all ocean pixels were required to have valid numerical data.

\paragraph{Multi-Modal Harmonization} Integrating Biomass, Biogeochemical (BGC), Colour, and Wind data presented several challenges due to varying coastal coverage and native resolutions:

\begin{itemize}
    \item \textbf{BGC and Biomass} data contain more coastal NaNs than physical variables due to the underlying numerical models. To ensure a gap-free data, we applied \textbf{biharmonic interpolation} to fill these coastal gaps before reapplying the master land mask. This was only done for a very small percentage of the data.
    \item \textbf{Ocean Colour:} Because of the higher resolution of this data, The computational cost of global interpolation was high, so we utilized a different approach: extending bounds, coarsening the resolution for the interpolation step, and then re-aligning to the master grid. This resulted in very negligible numerical differences.
    \item \textbf{Sea Ice} Sea ice concentration and thickness were imputed with 0 (no ice) where data was missing within the ocean mask, though velocity components remained as NaNs (not imputed to 0). If ice is not present, there should be no observable speed.
    \item \textbf{Wind} Wind data was upsampled using linear interpolation to match the 8~km grid.
    \end{itemize}

\paragraph{Static Data Integration} Bathymetry was sourced from GEBCO (2025 release). Because GEBCO and the Copernicus Physics products use slightly different geoids, coarsening leaves residual NaN gaps at coastal boundaries. We synchronised them by setting bathymetry to zero at any coordinate defined as land by the master mask. After this, we calculated mean, min, max, and std while coarsening. We addressed minor artifacts (e.g., positive values or extreme outliers).

\paragraph{Final Spatial Cropping} Following integration of all data streams, the merged dataset is cropped to the largest spatial extent over which every layer is simultaneously valid. Boundary rows and columns are excluded to eliminate edge effects introduced during regridding and interpolation  (in particular, the polar Arctic rows and the longitudinal wrap-around columns, where several source products do not provide full coverage, in particular at the polar boundaries and longitudinal extremes). The result is a spatially consistent domain in which all variables share an identical, artifact-free grid.

\subsection{Creation of the Downstream Task Dataset}
\label{sec:downstream_preprocessing}
For downstream validation, the Level 3 (L3) monthly composite chlorophyll-\textit{a} product is spatially aligned with the DARWIN pretraining grid. The raw L3 data is coarsened to a 8~km spatial resolution and interpolated to ensure consistent geographic extent with the Level 4 (L4) reanalysis data used during pretraining.


\subsection{Storage: Zarr}
To ensure maximal throughput during training, the final dataset was stored in the \texttt{Zarr v3} format. This format efficiently supports cloud-based streaming and optimises I/O operations. The Zarr format supports N-dimensional data and allows read and write arrays concurrently from multiple threads or processes.

Typical geospatial (xarray) datasets store each variable as a separate array with its own chunking along shared coordinates (lat, lon, time). When sampling a spatiotemporal window, this structure requires reading $n_{\text{chunks}} \times n_{\text{variables}}$ chunks. With 47 variables, this created excessive storage access which resulted in idle GPU time. To fix this, we stacked all variables along a new “variable” coordinate axis, converting the dataset into a unified array with shape \texttt{[variable, time, lat, lon]}. Since Zarr doesn’t support string values, variable names were mapped to integers. This restructuring reduced the chunk count per sample and resulted in a $\sim$10x speedup.

We chunk the full dataset in chunks of $H=32, W=32$ and $T=12$ to match the model’s chip input size. This chunking strategy minimized I/O overhead by ensuring that only the required chunks were loaded —no partial overlaps— which improves memory usage \footnote{\url{https://coenvde.github.io/blogs/training-at-larger-scale/part7.html}}. We tested multiple threading configurations with Dask and Zarr, and settled on Zarrv3-native threading. This had similar or better performance than Dask’s multiprocessing and left more workers available for the PyTorch dataloader.

Preprocessing and indexing of the dataset were handled using the \texttt{xarray} library, while sample generation was managed using \texttt{xbatcher} \footnote{\url{https://earthmover.io/blog/cloud-native-dataloader}}. Statistics deviation were precomputed per variable stored in json for standardization during preprocessing.\\

\subsection{Preprocessing}

After each data chip is loaded into memory, a series of preprocessing steps are applied before feeding it into the model. First, a land mask is extracted from the chip, identifying which values correspond to land (NaN). This mask is later used in the loss function to ensure that the model does not attempt to learn from land regions. Next, each variable within the chip is standardised. If a distribution is heavily skewed, we first log transform and then standardise via mean and standard deviation of the log distribution. Otherwise, we standardise using precomputed variable means and standard deviations. This step stabilizes gradients, enables reweighting of variables for a better learning signal, and improves training efficiency. We intentionally avoided techniques like batch normalization that are routinely employed in image-focussed applications. Since the variables in our dataset, such as temperature and chlorophyll concentration, represent units with measurements rather than RGB values, it is important for the model to see the (dataset, per variable standardised) absolute values. This standardization approach is also used in other geospatial foundation models \citep{szwarcman2024prithvi}. After standardization, all NaN values are replaced with zeros. This replacement ensures consistent tensor shapes for input to the model. Importantly, these zero values are not treated as meaningful inputs; the model is guided to ignore them through the "missingness embedder" and the application of the land mask in the loss function, thereby preventing gradients from propagating through invalid data regions. We set land values to zero specifically in the input (using the same embedding function), so they do not influence gradient updates to the patch embedding weights, though they still affect the bias.

\section{Pretraining hyperparameters}
\label{appendix:training_config}

We pretrain two variants of the DARWIN foundation model: \textbf{DARWIN}, the main model trained on the full set of 47 dynamic input variables, and \textbf{DARWIN-NoColour}, an ablation trained on a 26-variable subset that excludes all biological and ocean-colour variables (chlorophyll, phytoplankton functional types, net primary productivity, and related fields). Both variants share the same training recipe, masking strategy, and regularisation; they differ only in the dynamic input feature count and in the encoder/decoder widths and head counts. Hyperparameters were selected following established practice in masked autoencoder pretraining \citep{he2022masked,feichtenhofer2022masked} and adapted to the demands of marine geospatial data.

\begin{table}[H]
\centering
\caption{Optimisation and training schedule. Shared across both pretraining runs.}
\label{tab:training_hyperparams}
\begin{tabular}{ll}
\toprule
\textbf{Hyperparameter} & \textbf{Value} \\
\midrule
Optimiser                       & AdamW \\
Peak learning rate              & $5 \times 10^{-4}$ \\
Minimum learning rate           & $5 \times 10^{-6}$ \\
Adam $(\beta_1, \beta_2)$       & $(0.9,\, 0.95)$ \\
Weight decay                    & $0.05$ \\
Learning rate schedule          & Linear warmup, cosine decay \\
Warmup steps                    & $3{,}000$ ($3\%$ of total) \\
Gradient clipping ($\ell_2$)    & $1.0$ \\
Maximum training steps          & $100{,}000$ \\
Batch size (per GPU)            & $64$ \\
Mixed precision                 & \texttt{bf16-mixed} \\
Hardware                        & $8\times$ NVIDIA A100 \\
Random seed                     & $42$ \\
\bottomrule
\end{tabular}
\end{table}

\begin{table}[H]
\centering
\caption{Model architecture for the two pretraining runs. Both share the spatio-temporal chip and patch geometry; the variants differ in input channel count, encoder width, and decoder width.}
\label{tab:arch_hyperparams}
\begin{tabular}{lcc}
\toprule
\textbf{Hyperparameter} & \textbf{DARWIN} & \textbf{DARWIN-NoColour} \\
\midrule
Dynamic input features                          & $47$            & $26$ \\
Static input features                           & $5$             & $5$ \\
Chip spatial size                               & $45 \times 45$  & $45 \times 45$ \\
Chip timesteps                                  & $12$            & $12$ \\
Patch size (space $\times$ space $\times$ time) & $3 \times 3 \times 1$ & $3 \times 3 \times 1$ \\

Values per patch embedding & $423$ & $234$ \\
Encoder embedding dimension                     & $512$           & $256$ \\
Encoder depth / attention heads                 & $8\;/\;8$       & $8\;/\;4$ \\
Decoder embedding dimension                     & $384$           & $128$ \\
Decoder depth / attention heads                 & $4\;/\;4$       & $4\;/\;4$ \\
\bottomrule
\end{tabular}
\end{table}

\begin{table}[H]
\centering
\caption{Masking and regularisation. Shared across both pretraining runs.}
\label{tab:masking_hyperparams}
\begin{tabular}{ll}
\toprule
\textbf{Setting} & \textbf{Value} \\
\midrule
Random masking ratio                       & $0.75$ \\
Temporal-slice masking ratio               & $0.60$ \\
Spatial-tube masking ratio                 & $0.75$ \\
Dropout, absolute spatial positional enc.  & $0.10$ \\
Dropout, scale positional enc.             & $0.40$ \\
Dropout, absolute temporal positional enc. & $0.05$ \\
Dropout, yearly positional enc.            & $0.20$ \\
\bottomrule
\end{tabular}
\end{table}

\section{Finetuning hyperparameters (chlorophyll downstream)}
\label{appendix:finetune_hyperparams}

This appendix details the hyperparameters used for the Level~3 chlorophyll-\textit{a} dense prediction experiments. The target is the chlorophyll-\textit{a} concentration, standardised with a global $\log(1+x)$ z-score. Encoder inputs are identical to those used at pretraining for the DARWIN-NoColour variant: 26 dynamic physics, atmosphere, and biogeochemistry channels together with five static channels (ocean mask and bathymetry summaries). Patch tokens are obtained from the pretrained encoder and mapped to per-pixel chlorophyll predictions through a lightweight head. During fine-tuning the encoder is run without masking, so every patch contributes to the prediction.

\begin{table}[H]
\centering
\caption{Optimisation and training schedule. Shared across all chlorophyll fine-tuning runs.}
\label{tab:finetune_opt_hyperparams}
\begin{tabular}{ll}
\toprule
\textbf{Hyperparameter} & \textbf{Value} \\
\midrule
Optimiser                    & AdamW \\
Peak learning rate           & $1 \times 10^{-4}$ \\
Minimum learning rate        & $5 \times 10^{-6}$ \\
Adam $(\beta_1, \beta_2)$    & $(0.9,\, 0.95)$ \\
Weight decay                 & $0.05$ \\
Learning rate schedule       & Linear warmup, cosine decay \\
Warmup steps                 & $1{,}000$ ($2\%$ of total) \\
Gradient clipping ($\ell_2$) & $1.0$ \\
Maximum training steps       & $50{,}000$ \\
Batch size        & $64$ \\
Mixed precision              & \texttt{bf16-mixed} \\
Validation interval          & every $2{,}000$ steps \\
Random seed                  & $42$ \\
\bottomrule
\end{tabular}
\end{table}

\begin{table}[H]
\centering
\caption{Prediction-head architectures. All heads are applied to encoder patch tokens; the linear probe has no internal hyperparameters beyond the output projection. MLP and transformer-based heads use a GELU activation and an MLP expansion ratio of $4.0$ in their feed-forward blocks.}
\label{tab:finetune_head_hyperparams}
\begin{tabular}{lcccc}
\toprule
\textbf{Hyperparameter} & \textbf{Linear} & \textbf{MLP} & \textbf{Attention head} & \textbf{DARWIN decoder} \\
\midrule
Hidden dimension                & ---     & $512$   & $256$   & $256$ \\
Depth (transformer blocks)      & ---     & ---     & $1$     & $1$ \\
Attention heads                 & ---     & ---     & $4$     & $4$ \\
MLP expansion ratio             & ---     & $4.0$   & $4.0$   & $4.0$ \\
\bottomrule
\end{tabular}
\end{table}

\subsection{Parameter counts}
\label{appendix:param_counts}

\noindent Parameter counts are a function of architecture alone (independent of weight init), so the \emph{random init} and \emph{pretrained init} variants of each downstream head share the same totals. The No-Color encoder used as the downstream backbone has 6.7\,M parameters; when the backbone is frozen, only the head is trainable.

\begin{table}[H]
\centering
\caption{Pretraining model parameter counts. The full MAE includes both the encoder and the MAE reconstruction decoder; only the encoder is reused downstream.}
\label{tab:pretrain_param_counts}
\begin{tabular}{lrr}
\toprule
\textbf{Model} & \textbf{Full MAE} & \textbf{Encoder only} \\
\midrule
DARWIN FM (with colour) & 34.8\,M & 26.7\,M \\
DARWIN FM (No-Colour)    &  7.6\,M &  6.7\,M \\
\bottomrule
\end{tabular}
\end{table}

\begin{table}[H]
\centering
\caption{Downstream chlorophyll head parameter counts. ``Backbone'' is the NoColour encoder reused across all encoder-based heads (random or pretrained init); baselines have no encoder. Trainable counts equal the total only when the backbone is unfrozen.}
\label{tab:downstream_encoder_param_counts}
\small
\begin{tabular}{llrr}
\toprule
\textbf{Head} & \textbf{Backbone} & \textbf{Total} & \textbf{Trainable} \\
\midrule
\multicolumn{4}{l}{\textit{Encoder-based heads (config\_random\_init / config\_pretrained\_init \(\ast\))}} \\
Linear probe (\texttt{mode: linear}) & frozen & 6.72\,M & 0.00231\,M \\
Linear probe (\texttt{mode: linear}) & unfrozen & 6.72\,M & 6.72\,M \\
2-layer MLP (\texttt{mode: mlp}, hidden 512) & frozen & 6.85\,M & 0.136\,M \\
2-layer MLP (\texttt{mode: mlp}, hidden 512) & unfrozen & 6.85\,M & 6.85\,M \\
Vanilla attention head (\texttt{mode: attention\_head}, depth 1) & frozen & 7.57\,M & 0.858\,M \\
Vanilla attention head (\texttt{mode: attention\_head}, depth 1) & unfrozen & 7.57\,M & 7.57\,M \\
MAE-style decoder (\texttt{mode: decoder}, depth 1) & frozen & 7.85\,M & 1.13\,M \\
MAE-style decoder (\texttt{mode: decoder}, depth 1) & unfrozen & 7.85\,M & 7.85\,M \\
\midrule
\multicolumn{4}{l}{\textit{No-encoder baselines (config\_baseline\_\(\ast\))}} \\
Per-pixel linear (no static) & --- & \num{27} & \num{27} \\
Per-pixel linear (with static) & --- & \num{32} & \num{32} \\
Per-pixel MLP (no static, hidden 512) & --- & \num{14337} & \num{14337} \\
Per-pixel MLP (with static, hidden 512) & --- & \num{16897} & \num{16897} \\
Per-patch linear (no static) & --- & 0.00212\,M & 0.00212\,M \\
Per-patch linear (with static) & --- & 0.00252\,M & 0.00252\,M \\
Per-patch MLP (no static, hidden 512) & --- & 0.125\,M & 0.125\,M \\
Per-patch MLP (with static, hidden 512) & --- & 0.148\,M & 0.148\,M \\
Per-patch attention (no static, depth 1) & --- & 0.853\,M & 0.853\,M \\
Per-patch attention (with static, depth 1) & --- & 0.864\,M & 0.864\,M \\
\bottomrule
\end{tabular}
\end{table}

\subsection{Tree Regressor Implementation}
\label{appendix:tree_reg}

Each pixel-month observation is flattened into a tabular row containing the 26 dynamic reanalysis variables, five static GEBCO channels, latitude, longitude, cyclical month encoding ($\sin$ and $\cos$ of month), and year. Chlorophyll-\textit{a} is $\log_{10}$-transformed and cloud-masked pixels are dropped. A stratified spatial subsample is retained to keep the global dataset tractable on commodity hardware.

Hyperparameters were tuned with Optuna's Tree-structured Parzen Estimator (50 trials, 6-fold spatial block cross-validation using the six validation regions also used for DARWIN). The final models are retrained on a stratified subsample of the full training period with the six validation regions excluded from the fit, so that per-region test scores reflect genuine spatial generalisation rather than memorisation. All random seeds are fixed (matching the DARWIN seed) for reproducibility. Final hyperparameters are reported in \autoref{tab:tree_hparams}.

\begin{table}[H]
\centering
\caption{Tuned hyperparameters for XGBoost and LightGBM baselines.}
\label{tab:tree_hparams}
\begin{tabular}{lcc}
\toprule
\textbf{Hyperparameter} & \textbf{XGBoost} & \textbf{LightGBM} \\
\midrule
n\_estimators       & 1875   & 1701   \\
learning\_rate      & 0.0899 & 0.2011 \\
max\_depth          & 8      & 6      \\
min\_child\_weight / samples & 29 & 31 \\
subsample           & 0.898  & 0.647  \\
colsample\_bytree   & 0.850  & 0.821  \\
reg\_alpha ($L_1$)  & 4.616  & 7.789  \\
reg\_lambda ($L_2$) & 6.248  & 2.041  \\
\midrule
early\_stopping\_rounds & \multicolumn{2}{c}{50} \\
random\_state       & \multicolumn{2}{c}{42} \\
\bottomrule
\end{tabular}
\end{table}

\section{Supplementary downstream qualitative maps}
\label{appendix:downstream_supplementary}


\begin{figure}[H]
  \centering
  \includegraphics[width=0.95\textwidth]{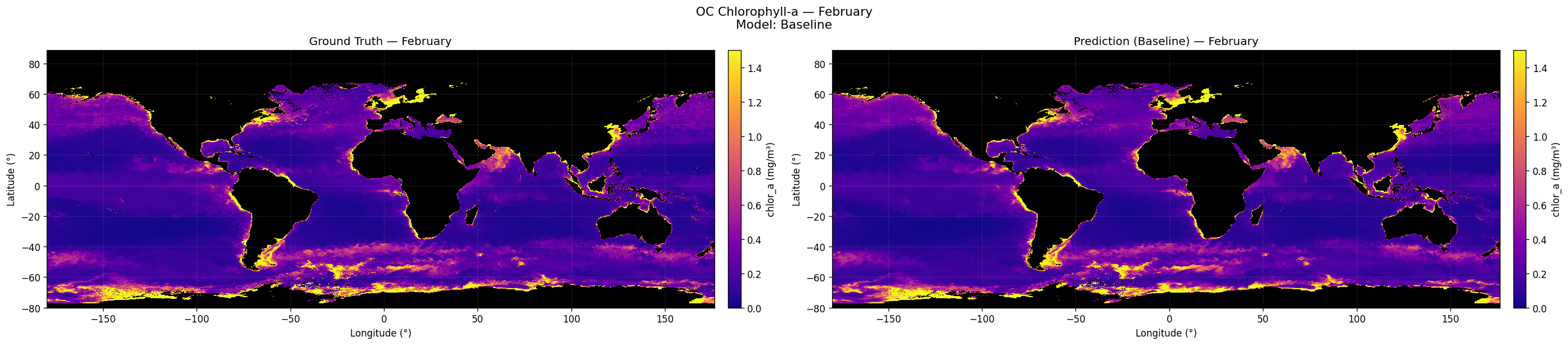}\\[0.3em]
  \includegraphics[width=0.95\textwidth]{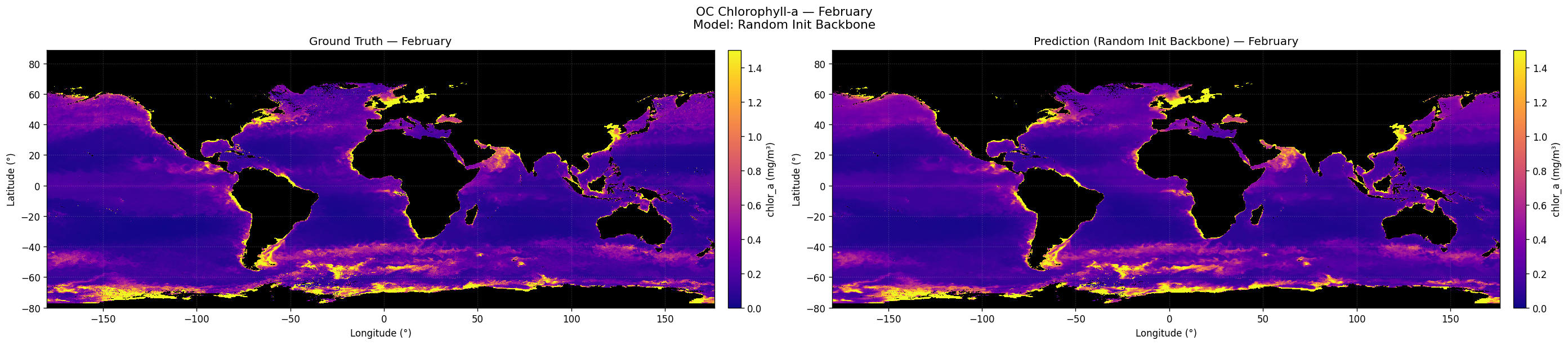}\\[0.3em]
  \includegraphics[width=0.95\textwidth]{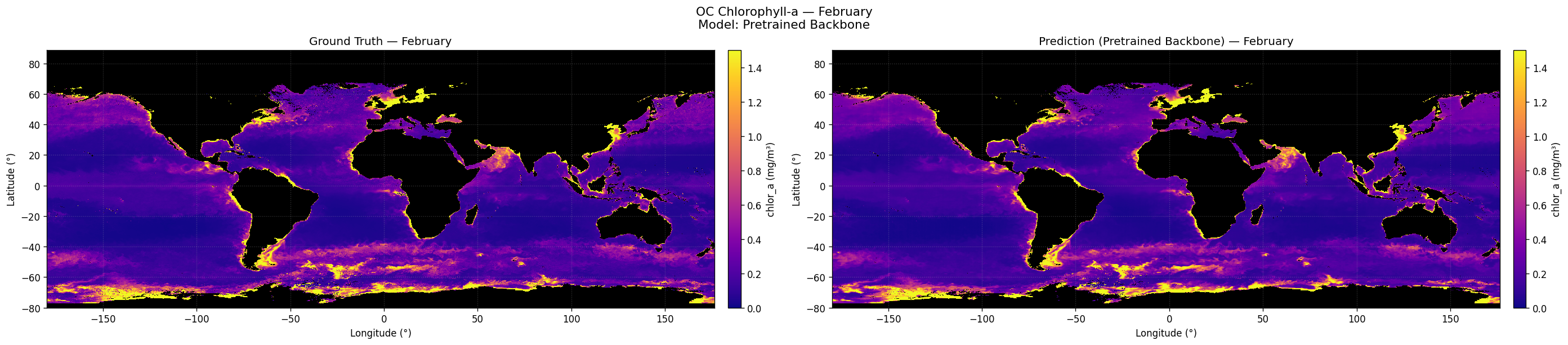}
  \caption{Predicted vs.\ observed chlorophyll-\textit{a} (linear scale, mg\,m$^{-3}$) for February 2022.
           Top to bottom: Baseline, Random Init (unfrozen), Pretrained DARWIN (unfrozen).
           Each row: ground truth (left) and model prediction (right).}
  \label{fig:qual_pred_feb}
\end{figure}

\begin{figure}[H]
  \centering
  \includegraphics[width=0.95\textwidth]{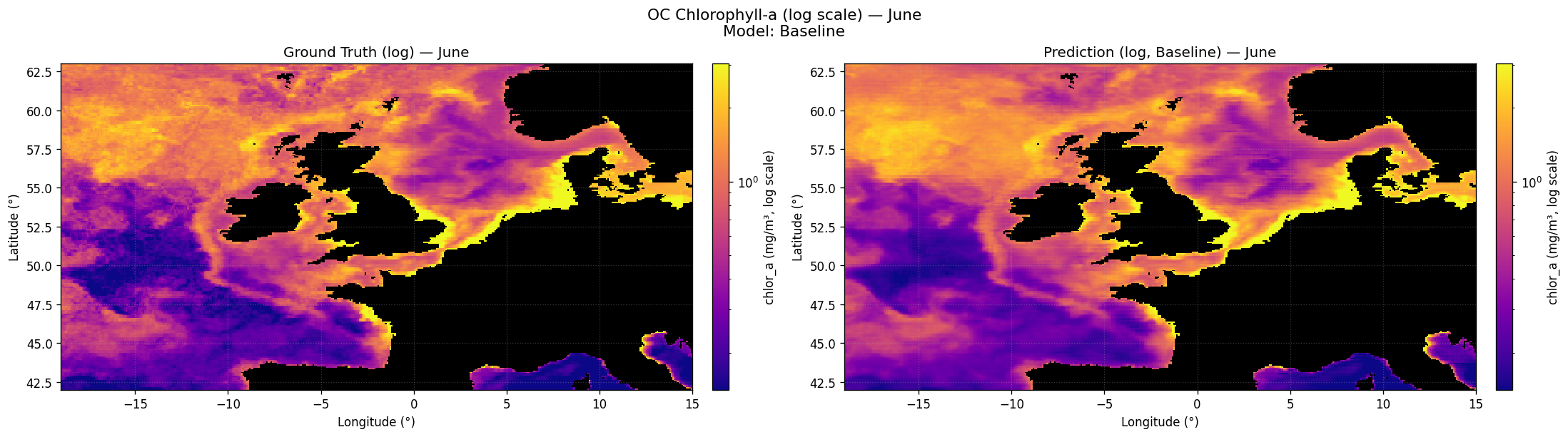}\\[0.3em]
  \includegraphics[width=0.95\textwidth]{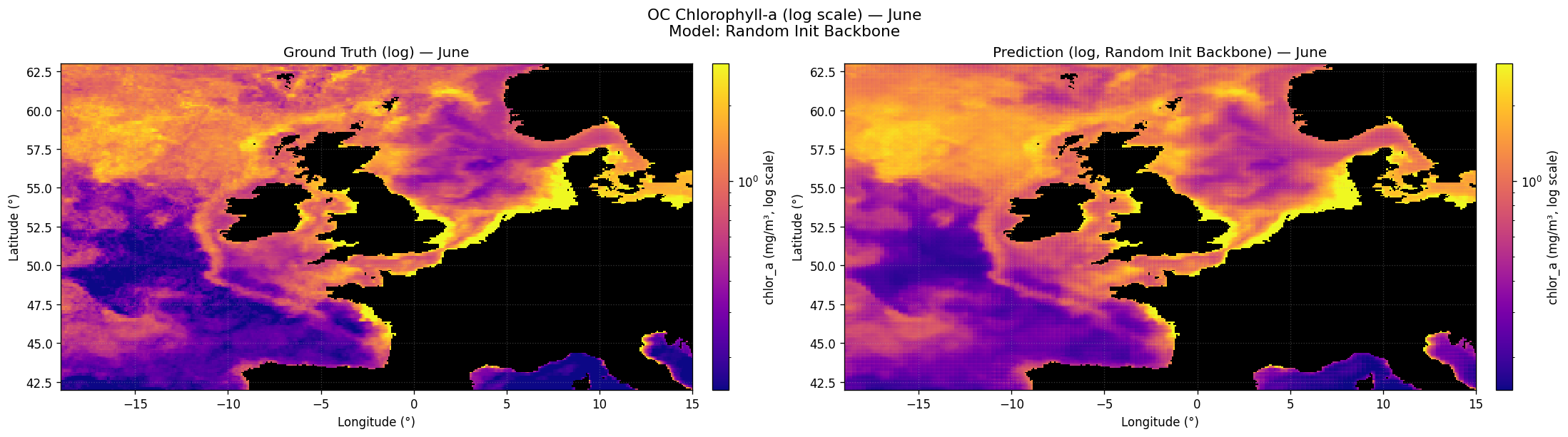}\\[0.3em]
  \includegraphics[width=0.95\textwidth]{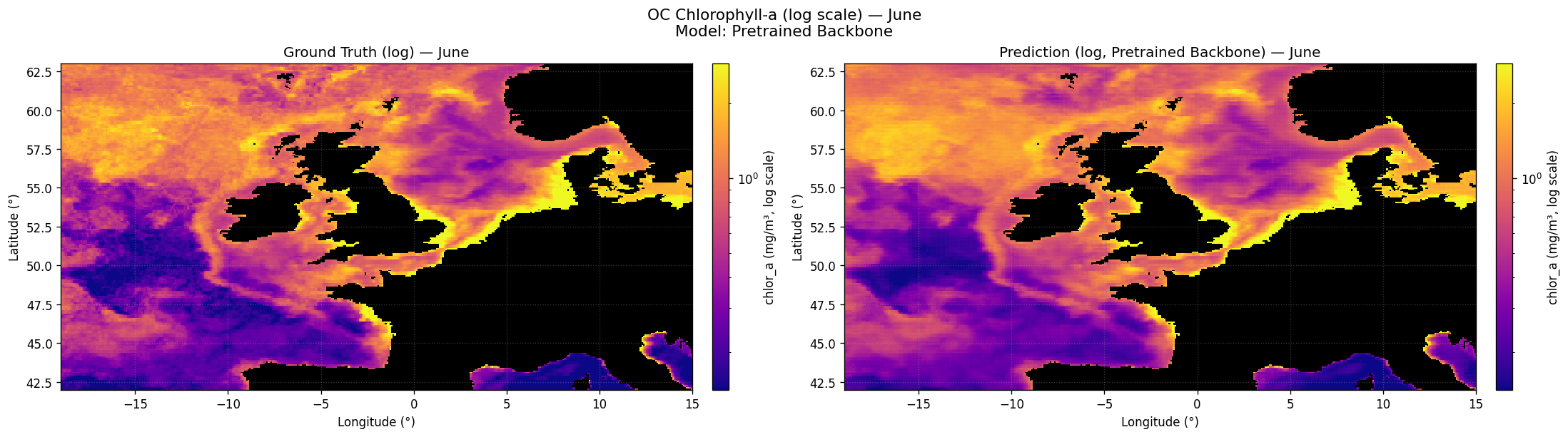}
  \caption{European-extent predicted vs.\ observed chlorophyll-\textit{a} ($\log_{10}$ scale) for June 2022.
           Row order as above.}
  \label{fig:supp_europe_pred_log_jun}
\end{figure}

\begin{figure}[H]
  \centering
  \includegraphics[width=0.95\textwidth]{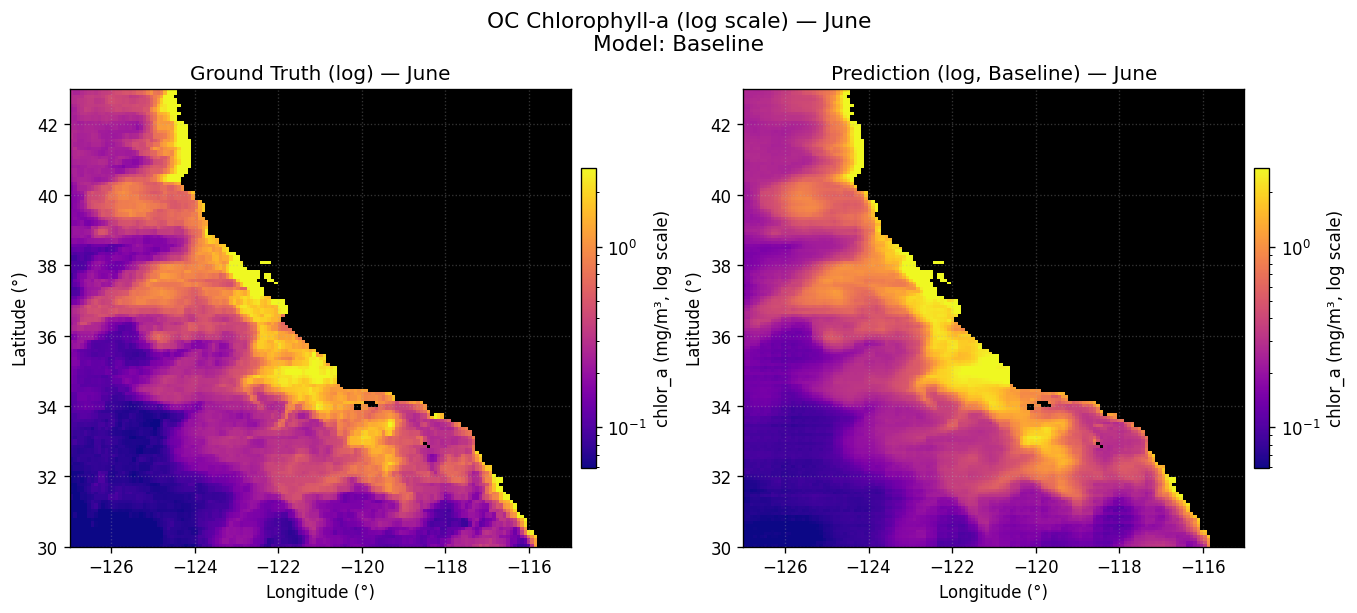}\\[0.3em]
  \includegraphics[width=0.95\textwidth]{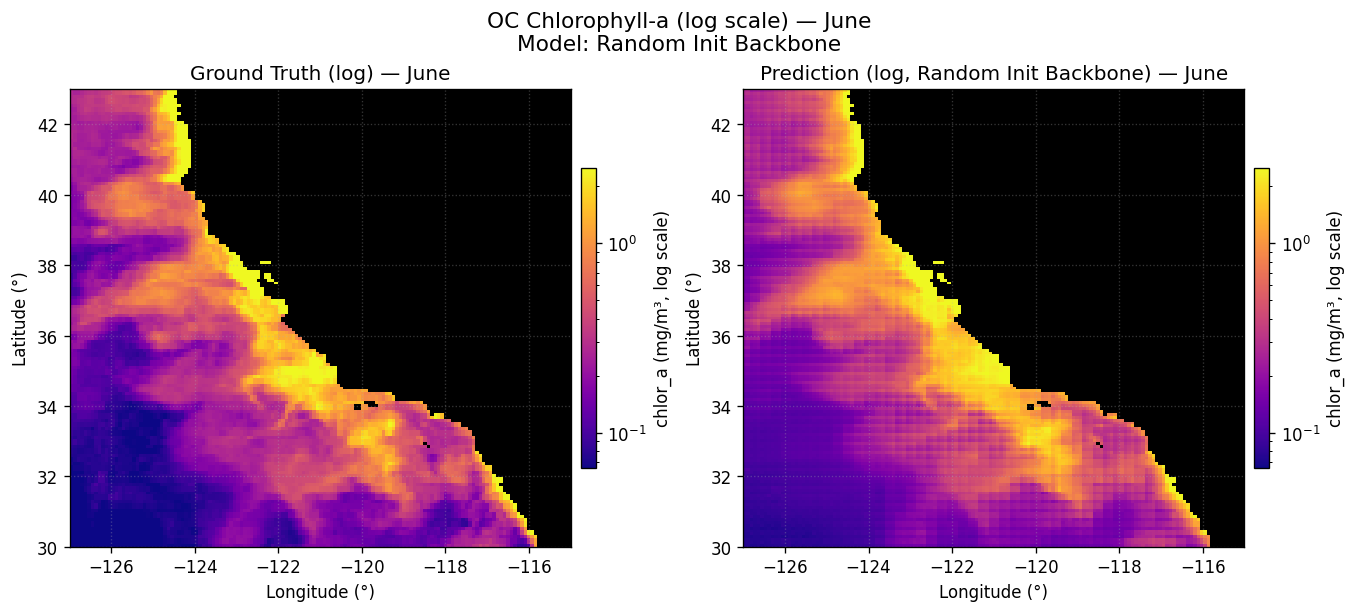}\\[0.3em]
  \includegraphics[width=0.95\textwidth]{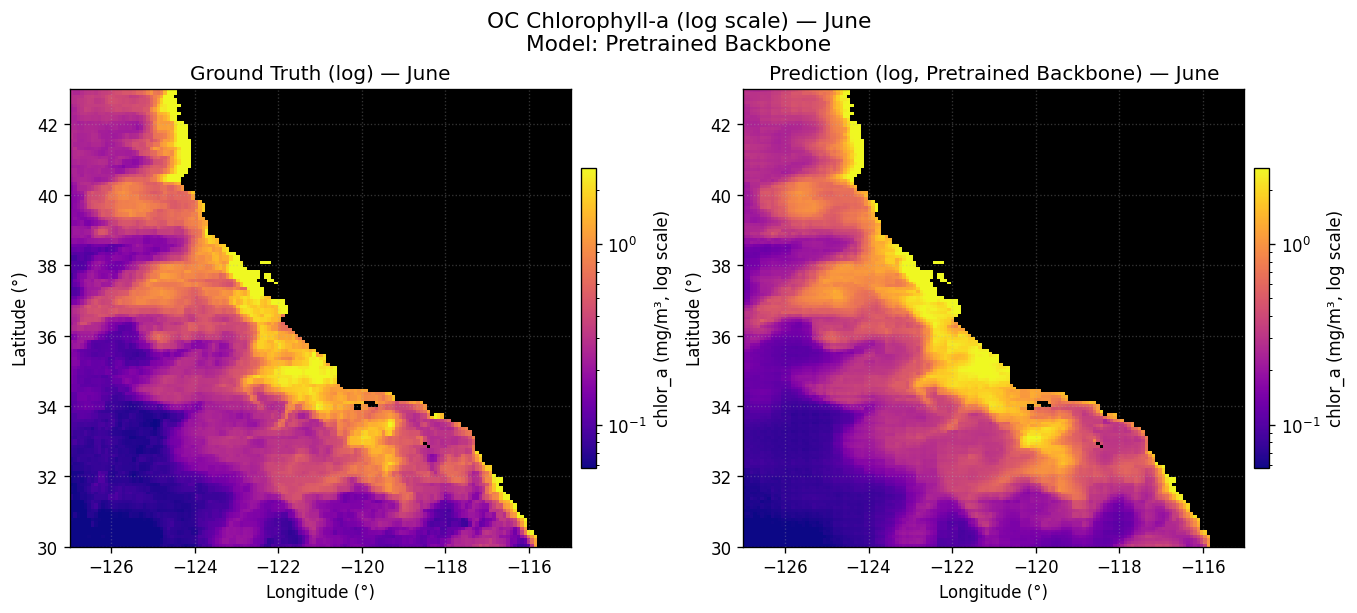}
  \caption{California-extent predicted vs.\ observed chlorophyll-\textit{a} ($\log_{10}$ scale) for June 2022.
           Row order as above.}
  \label{fig:supp_california_pred_log_jun}
\end{figure}

\begin{figure}[H]
  \centering
  \begin{tabular}{@{}cc@{}}
    \includegraphics[width=0.485\textwidth]{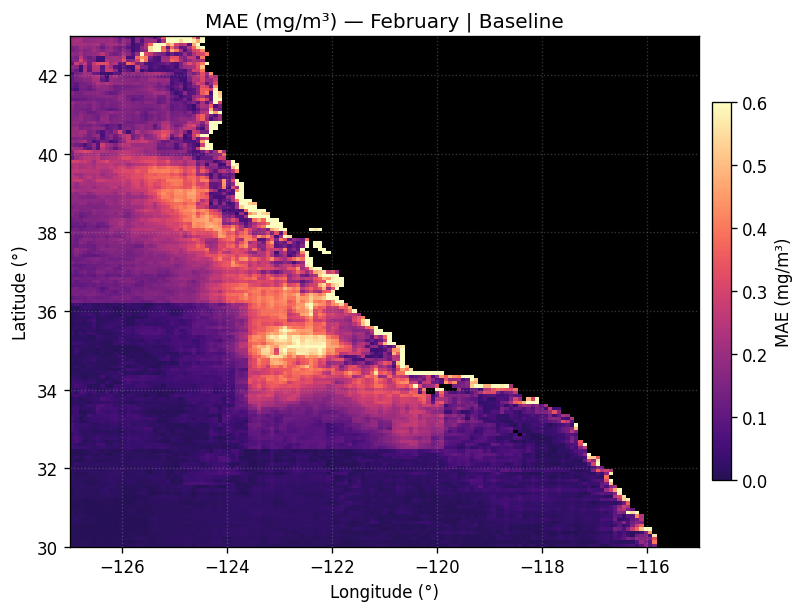} &
    \includegraphics[width=0.485\textwidth]{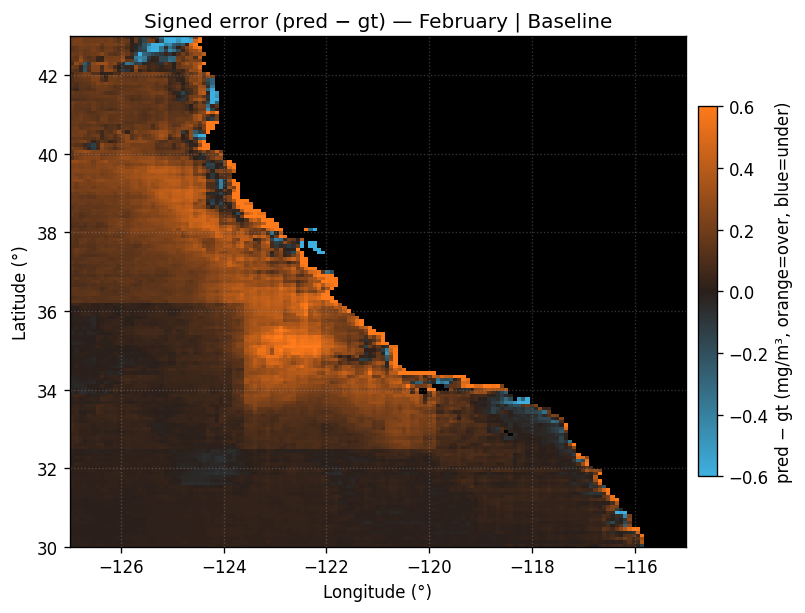} \\[0.2em]
    \includegraphics[width=0.485\textwidth]{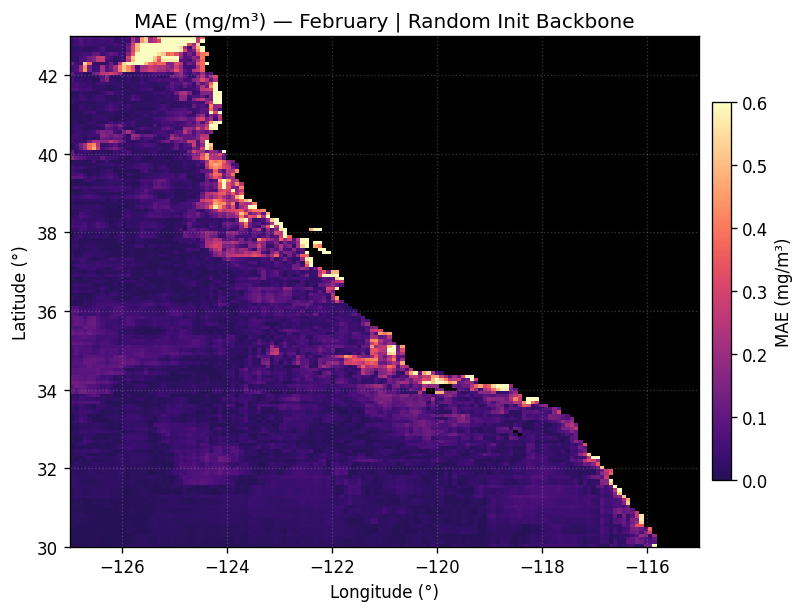} &
    \includegraphics[width=0.485\textwidth]{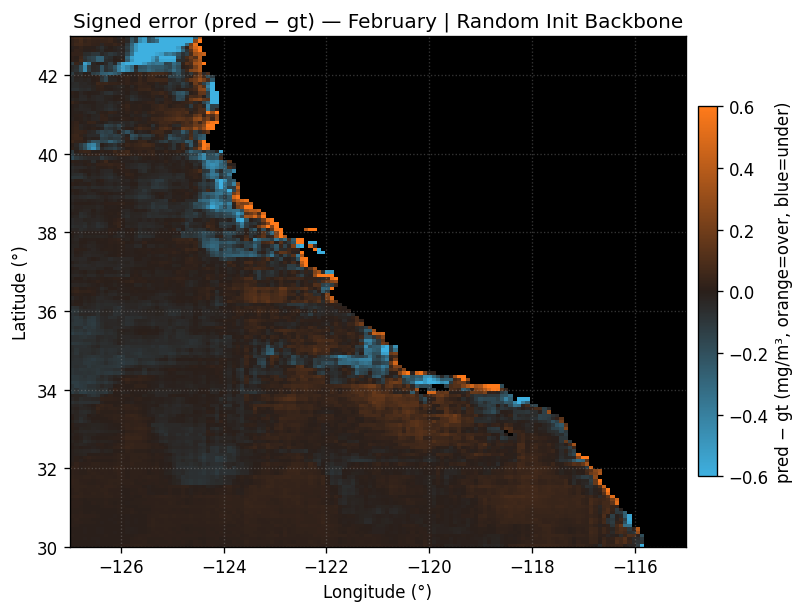} \\[0.2em]
    \includegraphics[width=0.485\textwidth]{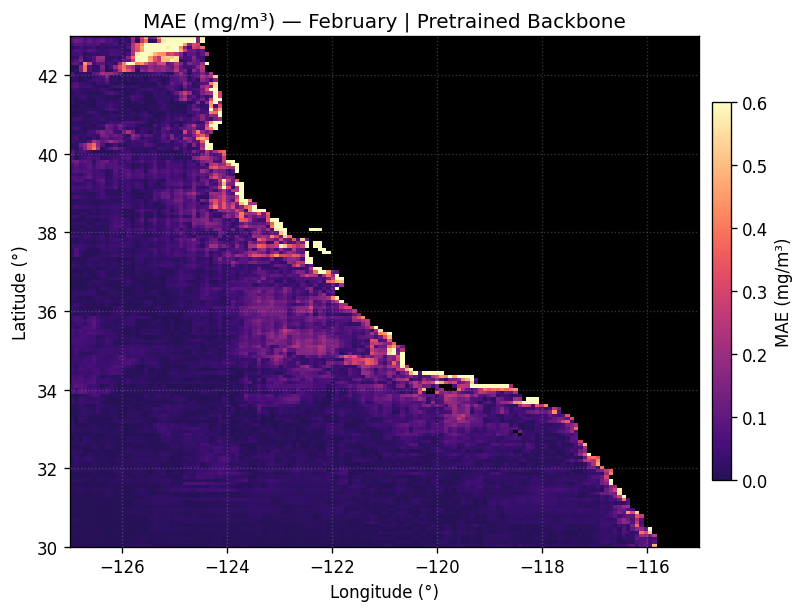} &
    \includegraphics[width=0.485\textwidth]{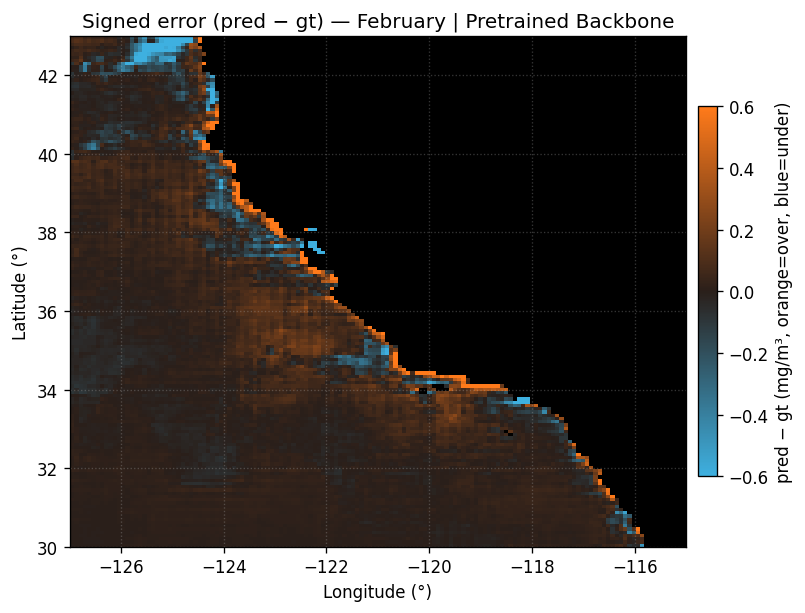}
  \end{tabular}
  \caption{Per-pixel MAE (left: absolute; right: signed) for California, February 2022.
           Rows: Baseline, Random Init, Pretrained DARWIN.}
  \label{fig:supp_california_mae_feb}
\end{figure}

\begin{figure}[H]
  \centering
  \begin{tabular}{@{}cc@{}}
    \includegraphics[width=0.485\textwidth]{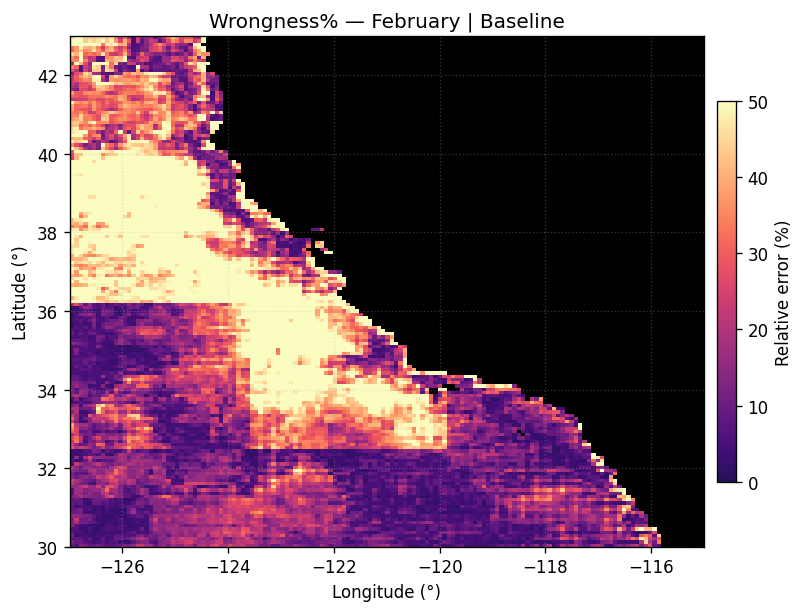} &
    \includegraphics[width=0.485\textwidth]{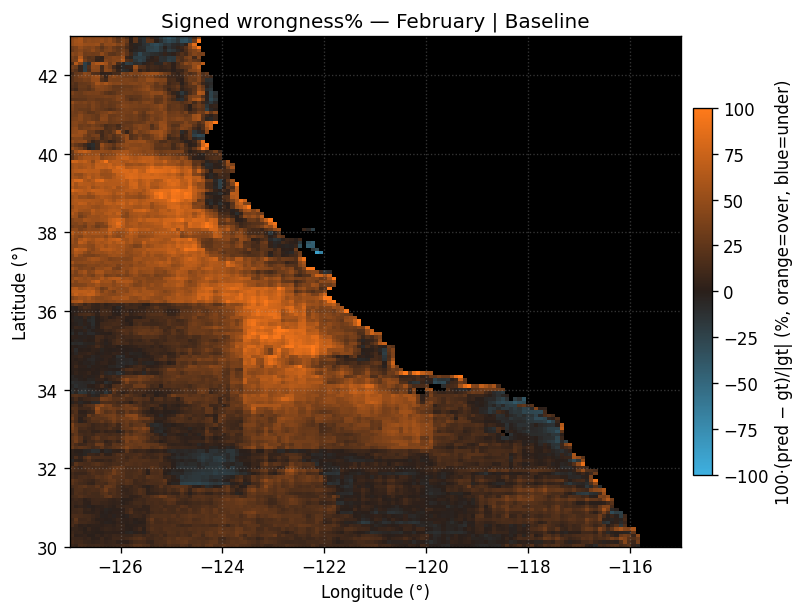} \\[0.2em]
    \includegraphics[width=0.485\textwidth]{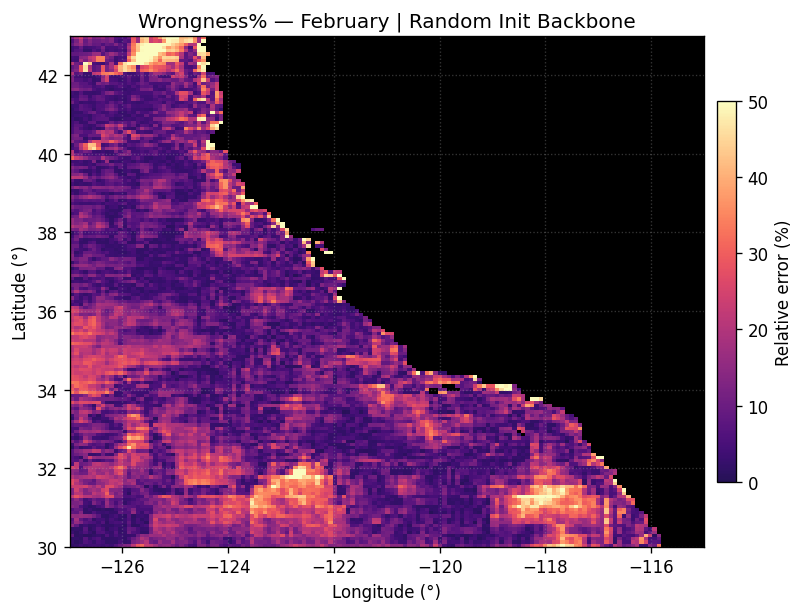} &
    \includegraphics[width=0.485\textwidth]{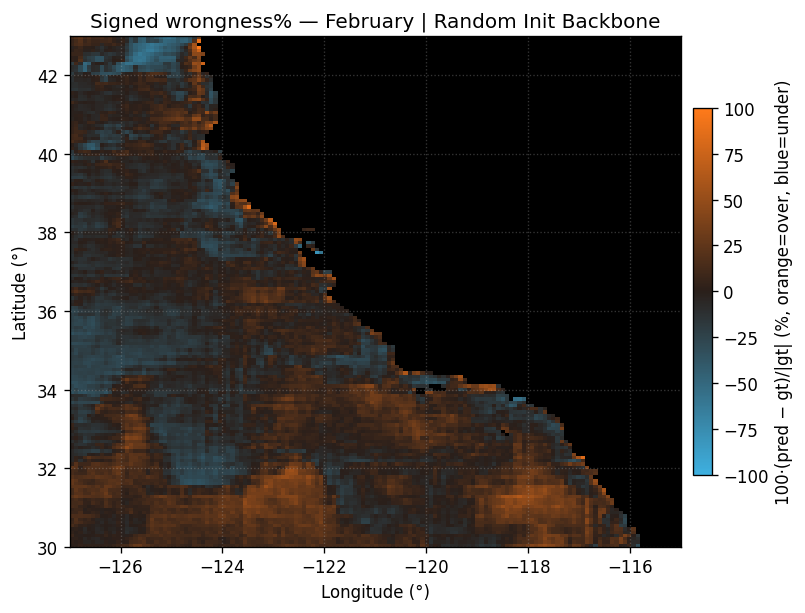} \\[0.2em]
    \includegraphics[width=0.485\textwidth]{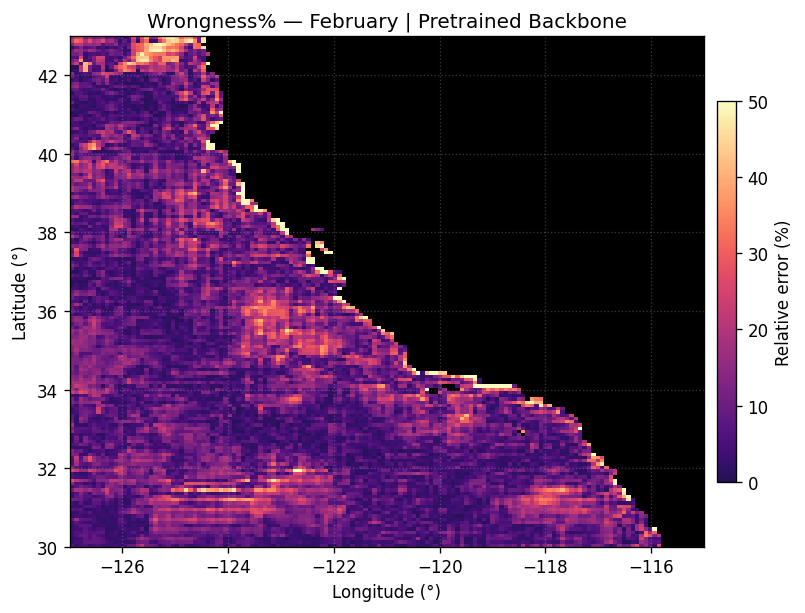} &
    \includegraphics[width=0.485\textwidth]{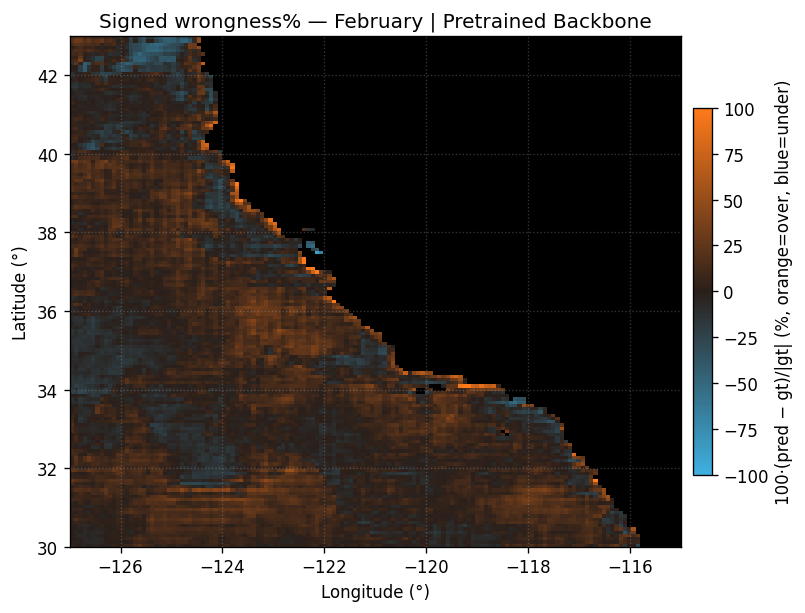}
  \end{tabular}
  \caption{Per-pixel relative error (wrongness, \%) for California, February 2022 (left: absolute; right: signed).
           Rows: Baseline, Random Init, Pretrained DARWIN.}
  \label{fig:supp_california_wrongness_feb}
\end{figure}

\begin{figure}[H]
  \centering
  \includegraphics[width=0.95\textwidth]{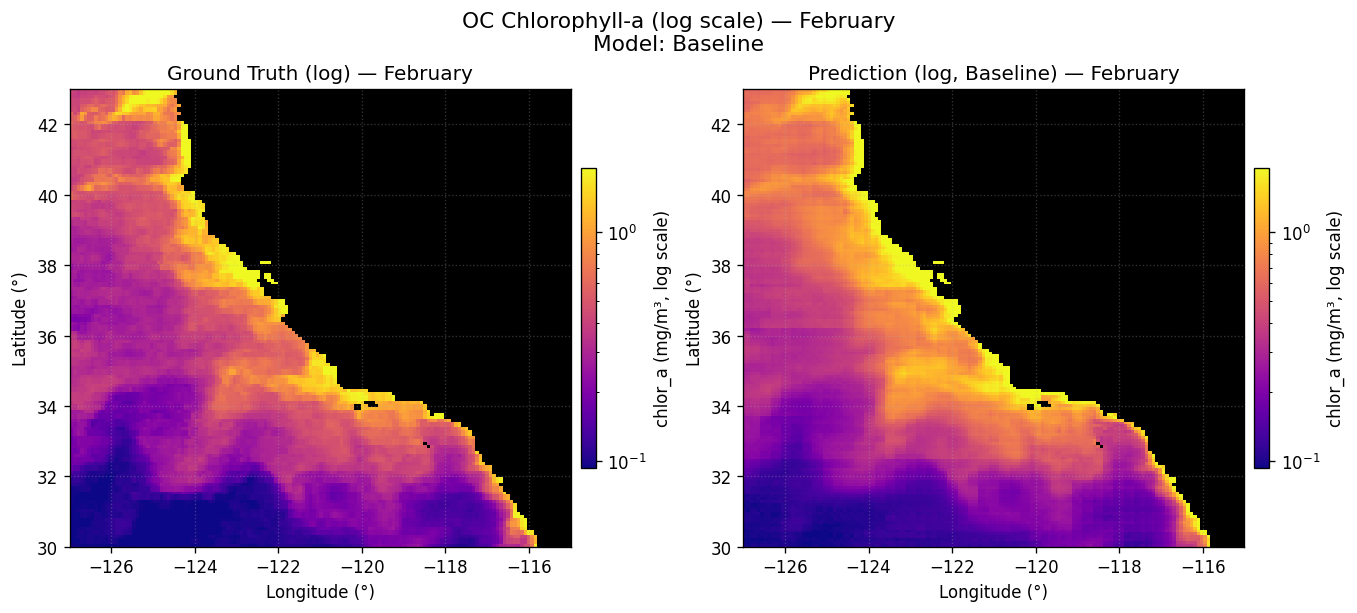}\\[0.3em]
  \includegraphics[width=0.95\textwidth]{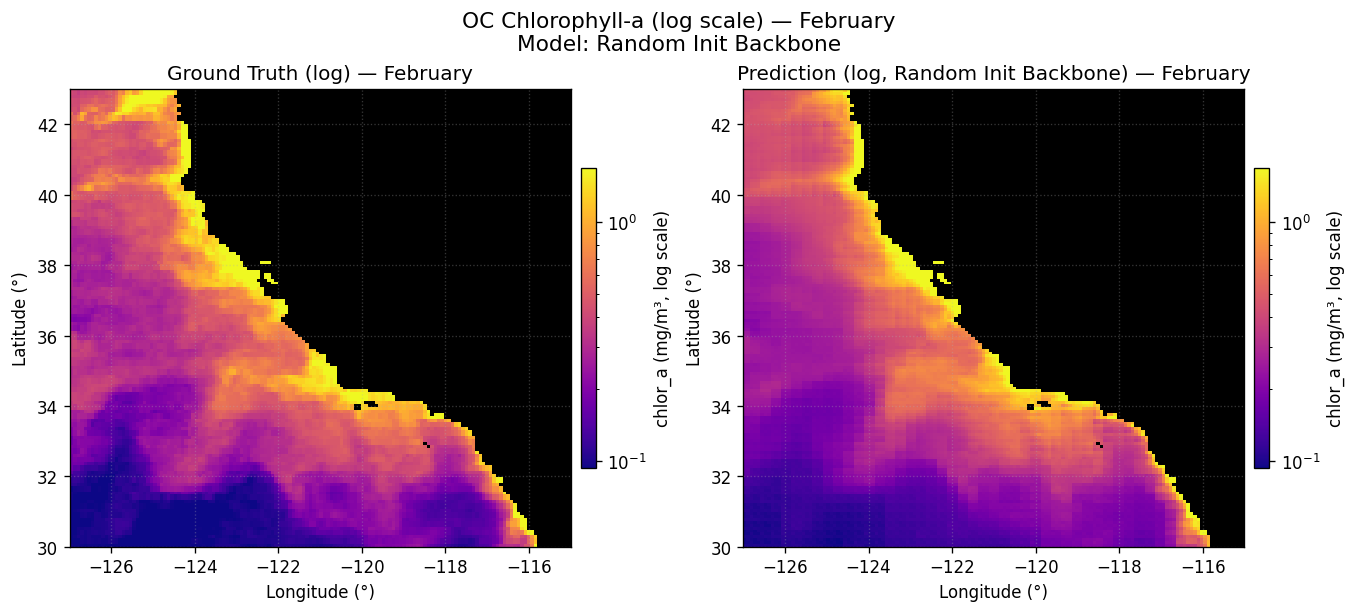}\\[0.3em]
  \includegraphics[width=0.95\textwidth]{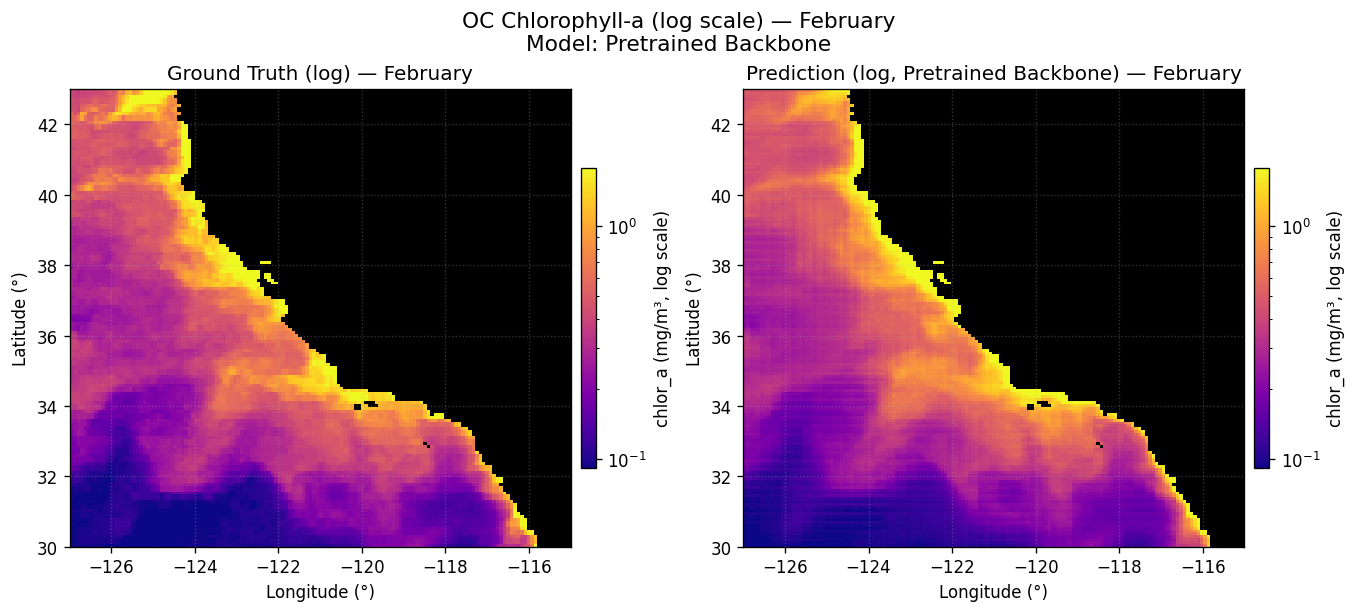}
  \caption{California-extent predicted vs.\ observed chlorophyll-\textit{a} ($\log_{10}$ scale) for February 2022.
           Top to bottom: Baseline, Random Init (unfrozen), Pretrained DARWIN (unfrozen).}
  \label{fig:supp_california_pred_log_feb}
\end{figure}

\section{Supplementary downstream quantitative figures}
\label{appendix:downstream_quantitative}

\begin{figure}[H]
  \centering
  \includegraphics[width=\textwidth]{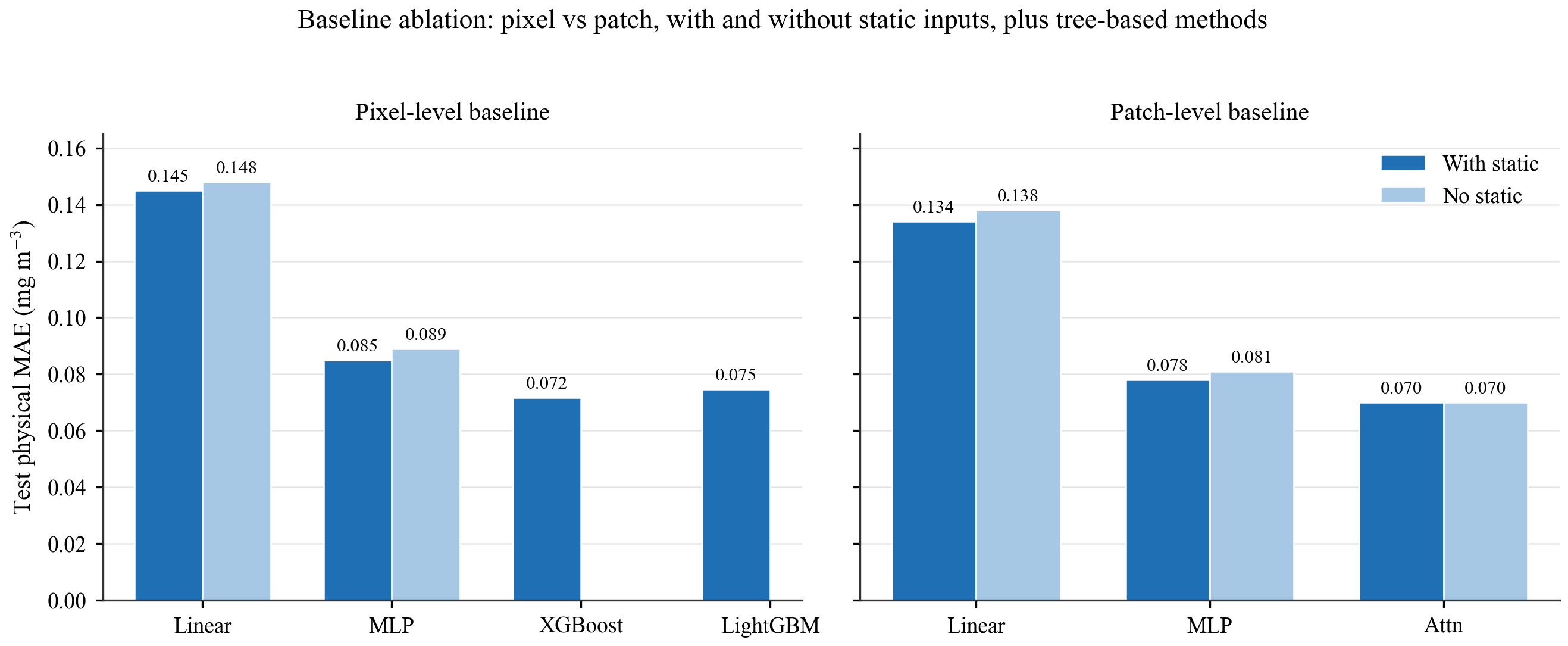}
  \caption{Baseline ablation: test physical MAE for the no-encoder baselines across prediction head, with and without static bathymetry/mask inputs. Pixel-level experiments with transformer-based methods (Attention and Decoder) could not be completed because the resulting sequence length exceeds GPU memory. Adding static features yields a small but consistent improvement across heads; the patch-level baselines outperform their pixel-level counterparts at every head.}
  \label{fig:downstream_baseline_ablation}
\end{figure}

\begin{figure}[H]
  \centering
  \includegraphics[width=\textwidth]{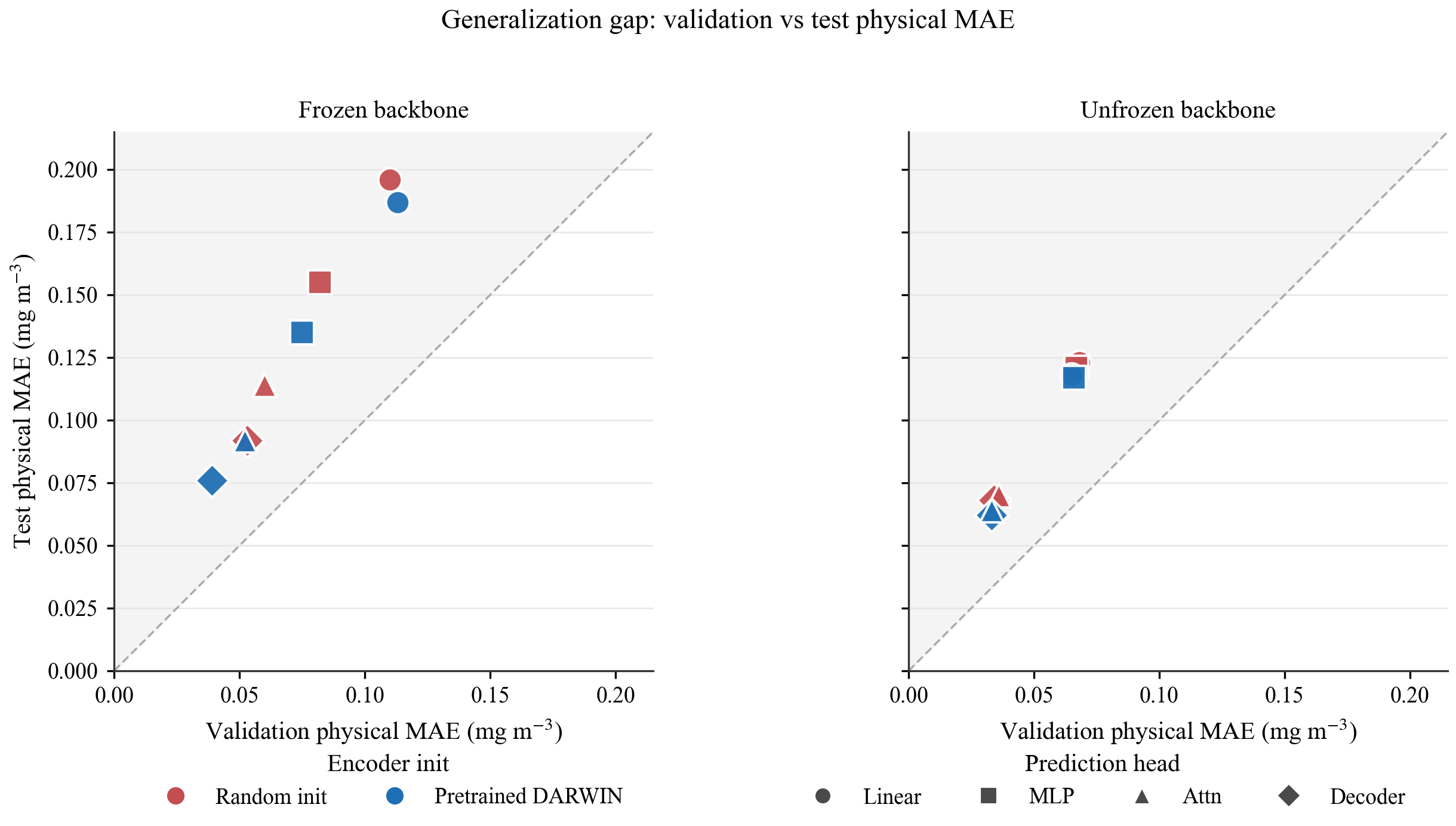}
  \caption{Validation vs.\ test physical MAE for every encoder-based configuration. The dashed $y=x$ line marks the zero-gap reference; the shaded region above the line corresponds to test-time degradation relative to validation. All configurations sit above the line, with comparable gaps across initialisations, suggesting that any DARWIN-vs-random differences are not explained by differential overfitting on validation.}
  \label{fig:downstream_generalization}
\end{figure}

\begin{figure}[H]
  \centering
  \includegraphics[width=\textwidth]{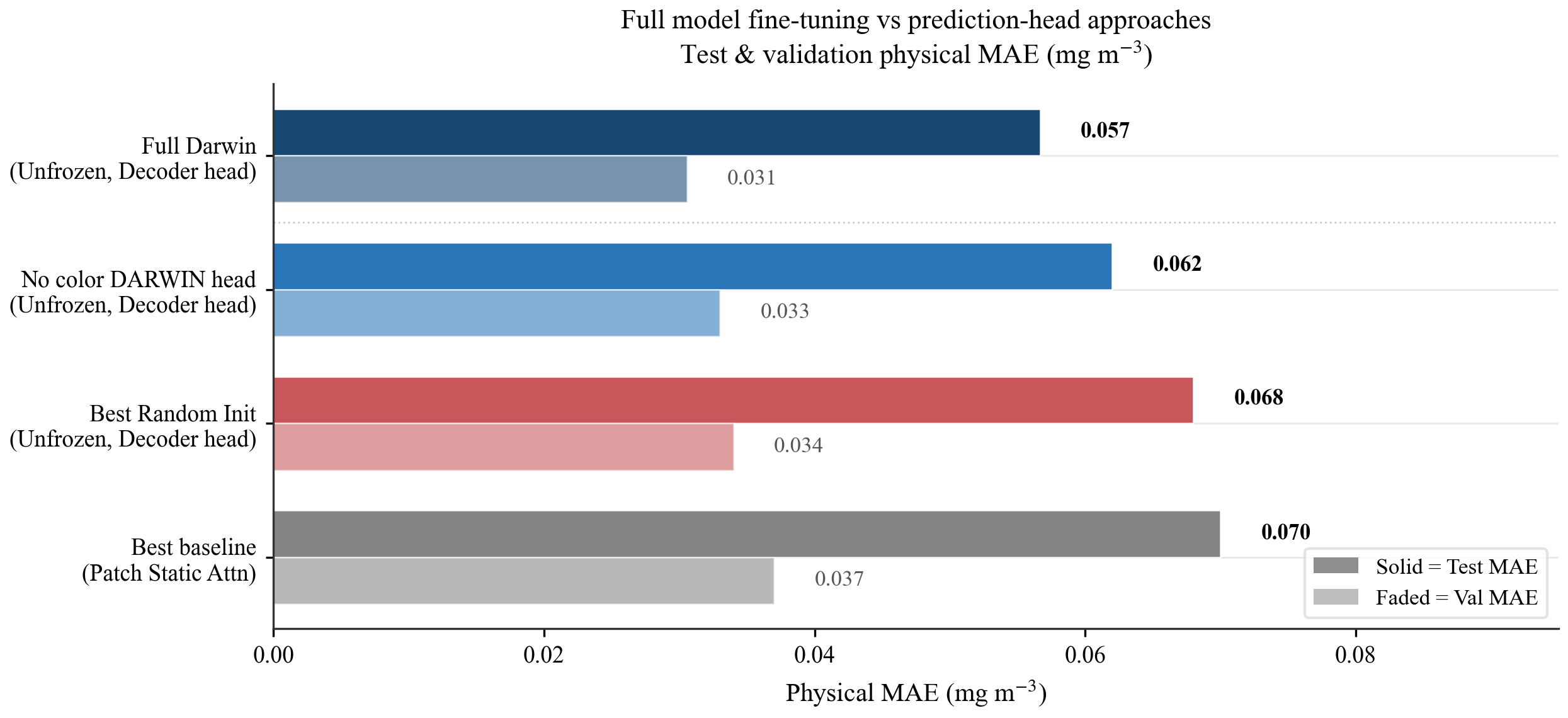}
  \caption{Downstream performance of the full pretrained foundation model compared against the other configurations. The full model attains the strongest downstream performance of all configurations.}
  \label{fig:full_darwin_comparison}
\end{figure}

\end{document}